\documentclass[letterpaper, 10 pt, conference]{ieeeconf}  %

\IEEEoverridecommandlockouts
\usepackage{floatrow}
\usepackage{subfigure}
\usepackage{wrapfig}
\newfloatcommand{capbtabbox}{table}[][\FBwidth]
\usepackage{epsfig}
\usepackage{graphicx}
\usepackage{amsmath}
\usepackage{amssymb}
\usepackage{pifont}%
\newcommand{\cmark}{\ding{51}}%
\newcommand{\xmark}{\ding{55}}%
\usepackage{booktabs}
\usepackage{mathtools}
\usepackage{algorithm,algorithmic}
\usepackage[utf8]{inputenc}
\usepackage{tabularx}
\usepackage{color}
\usepackage[export]{adjustbox}

\newcommand{\squeezeup}{\vspace{-1.5mm}}
\newcolumntype{Y}{>{\centering\arraybackslash}X}

\usepackage[font={footnotesize}]{caption}

\DeclareMathOperator*{\argmin}{arg\,min}
\newcommand{\norm}[1]{\left\lVert#1\right\rVert}

\usepackage[pagebackref=true,breaklinks=true,letterpaper=true,colorlinks,bookmarks=false]{hyperref}

\newcommand{\HB}[1]{{\color{blue}{HB: #1}}} %
\newcommand{\MON}[1]{{\color{green}{MON: #1}}} %
\newcommand{\ANURAG}[1]{{\color{magenta}{Anurag: #1}}} %

\def\etal{\emph{et al}. }
\def\ie{\emph{i.e}. } 
\renewcommand{\ANURAG}[1]{\ignorespaces}
\renewcommand{\HB}[1]{\ignorespaces}
\renewcommand{\MON}[1]{\ignorespaces}

\title{\LARGE \bf
Meta-Learning Deep Visual Words for Fast Video Object Segmentation
\author{Harkirat Singh Behl$^{1}$, Mohammad Najafi$^{1}$, Anurag Arnab$^{2}$ and Philip H.S. Torr$^{1}$}
\thanks{$^{1}$University of Oxford.
        $^{2}$Google Research. Work primarily done at Oxford.
    }%
}
\begin{document}
\maketitle

\begin{figure*}
\centering
\includegraphics[width=\textwidth]{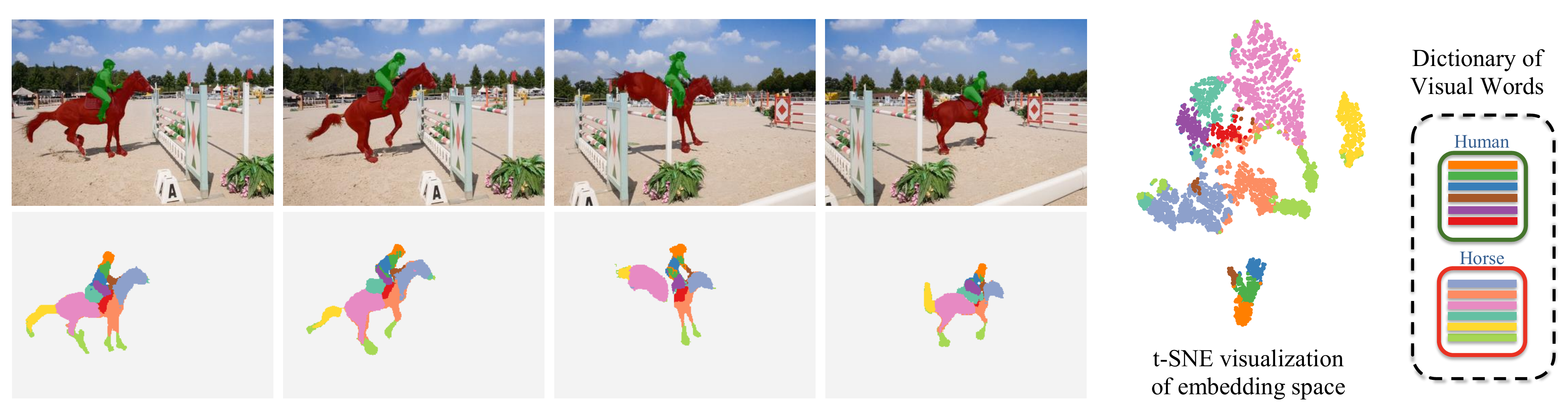}
\caption{\textbf{Video object segmentation using a dictionary of deep visual words.}
      Our proposed method represents an object as a set of cluster centroids in a learned embedding space, or ``visual words'', which correspond to object parts in image space (bottom row). 
      This representation allows more robust and efficient matching as shown by our results (top row).
      The visual words are learned in an unsupervised manner, using meta-learning to ensure the training and inference procedures are identical.
      The t-SNE plot \cite{maaten_jmlr_2008} on the right shows how different object parts cluster in different regions of the embedding space,
      and thus how our representation captures the multi-modal distribution of pixels constituting an object. 
      \label{fig:teaser}
     }
     \vspace{-0.75\baselineskip}
\end{figure*}

\thispagestyle{empty}
\pagestyle{empty}
\begin{abstract}
Personal robots and driverless cars need to be able to operate in novel environments and thus quickly and efficiently learn to recognise new object classes.
We address this problem by considering the task of video object segmentation.
Previous accurate methods for this task finetune a model using the first annotated frame, and/or use additional inputs such as optical flow and complex post-processing.
In contrast, we develop a fast, causal algorithm that requires no finetuning, auxiliary inputs or post-processing, and segments a variable number of objects in a single forward-pass.
We represent an object with clusters, or ``visual words'', in the embedding space, which correspond to object parts in the image space.
This allows us to robustly match to the reference objects throughout the video, because although the global appearance of an object changes as it undergoes occlusions and deformations, the appearance of more local parts may stay consistent.
We learn these visual words in an unsupervised manner, using meta-learning to ensure that our training objective matches our inference procedure.
We achieve comparable accuracy to finetuning based methods (whilst being 1 to 2 orders of magnitude faster), and state-of-the-art in terms of speed/accuracy trade-offs on four video segmentation datasets.
Code is available at \url{https://github.com/harkiratbehl/MetaVOS}.
\end{abstract}

\section{Introduction}
\vspace{-0.2cm}

Personal robots and driverless cars need to be able to operate in novel environments, and thus be able to quickly and efficiently learn to recognise object categories that they were not originally trained on.
Furthermore, detailed segmentations of objects are also required for applications such as robot manipulation~\cite{kenney_icra_2009, siam_icra_2019}, grasping and learning object affordances~\cite{do_icra_2018, siam_icra_2019}.
Finally, as live camera streams are processed in such applications, efficient and causal algorithms are required.
This paper addresses these problems by considering the task of video object segmentation, following the protocol defined in the DAVIS datasets~\cite{Caelles_arXiv_2018, davis_2017}.
Here, the ground-truth object mask of one or more objects are provided only in the first frame, which must then be tracked at a pixel-level throughout the rest of the video.
Since obtaining even a pixelwise segmentation for a single-frame may be too onerous in robotics applications, we also further extend the problem definition to only provide a bounding-box of each object in the first frame.

Accurate approaches to video segmentation trained a fully convolutional network (FCN) \cite{fcn} for foreground/background segmentation on existing datasets, and then adapted it to the testing video by finetuning the network on the first, fully-annotated frame ~\cite{OSVOS, DAVIS2018-Semi-Supervised-1st, onavos, lucid, Yoon_2017_ICCV, Ci_2018_ECCV}.
Although these methods produce accurate results (and can be improved further by using optical flow \cite{Tsai_2016_CVPR, Bao_2018_CVPR, Cheng_2017_ICCV, hu2017maskrnn, Xiao_2018_CVPR} or post-processing with DenseCRF \cite{densecrf, OSVOS, Bao_2018_CVPR, Cheng_2018_CVPR}), they are extremely time consuming, taking between 700s to 3h to finetune per DAVIS video \cite{OSVOS,lucid}, rendering them unsuitable for real-life applications and robotics.

This paper, in contrast, considers the more challenging (and practical) scenario where the network is not finetuned at all, and uses no optical flow or extra post-processing, in order to develop a fast and causal algorithm.
Our approach is inspired by metric-learning methods which embed pixels from the same object close to each other in a learned embedding space, and pixels from different objects far apart.
Chen~\etal\cite{Chen_2018_CVPR} used this idea %
to formulate video segmentation as a pixel-level retrieval task, where each pixel of the ground-truth mask was embedded in the first frame to form an index, and pixels in subsequent frames were classified with nearest neighbours.
Contrastingly, in the related context of few-shot learning, Prototypical networks \cite{snell_2017} represent each class with the mean of their embeddings and classify subsequent queries with a softmax over distances to each prototype.

Prototypical networks, although simple and fast, do not have sufficient capacity to model complex, multi-modal data distributions such as an object in a video that undergoes deformations, occlusions and viewpoint changes.
Nearest neighbour approaches \cite{Chen_2018_CVPR,Li_2018_CVPR, DAVIS2018-Interactive-2nd,Hu_2018_ECCV} have greater modelling capacity, but are more computationally expensive as the time and memory cost to perform a lookup grows linearly with the size of the index.
For pixel-level tasks, they also store many redundant pixels with similar appearance in the index.
Furthermore, they are more prone to overfitting and noise, which becomes more prevalent during the ``online adaptation'' of video segmentation models to account for variations throughout the video \cite{Chen_2018_CVPR, Hu_2018_ECCV,onavos,Ci_2018_ECCV}. %

Our flexible approach interpolates the spectrum of metric learning approaches by representing an object with a fixed number of cluster centroids in the embedding space.
We denote this as a dictionary of visual words, because each cluster centroid in the embedding space corresponds to a part of the object in the image space as shown in Fig.~\ref{fig:teaser}, even though these words are formed in an unsupervised manner.

The use of visual words enables more robust matching, because even though an object as a whole may be subject to occlusions, deformations, viewpoint changes, or disappear and reappear from the same video, the appearance of some of its more local parts may stay consistent.
Moreover, the robustness of this approach allows us to easily extend it to the scenario where we only have weak bounding-box supervision in the first frame.

These visual words are learned without any explicit supervision by clustering our embedding space, and using meta-learning to ensure that our training objective matches our inference procedure.
This is in contrast to related metric-learning based approaches \cite{Chen_2018_CVPR, DAVIS2018-Interactive-2nd,Li_2018_CVPR} which are trained with surrogate, and sometimes unstable, losses. 
Furthermore, as our method requires only a single forward-pass to segment a variable number of objects per video, it naturally scales to the multi-object setting.
Related methods \cite{Oh_2018_CVPR}, in contrast, segment each object independently before combining results and are thus slower for multiple objects.

The advantages of our simple and intuitive approach is reflected by its performance on multiple single- and multi-object video segmentation datasets (DAVIS 2016 \cite{DAVIS}, DAVIS 2017 \cite{davis_2017}, SegTrack v2 \cite{FliICCV2013}, YouTube-Objects \cite{youtube1, youtube2}) where we achieve comparable accuracy to finetuning-based methods (whilst being 1 to 2 orders of magnitude faster), and lie on the Pareto front as no other published methods to our knowledge are both faster and more accurate.

\section{Related Work}
\paragraph{Fine-tuning based approaches} The most accurate video segmentation methods using the DAVIS protocol \cite{DAVIS, davis_2017} currently finetune models on the first frame of the video \cite{DAVIS2018-Semi-Supervised-1st,OSVOS,onavos,lucid} and/or use optical flow \cite{hu2017maskrnn, Xiao_2018_CVPR, Cheng_2017_ICCV, Tsai_2016_CVPR} or DenseCRF \cite{OSVOS, Bao_2018_CVPR, Cheng_2018_CVPR} post-processing, or use self-paced learning \cite{zhang2017spftn} to improve performance.
Our proposed approach does not involve finetuning, or additional information such as optical flow, and still achieves comparable performance whilst being one to two orders of magnitude faster.

\paragraph{Fast approaches} Fast approaches to video segmentation, that do not finetune on the first frame or use optical flow, can broadly be divided into methods performing mask propagation or metric learning.
Mask propagation methods, such as \cite{masktrack, Oh_2018_CVPR,Jampani_2017_CVPR}, use the segmentation mask from the one frame to guide the network to predict the mask in the next frame (\ie pixel-level tracking).
These methods use the prior that objects move smoothly and slowly over time, and thus struggle when there are temporal discontinuities like occlusion or rapid motion.
Moreover, errors accumulate over time as the model ``drifts'', particularly if the algorithm loses track of the object.
Li \etal \cite{Li_2018_ECCV} addressed this issue using re-identification modules which traverse the video back-and-forth to recover any potential missed objects. %
However, this method is not causal as it looks at future frames.
Oh \etal \cite{Oh_2018_CVPR} do not only use the previous frame, but also the first reference frame, to guide the tracking. %
However, this does not completely alleviate the problem of model drift, as if the model loses track of the object, its appearance may have changed so much from the first frame that the reference frame is not effective in recovering it.
Moreover, since these methods match the entire object as a whole, they struggle with occlusions.
This is in contrast to our approach which represents objects by their constituent parts to be more robust to appearance changes.
Finally, \cite{Oh_2018_CVPR} is designed for tracking a single object, and thus handling multiple objects require processing each object individually before heurstically merging results.
Our method in comparison segments multiple objects in a single-forward pass.

\paragraph{Metric learning based approaches} Our work is more similar to methods using pixel-to-pixel matching or metric learning \cite{Chen_2018_CVPR, Hu_2018_ECCV, DAVIS2018-Interactive-2nd, Yoon_2017_ICCV,Li_2018_CVPR}. %
Chen \etal \cite{Chen_2018_CVPR} formulated video segmentation as a pixel-level retrieval problem, where embeddings from the first reference frame are used to form an index for a nearest neighbour classifier.
Note that the method of \cite{Chen_2018_CVPR} is trained with a variant of the triplet loss, which though common for metric learning, does not optimise explicitly for the nearest neighbour search at inference time.
The triplet loss is also difficult to train with, as it is very sensitive to triplet selection \cite{schroff_cvpr_2015,hermans_arxiv_2017}.
Siamese networks have also been employed in a similar manner \cite{Hu_2018_ECCV, DAVIS2018-Interactive-2nd, Yoon_2017_ICCV}, where one branch computes embeddings from the annotated first frame which are used to match to the embeddings computed by the other branch on the current frame.
When classifying query images, these methods all effectively search all the pixels from the reference frame. 
This approach is not only expensive in terms of time and memory (as it retains redundant embeddings of similar pixels), but is also more susceptible to noise.
This is an issue during the ``online adaptation'' \cite{Chen_2018_CVPR,Ci_2018_ECCV,Hu_2018_ECCV,onavos} of the model which may introduce incorrectly labelled embeddings.
Our method retains only cluster centroids in the embedding space (which correspond to exemplars of object parts), which enables faster and more robust matching.

Taking inspiration from classical computer vision, learned feature descriptors have been used in localization \cite{sarlin2019coarse}, motion removal \cite{sun2018motion} and feature matching \cite{ono2018lf} because of their robustness and efficiency. Note that our main contributions are that we can learn to segment new object classes in video from very few examples, and use a meta-learning technique to train our model so that the training and testing procedures match each other.
The utility of object parts for more robust matching for video segmentation has been identified before by \cite{Cheng_2018_CVPR}.
However, Cheng \etal \cite{Cheng_2018_CVPR} use handcrafted heuristics to form object parts based on bounding boxes in the image space.
In contrast, we cluster our embedding space in an unsupervised manner to obtain visual words which resemble object parts (as pixels with similar appearance cluster together).
Moreover, the method of Cheng \etal \cite{Cheng_2018_CVPR} -- which tracks bounding boxes of object parts and then merges foreground segmentations within these boxes -- consists of two separately trained modules (using different datasets), whereas our method is trained via meta-learning with a single objective function that matches our inference procedure. 

\paragraph{Meta-learning} Finally, we note that meta-learning has not been explored much in the context of video segmentation.
Yang \etal \cite{Yang_2018_CVPR} used meta-learning to adapt the weights of the final layer of a segmentation network at test time.
This is in contrast to our method which adaptively computes the initial dictionary of visual words from the first labelled frame in the video.
Note that our method can be viewed as a generalisation of Protoypical networks \cite{snell_2017} and Matching networks \cite{vinyals_2016} for few-shot classification.
Prototypical networks represents the training data from each class as a single prototypical vector.
Matching networks, on the opposite end of the spectrum, 
consider all training data samples of a particular class to make a classification regardless of how redundant or noisy these samples may be.
Our method interpolates these two methods by representing an object class via a fixed number of cluster centroids, which correspond to exemplars of object parts in the case of video segmentation.

\vspace{-0.3\baselineskip}
\section{Proposed Approach}\vspace{-0.2\baselineskip}
\begin{figure*}[]
	\vspace{-0.1\baselineskip}
	\centering
	\includegraphics[width=\textwidth]{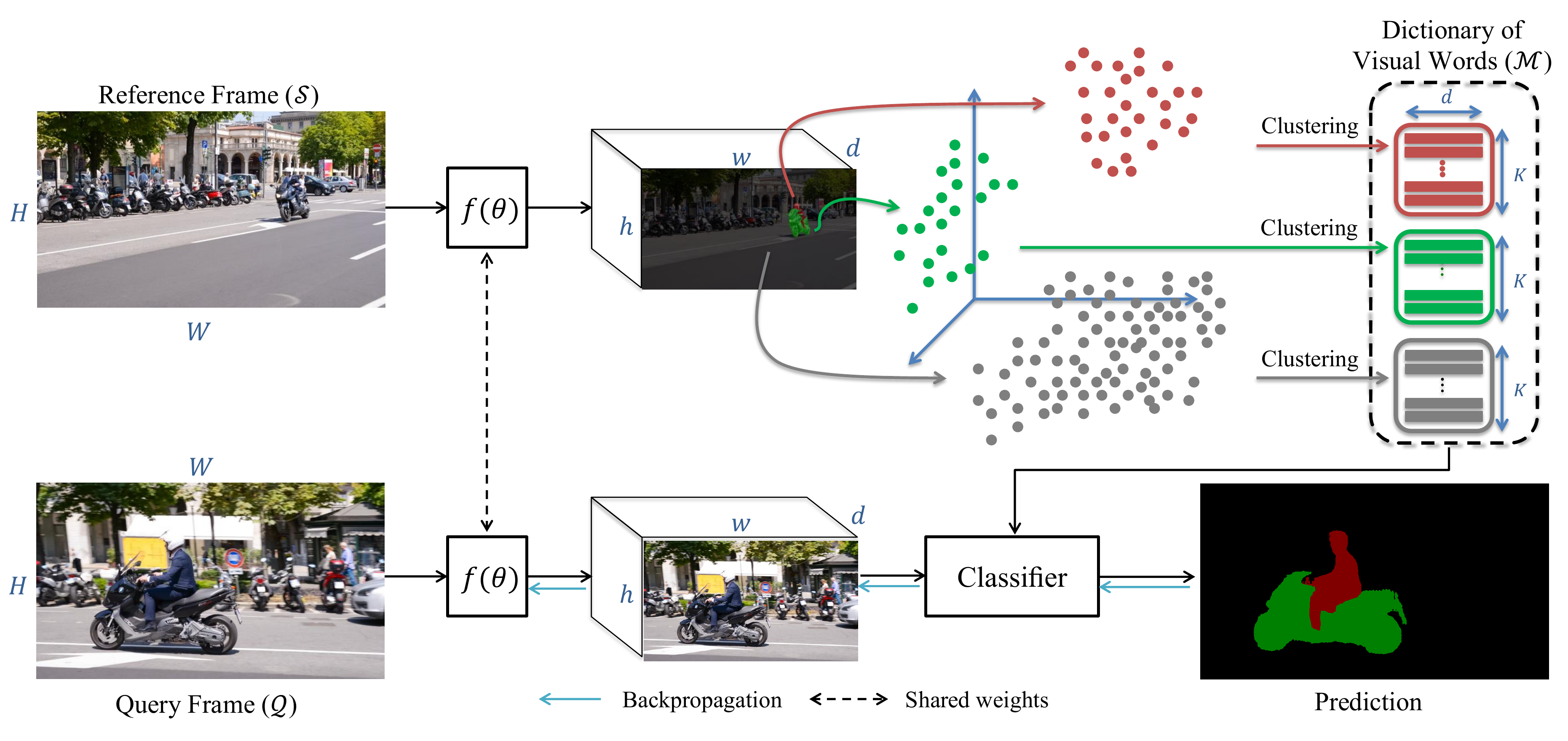}
	\caption{{\bf Overview of the proposed method.} 
	The first frame of the video (reference frame), which forms the support set $\mathcal{S}$ in our meta-learning setup, passes through a deep segmentation network $f(\theta)$ to compute a $d = 128$ dimensional embedding for each pixel. 
	A dictionary of deep visual words are then learned by clustering these embeddings for each object in the reference frame (Eq.~\ref{eq:unsupervised}). 
	Pixels of the query frame are classified as one of the objects based to their similarities to the visual words (Eq.~\ref{eq:cluster_prob} and Eq.~\ref{eq:label_prob}). 
	The model is meta-trained by alternately learning the visual words given model parameters $\theta$, and learning model parameters given the visual words. During testing, the bottom path, without blue lines, is applied to all frames, whereas, the top path is only applied to the first frame and the online adaptation frames.
	}
	\vspace{-1\baselineskip}
	\label{fig:ourmodel}  
\end{figure*}

We first describe the formulation of video object segmentation as a meta-learning problem.
This allows us to train our model in same way that it will be tested, unlike other metric-learning based approaches to video segmentation~\cite{Chen_2018_CVPR, Hu_2018_ECCV}. %

\subsection{Video Object Segmentation as Meta-Learning}
\label{sec:video_seg_metalearning}

Meta-learning, or learning to learn, is often defined as learning from a number of tasks in the training set, to become better at learning a new task in the test set \cite{schmidhuber_thesis_1987,naik_ijcnn_1992,hochreiter_icann_2001,MAML}.
In the context of video object segmentation, the task is to learn from the ground-truth masks of the objects in the first frame of the video (support set) to segment and track them in rest of the video (query set).
Our meta-learning objective is to learn model parameters $\theta$ on a variety of tasks (videos), which are sampled from the distribution $p(\mathcal{T})$ of training tasks (\ie meta-training set), such that the learned model performs well on a new unseen task (test video).
Denoting the loss of the model on the $n^{th}$ task, $\mathcal{T}_n$, as $\mathcal{L}_{\mathcal{T}_n}(\theta)$, the meta-training objective is thus
\begin{align}
\theta^*
&= \argmin_\theta \sum_{\mathcal{T}_n \sim p(\mathcal{T})} \mathcal{L}_{\mathcal{T}_n}(\theta).
\label{eq:loss}
\end{align}
The support set $\mathcal{S}$ is the set of all labeled pixels in the first frame, $\mathcal{S} = \{x_i, y_i \}_{i=1}^{N}$. 
Here $x_i$ represents the pixel $i$ in the first frame, $y_i \in \mathcal{C} = \{ 1,...,C \}$ is the ground truth class label of pixel $x_i$, $N$ is the number of labeled pixels in the frame, and $C$ is the number of object classes that need to be tracked and segmented in the video.
Similarly the query set is defined by $ \mathcal{Q} = \{ x_j, y_j \}_{j=1}^{N_Q}$, where $N_Q$ is the number of labelled pixels in the video after the first frame, $j$ is the index.
The output of each task $\mathcal{T}$ is the set of predicted class labels for the pixels in $ \mathcal{Q}$, $\mathcal{\hat{Y}} = \{\hat{y}_j\}_{j=1}^{N_Q}$. 

Next, we describe our model for estimating the outputs of each task, \ie the object label for every pixel in the query frames of the video.
\vspace{-0.3\baselineskip}
\subsection{Model}
\label{sec:model}
\vspace{-0.2\baselineskip}
In order to predict the label for each pixel in the query set $\mathcal{Q}$, we need to learn a representation for each object using information from the support set $\mathcal{S}$.
We represent each object in the video using a dictionary of deep visual words.
Each pixel in the query set is then labelled based on the deep visual word that it is assigned to.
Our method is an online method does not use future-frames during runtime, i.e, to segment a new frame, only the information upto that point is used.

Learning visual words is a challenging task, as we do not have any ground truth information of the object parts that they correspond to.
Consequently, as summarised in Fig.~\ref{fig:ourmodel}, we use a meta-training algorithm, where we alternate between the unsupervised learning of deep visual words (Sec.~\ref{sec:unsupervised_learning_visual_words}) and supervised learning of pixel classification given these visual words (Sec.~\ref{sec:supervised_pixel_classification}).
Our model thus learns to learn a better classifier, by optimising the visual words that it will produce at test-time.

\subsubsection{Unsupervised Learning of Deep Visual Words}
\label{sec:unsupervised_learning_visual_words}

We initially pass the first frame of the video, which is the support set $\mathcal{S}$, through a deep neural network $f(\theta)$, which is an artificial neural network with multiple layers between input and output, to compute the embedding for each pixel $x_i$ in $\mathcal{S}$, $f_\theta(x_i)$. 
We then compute a set of deep visual words for all the pixels in each object class. 
Let $\mathcal{S}_c$ be the set of pixels in $\mathcal{S}$ with class label $c$. 
Each set $\mathcal{S}_c$ is partitioned into $K$ clusters $\mathcal{S}_{c1}, ...., \mathcal{S}_{cK}$ using the k-means algorithm~\cite{kmeans}, with $\mu_{ck}$ being the respective centroids of the clusters, using the objective:
\begin{subequations}\label{eq:unsupervised}
	\begin{equation}
	\mathcal{S}_{c1}, ...., \mathcal{S}_{cK} = \argmin_{\mathcal{S}_{c1}, ...., \mathcal{S}_{cK}} \sum_{k=1}^{K}\sum_{x_i \in \mathcal{S}_{ck}} \left\lVert f_\theta(x_i) - \mu_{ck} \right\rVert _2^2, \\
	\end{equation}
	\begin{equation}\label{eq:mu}
	\mu_{ck} = \frac{1}{\lvert \mathcal{S}_{ck} \rvert}\sum_{x_i \in \mathcal{S}_{ck}} f_\theta(x_i).
	\end{equation}
\end{subequations}
In other words, we represent the distribution of the pixels within each set $\mathcal{S}_c$ in the learned embedding space with a set of deep visual words $\mathcal{M}_c = \{\mu_{c1},...,\mu_{cK}\}$.
We can, in principle, use any clustering algorithm here and choose k-means as it is computationally efficient and simple.

\subsubsection{Supervised Learning for Pixel Classification}
\label{sec:supervised_pixel_classification}
Once the deep visual words for each object have been constructed, the probability of assigning a pixel $x_j \in \mathcal{Q}$ to the $k^{th}$ visual word from the $c^{th}$ object class is computed using a non-parametric softmax classifier,
\begin{equation}
p(c_k|x_j) =  \frac{\exp\big(cos(\mu_{ck}, f_{\theta}(x_j))\big)}{\sum_{\mu_{i} \in \mathcal{M}}\exp\big(cos(\mu_{i}, f_{\theta}(x_j))\big)},
\label{eq:cluster_prob}
\end{equation}
where $\mathcal{M} = \bigcup_{c=1}^{C} \mathcal{M}_c$ is the dictionary of deep visual words for all objects present in the video, and $cos$ is the cosine similarity function.
We enable our model to account for intra-class variations by encouraging each pixel to resemble only one relevant visual word. %
As a result, the probability of pixel $x_j$ taking the object class label $c$ is defined as
\begin{equation}
p(\hat{y_j}=c|x_j) = \frac{\operatorname*{max}_{k \in \{1,..,K\}}p(c_k|x_j)}{\sum_{c'=1}^{C} \operatorname*{max}_{k \in \{1,..,K\}}p(c'_k|x_j)}.
\label{eq:label_prob}
\end{equation}
This allows our model to learn meaningful visual words that correspond to the diverse object parts that constitute an object.
Note from the T-SNE visualisation of our embeddings in Fig.~\ref{fig:teaser} that pixels from different parts of the same object cluster in separate regions of the embedding space. Finally, our loss function for this pixel classification problem is the cross-entropy loss.

\vspace{-0.4\baselineskip}
\subsection{Meta-training procedure}
\label{sec:meta_training}
\vspace{-0.3\baselineskip}
Each iteration of our meta-training algorithm consists of an unsupervised learning process to construct a dictionary of visual words from the support set $\mathcal{S}$, followed by a supervised learning step where the segmentation network parameters, $\theta$, are updated by minimsing the cross-entropy loss function according to Eq.~\ref{eq:loss}.
In other words, the model learns to learn deep visual words in the first frame of the video to minimise a pixel-level loss over the rest of the video.

Our method is a form of non-parameteric meta-learning, as described in \cite{meta_learning_tutorial_2019}. Note that the cluster centroids can be seen as the parameters of the final classification layer (Eq.~\ref{eq:cluster_prob}).
Prototypical networks \cite{snell_2017} represent each class with a single prototypical vector (\ie one visual word), whilst Matching networks \cite{vinyals_2016} represent each class using all the samples of that class in the support set (\ie the embedding of each pixel in $\mathcal{S}$ would form a visual word).
Our method interpolates between these two approaches to build a more robust representation of the support set $\mathcal{S}$.
Also note that previous metric-learning approaches to video object segmentation such as \cite{Chen_2018_CVPR} learn an embedding using variants of the triplet loss and perform nearest neighbour classification at test time. %
This approach is thus similar to Matching networks \cite{vinyals_2016}, with the key difference being that the training objective (triplet loss) does not correspond to the inference procedure (nearest neighbour search), which is an essential component for meta-learning \cite{meta_learning_tutorial_2019}. Our method ensures that meta-train and meta-test setup match.

\vspace{-0.3\baselineskip}
\subsection{Online Adaptation}
\label{sec:online_adaptation}
\vspace{-0.2\baselineskip}
The objects of interest from the first frame, as well as the  background, often undergo deformations, occlusions, viewpoint changes.
As a result, adapting the model throughout the video is vital to achieve good performance and done by state-of-art video methods \cite{onavos, Chen_2018_CVPR, Hu_2018_ECCV, behlincremental}.

We adapt our model by simply updating the set of visual words that represent the object.
Concretely, given a dictionary of visual words $\mathcal{M}$, captured up to the frame $t_j$, we predict the segmentation map in frame $t_{j+\delta}$, and treat it as a new support set $ \mathcal{S^\delta} = \{ x_i^\delta, y_i^\delta \}_{i=1}^{N}$, where $y_i^\delta$ is the predicted object class for pixel $x_i^\delta$.
Next, we compute an updated set of deep visual words $\mathcal{M^\delta}$ from the new support set using k-means as described in Sec.~\ref{sec:unsupervised_learning_visual_words}, and compute their corresponding cluster centroid representations by
\begin{equation}
\mu_{ck}^\delta = \frac{1}{\lvert \mathcal{S}_{ck}^\delta \rvert}\sum_{x_i \in \mathcal{S}_{ck}^\delta} f_\theta(x_i).
\label{online_mu}
\end{equation}
To filter out incorrect predictions and prevent errors %
from compounding, we only add new words that still resemble the existing ones.
This is based on the assumption that within a time interval $\delta$, where $\delta$ is chosen moderately, the objects will deform slowly and their pixel-level embeddings will also not vary greatly.
Concretely, we update the main visual word set $\mathcal{M}$ with the new set $\mathcal{M}^\delta$, if there are $m^\delta \in \mathcal{M}_c^\delta$ and $m \in \mathcal{M}_c$, for which $\norm{\mu_m^{\delta} - \mu_m} \leq \alpha$.

Additionally, to ensure that online adaptation uses reliable and confident pixel-level predictions to update the visual words, we apply a simple outlier removal process (that assumes spatio-temporal consistency of objects over time) to the pixel-level predictions.
Specifically, we discard regions from the adaptation process if they have no intersection with the predicted object mask in the previous frame, using connected components.
Note that during online adaptation, none of the existing words within $\mathcal{M}$ are discarded, because each object may revert to its original shape, appearance or viewpoint during a video.
This is also why the Eq.~\ref{eq:label_prob} takes the maximum value to match to only the most relevant visual word.
The effect of this online adaptation procedure, and other design choices, are experimentally validated next.

\vspace{-0.3\baselineskip}
\section{Experimental evaluation}
\vspace{-0.1\baselineskip}
\subsection{Experimental setup}
\label{sec:experimental_setup}
\vspace{-0.1\baselineskip}
\paragraph{Model}
We use a Deeplab v2 architecture as the encoder \cite{deeplab}, $f(\theta)$, which uses a ResNet-101 \cite{resnet} backbone with dilated convolutions.
The encoder maps an input frame of size $H \times W$ to a feature of size $H \times W \times 2048$.
We add an additional convolutional layer to produce an embedding of $d=128$ dimensions, and bilinearly upsample this to the original image size.
These $128$-dimensional embeddings are then clustered to form our visual words.
Unless otherwise specified, we use $k = 50$ visual words for the foreground object classes.
As the background typically contains more variation, we use four times as many%
clusters for the background.
For online adaptation, we set $\alpha = 0.5$.
The ablation study in Sec.~\ref{sec:exp_ablation} shows the effect of the number of visual words, $k$.

\paragraph{Datasets}
We evaluate on standard video segmentation datasets for tracking both single objects (DAVIS-2016 \cite{DAVIS}, YouTube-Objects \cite{youtube1, youtube2}) and multiple objects (DAVIS-2017 \cite{davis_2017}, SegTrack v2 \cite{FliICCV2013}) given fully-annotated object masks in the first frame.
DAVIS-2016 contains 30 training and 20 validation videos.
DAVIS-2017 extends DAVIS-2016 to 60 training and 30 validation videos.
Furthermore, multiple objects (ranging from 1 to 5, with an average of 2) are annotated in the first frame and must be tracked through the video, making it considerably more challenging than DAVIS-2016.
YouTube-Objects and SegTrack v2 do not have a training split, so we evaluate our model trained on DAVIS-2017 on them.

\paragraph{Training}
Following competing methods which use a model pretrained on image segmentation datasets \cite{Hu_2018_ECCV, Oh_2018_CVPR, Chen_2018_CVPR, Yang_2018_CVPR, OSVOS-S, onavos}, we initialise the encoder of our network using the public Deeplab-v2 model \cite{deeplab} that has been trained on COCO \cite{coco}.
Thereafter, we meta-train our model following the ``episodic training'' procedure, which is the standard practice %
\cite{vinyals_2016, snell_2017, MAML, behl2019alpha}.
Each training episode is formed by sampling a support set $\mathcal{S}$ and a relevant query set $\mathcal{Q}$.
The idea of episodic training is to, at each iteration, mimic the inference procedure.
In other words, the query set should be classified given only the support set.
Here, we build each episode by first randomly sampling a video from the training dataset, treating the pixels of the first frame of the video as $\mathcal{S}$, and randomly selecting a set of query frames from the rest of the video and treating their pixels as $\mathcal{Q}$. Randomly selecting sets of frames in the video make the method robust to temporal discontinuties, occlusions and object complexity, thereby making it more generic. We sample from as low as 3 frames to the entire video length.

\paragraph{Evaluation metrics}

We report standard metrics defined by the DAVIS protocol \cite{DAVIS}: The mean IoU ($\mathcal{J}$), the F-score along the boundaries of the object ($\mathcal{F}$) and the mean of these two values ($\mathcal{J} \& \mathcal{F}$).
We also report the ``decay'' \cite{DAVIS} in $\mathcal{J}$.
This is calculated by splitting a video temporally into four clips, and taking the difference of the IoU in the last clip to the first clip.
Lower scores of ``decay'' are better, and was proposed by \cite{DAVIS} to measure whether a model is robust or if its predictions degrade over time.

Finally, we also report our runtime per-frame.
Our runtime is measured on a desktop machine with a single Titan X (Pascal) GPU, and an Intel i7-6850K CPU with six cores.
More details and experiments are available in the supplementary material\footnote{Supplementary materials \url{https://harkiratbehl.github.io/}}.

\vspace{-0.2\baselineskip}
\subsection{Comparison to state-of-art}
\label{sec:exp_comp_sota}

\begin{figure}
	\vspace{-0.75\baselineskip}
		\ffigbox{%
			\includegraphics[width=\textwidth]{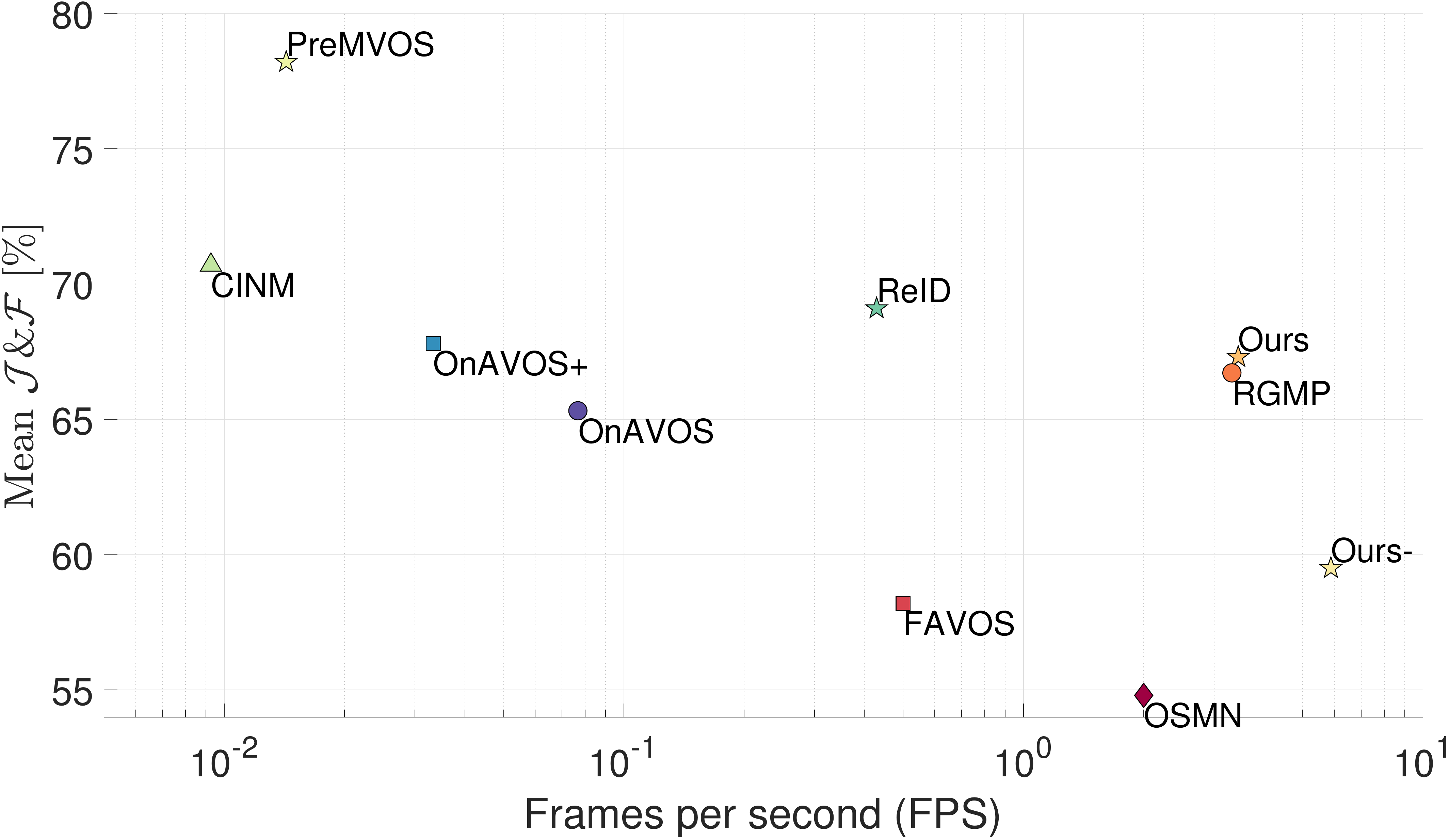}
		}{%
			\vspace{-1.5\baselineskip}
			\caption{{\bf Comparison of speed and accuracy on DAVIS 2017.} 
				Entries on the Pareto front (\ie no other method is both faster and more accurate) are marked by a star.
				Note the speed axis uses a logarithmic scale.}%
			\label{fig:pareto_davis_2017} 
		}
\vspace{-\baselineskip}
\end{figure}

\begin{table*}[t]
    \centering
    \small
   \vspace{0.2cm}
   \caption{{\bf State-of-art results among methods not performing finetuning on four common video object segmentation datasets.}
   		Legend: FT: Fine-Tuning on the first frame of the test video; PP: Post-Processing; OF: Optical Flow;
   		$\textrm{Ours}^-$: Our model without online adaptation; $\dagger$: An ensemble of models are used. *As the original authors did not report the runtime, we timed it using the public inference code.
   		Evaluation metrics are detailed in Sec.~\ref{sec:experimental_setup}. 
   	}
    \resizebox{\textwidth}{!}{
	\begin{tabular}[t]{lccc ccccc ccccc c c} 
	    \toprule
		 &  &  &  & \multicolumn{5}{c}{DAVIS-2017} & \multicolumn{5}{c}{DAVIS-2016} & YouTube-Objects & SegTrack-v2\\
		\cmidrule(lr){5-9} \cmidrule(lr){10-14} \cmidrule(lr){15-15} \cmidrule(r){16-16}
		Method & FT & PP & OF &  ${\mathcal{J}}\&\mathcal{F}${\footnotesize (\%)} & ${\mathcal{J}}${\footnotesize (\%)} & ${\mathcal{J}}$ Decay{\footnotesize (\%)}& ${\mathcal{F}}${\footnotesize (\%)}  & Time(s)  & ${\mathcal{J}}\&\mathcal{F}${\footnotesize (\%)} & ${\mathcal{J}}${\footnotesize (\%)} & ${\mathcal{J}}$ Decay{\footnotesize (\%)} & ${\mathcal{F}}${\footnotesize (\%)} &  Time(s) & ${\mathcal{J}}${\footnotesize (\%)} & ${\mathcal{J}}${\footnotesize (\%)}\\
		\cmidrule(r){1-4} \cmidrule(lr){5-9} \cmidrule(lr){10-14} \cmidrule(lr){15-15} \cmidrule(r){16-16}
        OnAVOS~\cite{onavos} &\cmark &\cmark & & 65.3 & 61.6 & 27.9  & 69.1 & 13  & 85.5 & 86.1 & 5.2  & 84.9 & 13 & 77.4 & --\\
		$\textrm{OSVOS}^S$~\cite{OSVOS}  &\cmark &\cmark & & {68.0} &{64.7} & 15.1  & {71.3} &  --   & {86.5} & {\bf 85.6} & 5.5  & { 87.5} & 4.50 &{\bf 83.2} & 65.4\\	
		$\textrm{PReMVOS}^{\dagger}$~\cite{DAVIS2018-Semi-Supervised-1st}  &\cmark & &\cmark & {\bf 78.2} & {\bf 74.3} & 16.2  & {\bf 82.2} & $\sim$ 70    & {\bf 87.0} & { 85.5} & 8.8 & {\bf 88.6} & $\sim$ 70 & -- & --\\
		\cmidrule(r){1-4} \cmidrule(lr){5-9} \cmidrule(lr){10-14} \cmidrule(lr){15-15} \cmidrule(r){16-16}
		MaskRNN~\cite{hu2017maskrnn} & & &\cmark & -- &45.5 & --  & -- &  0.60 & --  & -- & -- & -- & -- & -- & --\\
		FAVOS~\cite{Cheng_2018_CVPR} & &\cmark & & 58.2 & 54.6 & {\bf14.1}  & 61.8 &  $>$1.80 & 80.9 & {\bf 82.4} & \textbf{4.5}  & 79.5 & 1.80  & -- & --\\ 
		CTN~\cite{Jang_2017_CVPR} & & &\cmark  & -- & --  & -- & -- & -- & 71.4 & 73.5  & 15.6 & 69.3 & 1.33 & -- & --\\  
		FAVOS~\cite{Cheng_2018_CVPR} & & &  & -- & --  & -- & -- & -- & 76.9 & 77.9  & -- & 76.0 & 0.60 & -- & --\\ 
		VPN~\cite{Jampani_2017_CVPR} & & &  & -- & --  & -- & -- & -- & 67.8 & 70.2  & 12.4 & 65.5 & 0.63 & -- & --\\ 
		BVS~\cite{bvs} & & &  & -- & --  & -- & -- & -- & 59.4 & 60.0  & 28.9 & 58.8 & 0.37 & 68.0 & 60.0\\
		OSMN~\cite{Yang_2018_CVPR} & & &  & 54.8 & 52.5 & 21.5  & 57.5 &  0.50* & -- & 74.0 & 9.0 & -- & 0.14  & 69.0 & --\\
		VideoMatch~\cite{Hu_2018_ECCV} & & & & -- & 56.5 & --  & -- & 0.35 & -- & 81.0 & --  & -- & 0.32 & 79.7 & --\\
		RGMP~\cite{Oh_2018_CVPR} & & & & 66.7 & {\bf 64.8} & 18.9  & 68.6 & 0.30* & 81.7 & {81.5}  & 10.9 & 82.0 &  {\bf0.13} & -- & {71.1}\\
	    $\textrm{Ours}^-$ & & & &{63.1} & { 59.5}  & {55.8} & 24.6 & {0.17} & 76.9 & 76.2  & 11.2 & 77.6 & 0.17 & 77.4 & 64.6 \\
	    Ours & & & & {\bf 67.3}  & {63.9} & {14.4} &{\bf 70.7} &{\bf 0.29} & {\bf 82.1} & 81.5  & 5.0 & {\bf 82.7} & 0.25 & {\bf 81.1} & {\bf 72.0}\\
	    \bottomrule
	\end{tabular}
	}
	\label{tab:comparison_all}
	
\end{table*}
Table \ref{tab:comparison_all} shows our state-of-art results on DAVIS-2017, DAVIS-2016, YouTube-Objects and SegTrack-v2.
On all these datasets, there is no method that is both faster and more accurate than us.
This Pareto front is also visualised in Fig.~\ref{fig:pareto_davis_2017} for DAVIS-2017, the most challenging dataset.
The methods that are more accurate than us all finetune on the first frame, use optical flow or additional post-processing such as CRFs \cite{densecrf} and thus have a runtime that is larger by a factor of at least $8$ \cite{Li_2018_ECCV, OSVOS}. 

The only method that is close to us in terms of speed and accuracy is RGMP \cite{Oh_2018_CVPR}.
However, the runtime of RGMP increases linearly with the number of objects being tracked from the first frame. 
This is because RGMP processes each object instance independently through the ``encoder'' part of their network \cite{Oh_2018_CVPR}, and combine their results together at the end.
Therefore, even though RGMP is faster than our method on DAVIS-2016 (a single-object dataset), it is actually slower on DAVIS-2017 (a multi-object dataset).
As the authors did not report the runtime of their method on DAVIS-2017, we ran their publicly available inference code, and obtain an average runtime of 0.30s per frame on DAVIS-2017 (Tab.~\ref{tab:comparison_all}).
However, DAVIS-2017 only averages 2 objects per video.
We measured RGMP to average 0.11s, 0.41s and 0.60s per frame for videos with 1, 3 and 5 objects respectively.
The runtime of our method, in contrast, increases much slower, taking 0.25s, 0.38s and 0.53s respectively.  %
Thus, the runtime of RGMP increases by $5.4\times$ from 1 object to 5 objects, whilst our runtime only increases by $2.1\times$.
This is because our model only requires a single forward-pass through the network, irrespective of the number of objects being tracked.
Thus, there is only a minor increase in runtime from DAVIS-2016 to DAVIS-2017 as there are more visual words.
Note that the runtime advantage of our method would increase over RGMP \cite{Oh_2018_CVPR} if even more objects were to be tracked in a video.
Moreover, our speed could also be improved by implementing CUDA kernels for k-means clustering.

Note that our method also achieves a lower $\mathcal{J}$ Decay \cite{DAVIS} than RGMP \cite{Oh_2018_CVPR} indicating greater robustness.
This is also shown qualitatively in Fig.~\ref{fig:comparison_qualitative_main} where our method overcomes occlusions and can recover from errors made in previous frames, unlike mask propagation methods like RGMP.
We believe that representing objects with cluster centroids in the embedding space (visual words), which correspond to object parts in the image space, increases the robustness of our matching,
as the appearance of more local parts typically stays consistent even though the object as a whole transforms.
And as our online adapatation process retains a memory of previous visual words, our method can handle objects disappearing and reappearing (Fig.~\ref{fig:comparison_qualitative_main}) unlike RGMP.

\def \imwidth {0.18}
\def \imopts {valign=m}

\setlength{\tabcolsep}{1pt}

\begin{figure*}[]
\vspace{0.2cm}
	\begin{tabular}{m{1cm}ccccc}
		
	\footnotesize{Ours}	&   \includegraphics[valign=m,width=\imwidth\textwidth]{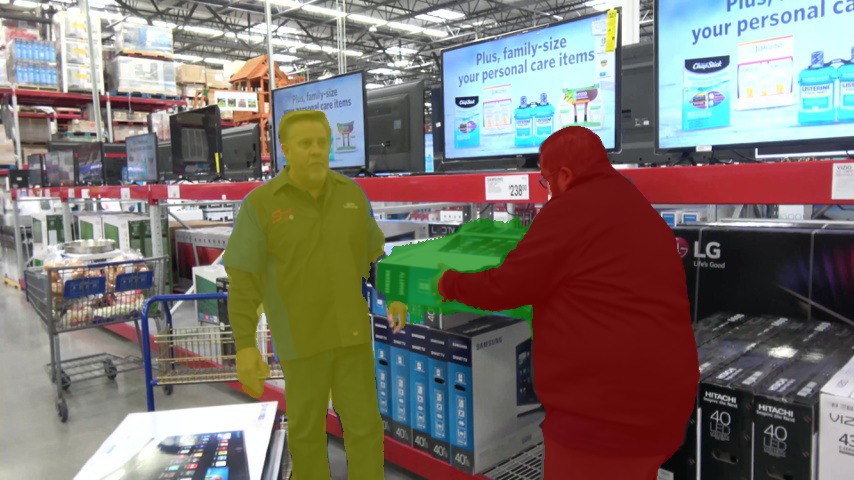}
		& 
	\includegraphics[valign=m,width=\imwidth\textwidth]{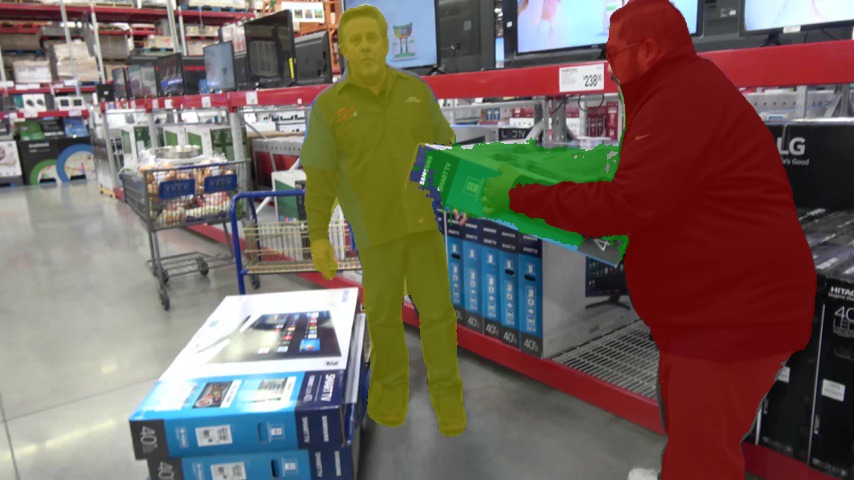}
	 & 
 	\includegraphics[valign=m,width=\imwidth\textwidth]{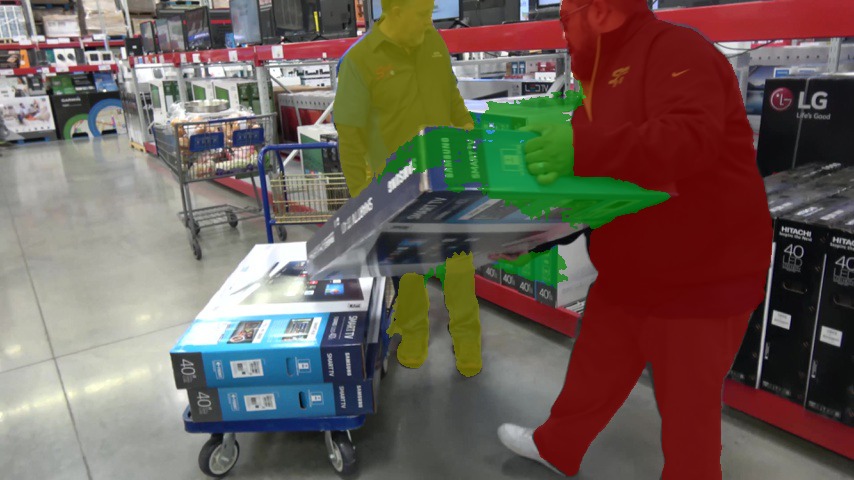}
	&
 	\includegraphics[valign=m,width=\imwidth\textwidth]{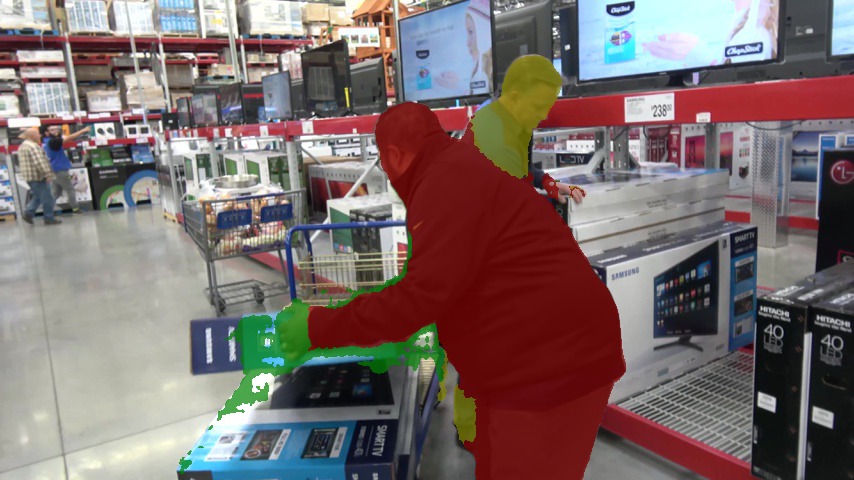}
	&
 	\includegraphics[valign=m,width=\imwidth\textwidth]{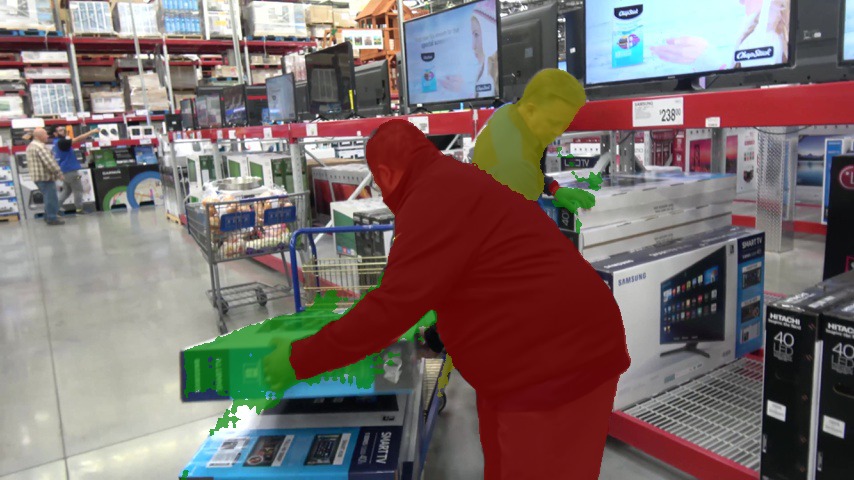}
	\\ [0.8cm]
	\footnotesize{RGMP \cite{Oh_2018_CVPR}}	& 
	\includegraphics[valign=m,width=\imwidth\textwidth]{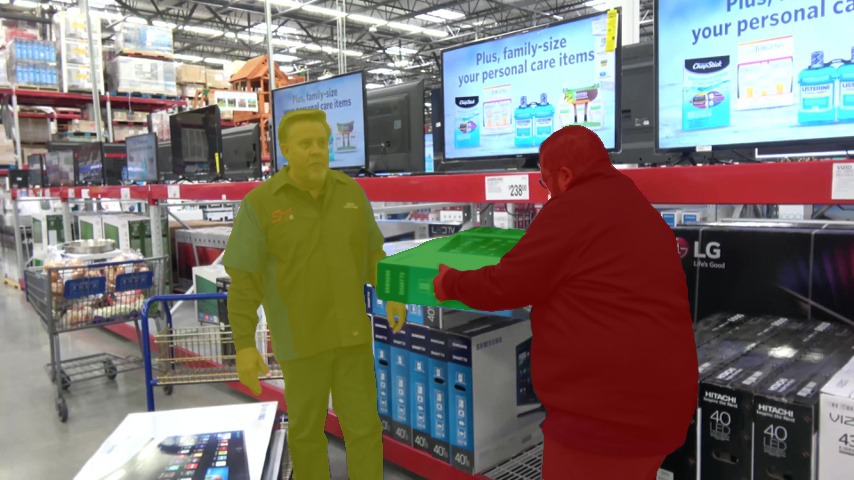}
	&
	\includegraphics[valign=m,width=\imwidth\textwidth]{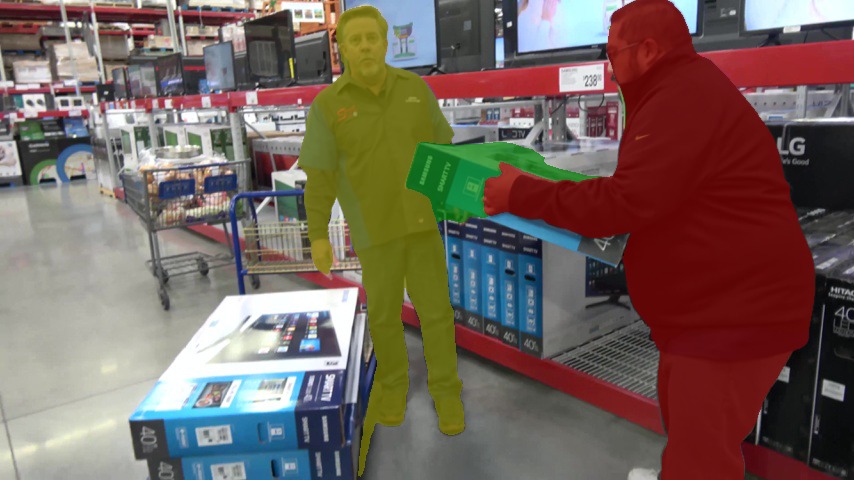}
	&
	\includegraphics[valign=m,width=\imwidth\textwidth]{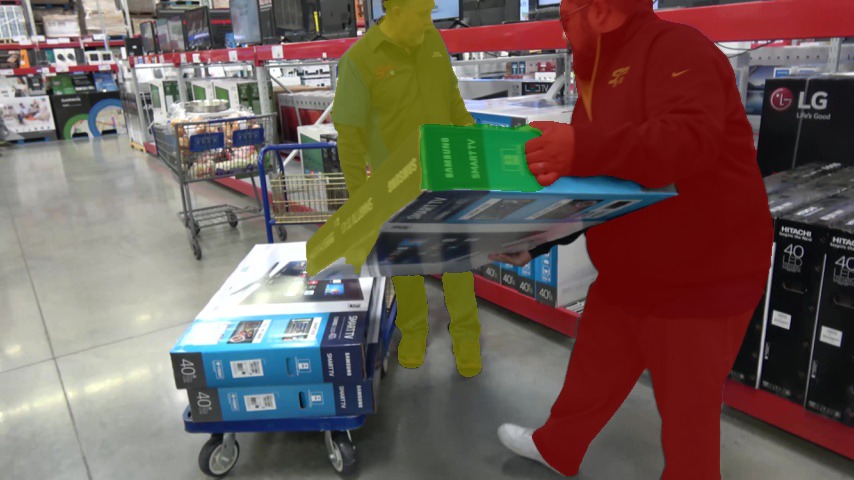}
	&
	\includegraphics[valign=m,width=\imwidth\textwidth]{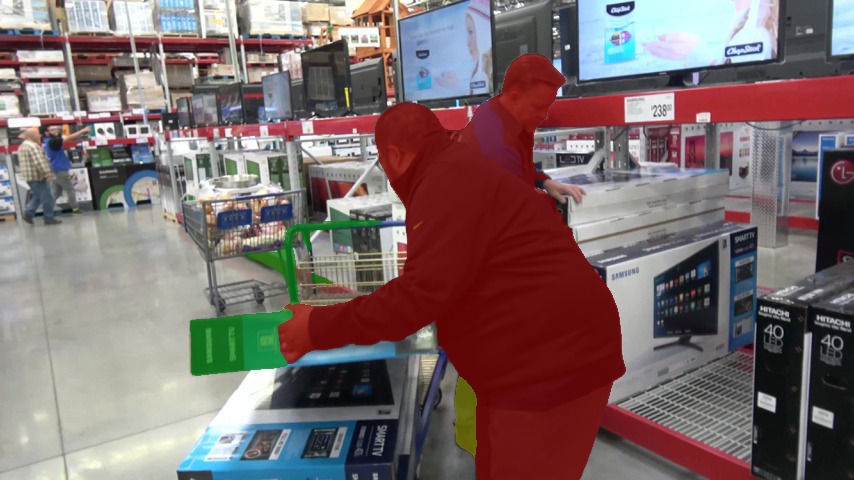}
	&
	\includegraphics[valign=m,width=\imwidth\textwidth]{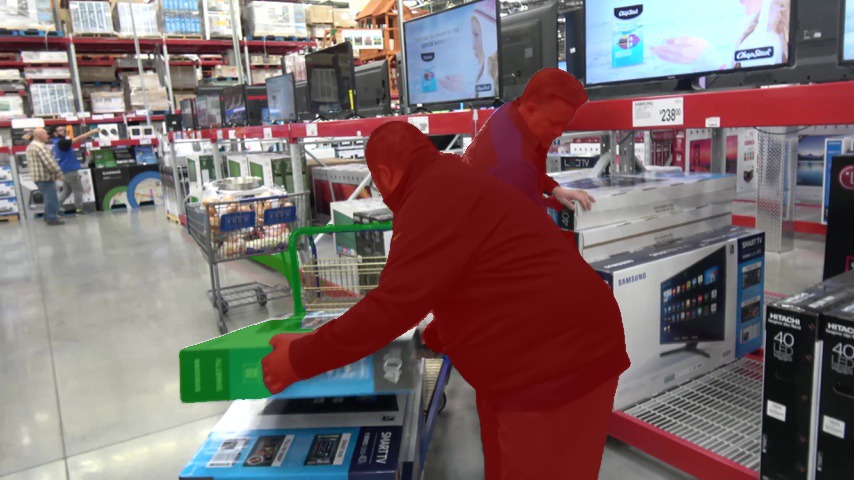}\\ [0.85cm]
	
	\footnotesize{Ours}	& 
		\includegraphics[valign=m,width=\imwidth\textwidth]{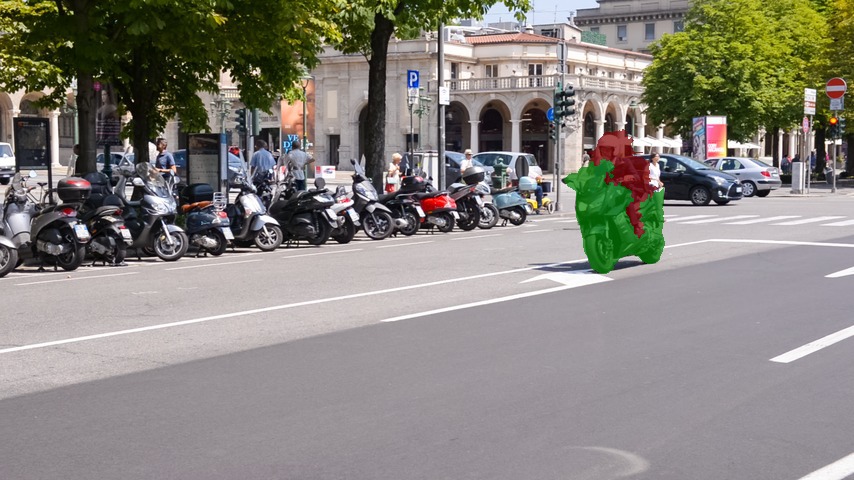}
		&
		\includegraphics[valign=m,width=\imwidth\textwidth]{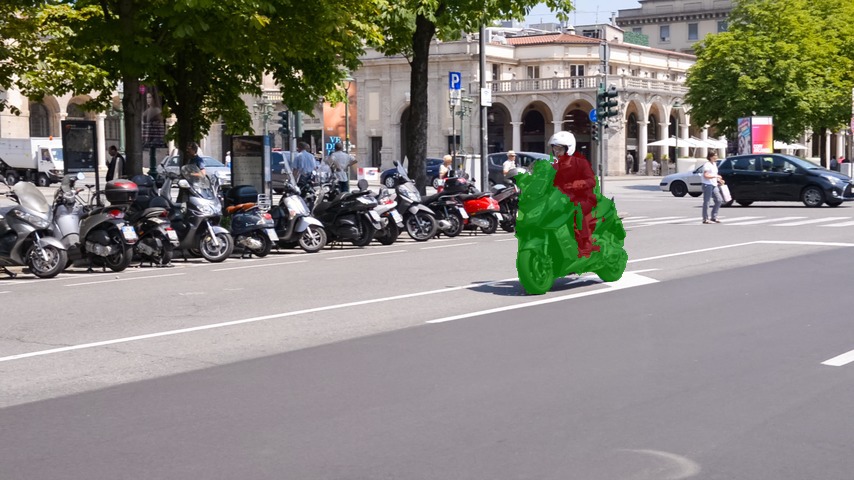}
		&
		\includegraphics[valign=m,width=\imwidth\textwidth]{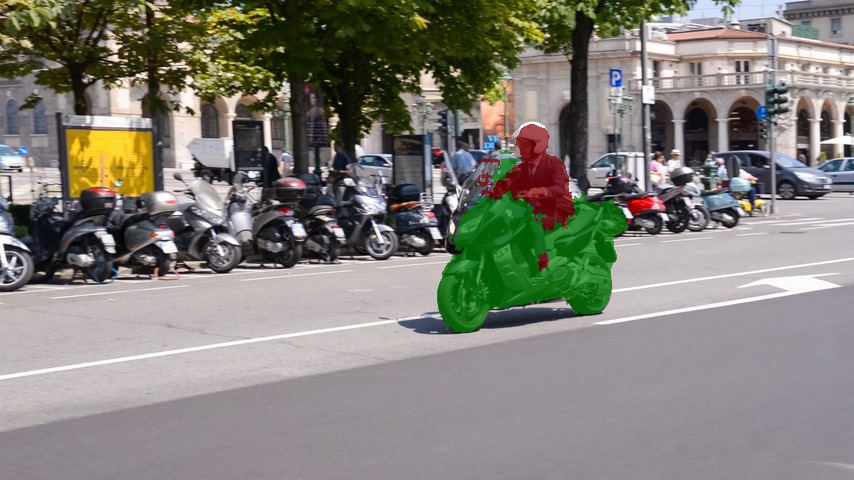}
		&
		\includegraphics[valign=m,width=\imwidth\textwidth]{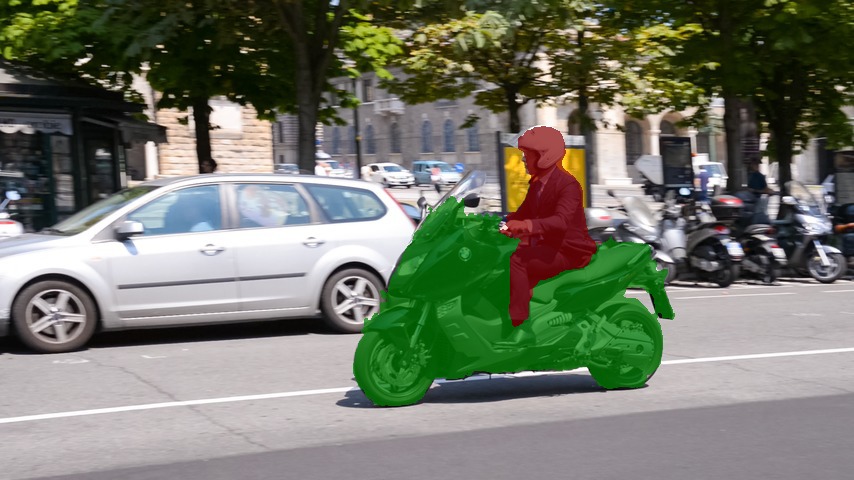}
		&
		\includegraphics[valign=m,width=\imwidth\textwidth]{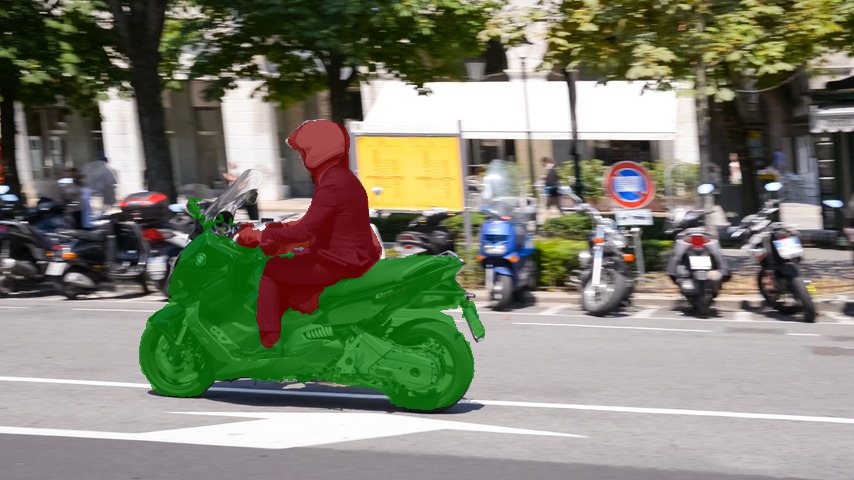} \\ [0.8cm]
		
	\footnotesize{RGMP \cite{Oh_2018_CVPR}}	&
		\includegraphics[valign=m,width=\imwidth\textwidth]{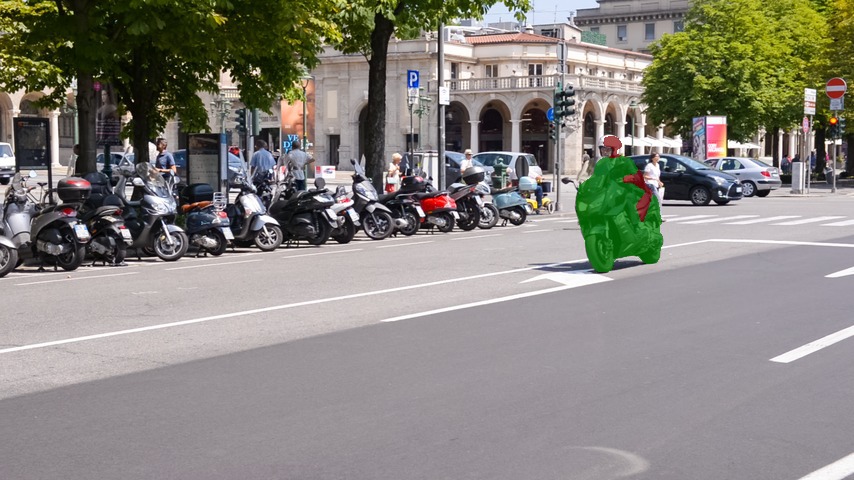}
		&
		\includegraphics[valign=m,width=\imwidth\textwidth]{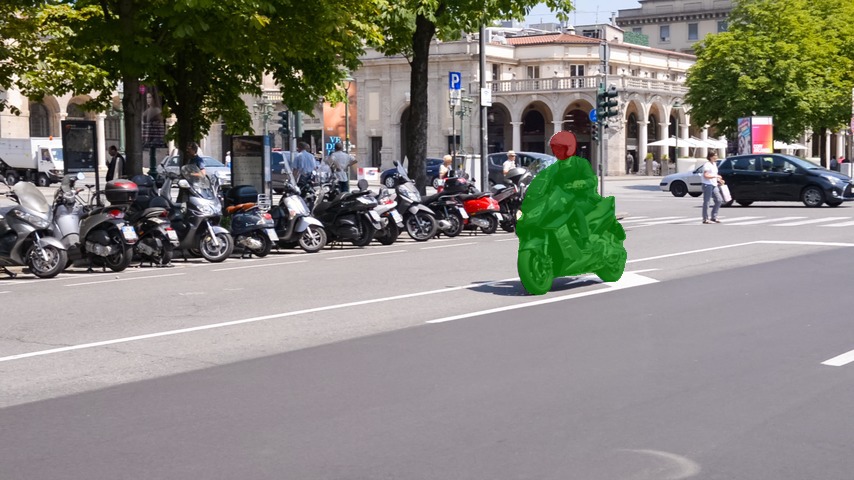}
		&
		\includegraphics[valign=m,width=\imwidth\textwidth]{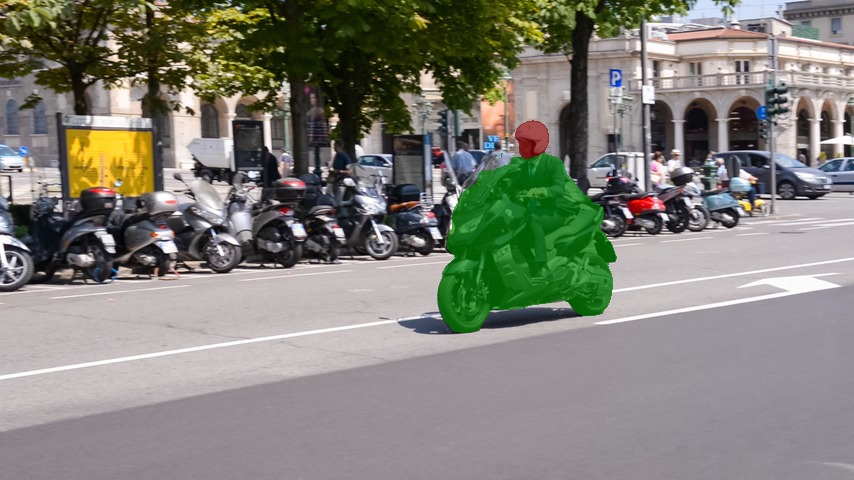}
		&
		\includegraphics[valign=m,width=\imwidth\textwidth]{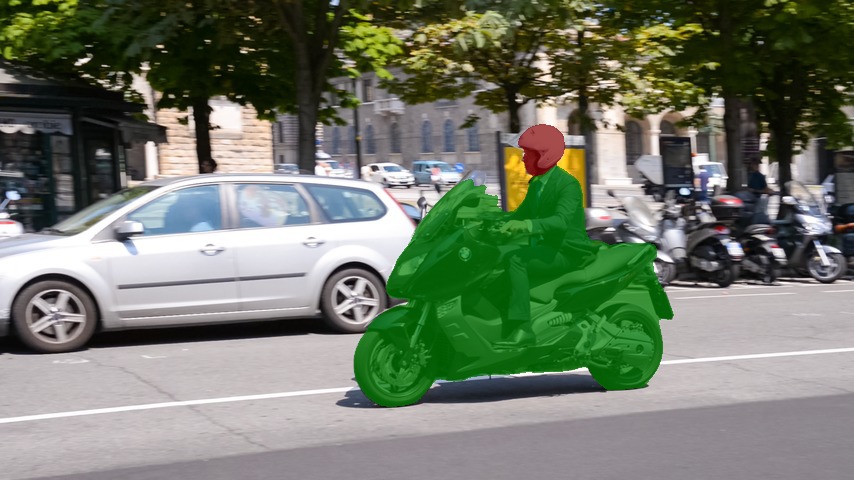}
		&
		\includegraphics[valign=m,width=\imwidth\textwidth]{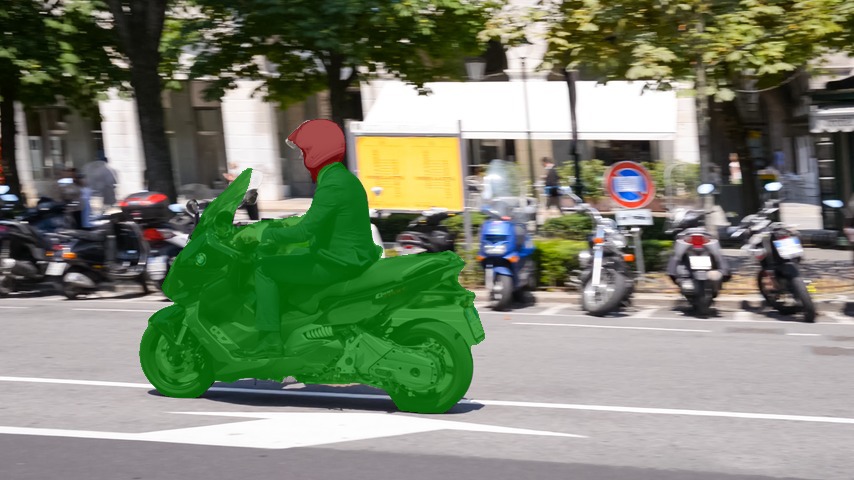} \\ [0.85cm]
	
	\end{tabular}
	\caption{\textbf{Qualitative comparison of our method to RGMP \cite{Oh_2018_CVPR}.} RGMP obtains good results initially in the video (first two columns), but cannot recover after making errors (third column). 
	Note how it misclassifies the yellow person (first example) and loses track of the rider (second example).
	In contrast, our method overcomes occlusions in all of these cases by robustly matching an object to its constituent parts.
	} 
	\label{fig:comparison_qualitative_main}
\end{figure*}

\vspace{-0.23cm}
\subsection{Bounding box based initial mask}
In this section, we do experiments when only bounding box supervision is available in the first frame. Hence the aim is to segment the object in the video, given only the bounding box in the first frame. 
To generate the dictionary of visual words for a given object, we use a mechanism similar to our online adaptation.
We first take the embeddings of all pixels in the bounding box of the object and segment them into clusters. We then discard the ones that resemble the background, with the difference that here we use resemblance with the background to eliminate some regions of the bounding box.
The remaining clusters are then used to construct the visual word dictionary for this object. 
The rest of the algorithm remains the same.
The results are shown in Table \ref{tab:bounding_box}. It can be seen that there is not a substantial drop in performance in comparison to the mask based case. 
We believe that our part-based approach makes our model more robust to the noise in the input in this scenario.
\begin{table}[t]
    \centering
    \small
   \caption{{\bf Results of our method ($\mathcal{J\&F}$) with only bounding box based initialization.}
   		$\textrm{Ours-BB}$: Our model with only bounding box mask provided in first frame.
   	}
	\begin{tabular}[t]{l c c c c} 
	    \toprule
		 & DAVIS-2017 & DAVIS-2016 & YouTube-Objects & SegTrack-v2\\ \midrule
	    Ours & { 63.8} & { 81.5} & { 81.1} & { 72.0}\\
	    Ours-BB & { 51.5} & { 77.5} & { 75.8} & { 66.5}\\
	    \bottomrule
	\end{tabular}
	\label{tab:bounding_box}
	\vspace{-0.5\baselineskip}
\end{table}

\vspace{-0.2\baselineskip}
\subsection{Ablation study}
\label{sec:exp_ablation}
\vspace{-0.1\baselineskip}
This section studies how different design choices in our algorithm impact overall performance on DAVIS-2017. %

\paragraph{Effect of Meta-Learning}
To evaluate the efficacy of meta-learning, we evaluated our MS-COCO initialised network and obtained a mean IoU of ($\mathcal{J}$) of $50.7$. 
Meta-training significantly improves our IoU to $63.9\%$ (Tab.~\ref{tab:comparison_all}).
Perhaps surprisingly, our initialisation already outperforms published work like Mask-RNN \cite{hu2017maskrnn} (Tab.~\ref{tab:comparison_all}).

\paragraph{Object representation}

\begin{table}[]
	\centering
	\small
	\vspace{-0.1\baselineskip}
	\begin{tabular}[t!]{ccc}
		\toprule
		Model                  & $\mathcal{J}${\footnotesize (\%)} & Time(s)\\ \midrule
		Single prototype  & 32.9 & \textbf{0.14} \\
		5 Nearest neighbours      & 45.9 & 5.50\\
		Visual words ($k = 50$)    & \textbf{48.4} & 0.17\\
		\bottomrule
	\end{tabular}
	\caption{{\bf The effect of different object representations} 
	The same MS-COCO pretrained network is used, without any online adaptation.
	We use the 5 nearest neighbours, following \cite{Chen_2018_CVPR}.}
	\label{tab:representation_comparison}
	\vspace{-0.5\baselineskip}
\end{table}
\squeezeup
We represent the object given in the first frame with a dictionary of $k$ visual words in the embedding space.
An alternative is to represent each object with a single vector, \ie $k = 1$ (as in Prototypical networks \cite{snell_2017}). 
In our case, this prototype is formed by taking the mean embedding of all pixels of the object labelled in the first frame.
The other end of the spectrum is to represent each object with separate embeddings for all of its pixels, \ie $k = n$ where $n$ is the number of labelled pixels in the first frame, like \cite{Chen_2018_CVPR} and Matching networks \cite{vinyals_2016}.

Table \ref{tab:representation_comparison} compares these approaches for our MS-COCO pretrained network.
It shows that using $k = 50$ clusters outperforms both nearest neighbour classification and a single prototypical vector per class.
This motivates our reason for using $k$ visual words to represent an object and suggests why we outperform methods such as \cite{Chen_2018_CVPR} in Tab.~\ref{tab:comparison_all}.

Note how matching using our visual words representation has a similar runtime to a single prototype and is significantly faster than performing a nearest neighbour search.
This is because the search time is linear in the number of pixels, $O(n)$. 
And like the other approaches we compare to in Tab.~\ref{tab:representation_comparison}, we do the matching at full resolution. %
The runtime could be greatly reduced by doing the look-up on a subsampled image (for example, \cite{Chen_2018_CVPR} do the look-up at $1/8$ resolution which reduce the time by about a factor of $64$).

\paragraph{Number of visual words}
\begin{table}
\vspace*{-\baselineskip}
\resizebox{\textwidth}{!}
{\begin{tabular}{cccccccccc}
		\toprule
		Dictionary Size ($k$) & 1 & 5 & 10 & 50 & 100 & 200 & 400 & 1500 & 3000\\ \midrule
		$\mathcal{J}${\footnotesize (\%)} & 49.9    & 54.5 & 54.8 & 55.8 & 56.3 & 56.3 & 56.4 & 56.2 & 54.5\\
		Time (s)                          & 0.140 & 0.168 & 0.170  & 0.173 & 0.199 & 0.254 & 0.373 & 0.97 & 1.42\\
		\bottomrule
	\end{tabular}
}
\caption{{\bf The effect of the size of visual word dictionary on model performance}
				Results are on DAVIS-2017, without any online adaptation.
			}
			\label{tab:ablation_num_clusters}
\vspace{-0.5\baselineskip}
\end{table}

Table \ref{tab:ablation_num_clusters} examines the effect that the number of visual words in the dictionary has on accuracy and runtime on DAVIS-2017.
We can see that accuracy steadily increases as the number of clusters is increased from $k = 1$ (which corresponds to Prototypical networks \cite{snell_2017}) and plateaus at $k = 50$.
We believe that complex objects with high intra-object variations produce embeddings with multi-modal distributions, which is why they are better represented with multiple visual words.
Although increasing $k$ beyond 50 does not substantially change the accuracy, it does increase the runtime, which is why we use $k = 50$ when comparing to existing methods in Tab.~\ref{tab:comparison_all}. 
Setting the number of visual words to $n$, the number of pixels in the first frame, would amount to the nearest neighbour search done by \cite{Chen_2018_CVPR}.%

\section{Conclusion and Future Work}
\vspace{-0.2cm}

We proposed a novel representation of objects by their cluster centroids in the embedding space (visual words) which correspond to object parts.
These visual words were learned without supervision, using meta-learning.
Visual words enable robust matching, as the appearance of local parts may stay consistent whilst the object as a whole deforms or is occluded.
Our novel representation, and meta-training procedure enabled our method to achieve state-of-art performance on four common datasets in terms of speed and accuracy trade-offs (with comparable accuracy to expensive finetuning-based methods that take at least $8$ times longer).
Moreover, our method readily scales to multiple objects in videos, with its runtime only increasing slightly from single-object DAVIS-2016 to multi-object DAVIS-2017.
Finally, the robustness of our part-based algorithm allows us to easily extend it to the scenario where we only have bounding-box supervision in first frame.
Future work is to learn the number of clusters automatically, and learn to generate synthetic data \cite{behl-2020-autosimulate} for video segmentation.

\textbf{Acknowledgements}
Harkirat is wholly funded by a Tencent grant. This work was supported by the Royal Academy of Engineering, EPSRC/MURI grant EP/N019474/1 and FiveAI.

\vspace{-0.75\baselineskip}
\bibliographystyle{IEEEtran}
\bibliography{egbib}

\section*{Appendix}

In this supplementary material, we present additional ablation studies (Sec.~\ref{sec:supp_ablation}), more qualitative results (Sec.~\ref{sec:supp_qualitative_res}) and also further quantitative results (Sec.~\ref{sec:supp_quantitative}).

\section{Additional ablation studies}
\label{sec:supp_ablation}

In this section, we present an additional experiment on the temporal consistency of our visual words, the effect of online adaptation and the size of the visual word dictionary.

\subsection{Temporal consistency of our visual words}
In this experiment we aim to infer whether our visual words remain in the same region/part on the object throughout the video. 
We use the Physical Parts Discovery Dataset \cite{del2016discovering} for this experiment. 
This dataset contains videos for two object classes, namely, ``tiger'' and ``horse''. 
There are 16 videos for each class, with 8 videos with the animal facing right and the other 8 videos with the animal facing left. 
Ground truth annotation for 10 body parts of the object is provided for a subset of frames in the video. 

\def \imwidth {0.18}
\def \imopts {valign=m}

\setlength{\tabcolsep}{1pt}

\begin{figure*}[t]
	\vspace{-0.5\baselineskip}
	\begin{tabular}{m{1.3cm}cccccc}
		
	\begin{tabular}[c]{m{1cm}}
		Original Image \\ [0.5cm]
		Ground Truth Part Labels \\ [0.5cm]
		Visual words\\ [0.7cm]
		Predict-ion 
	\end{tabular}
	&  
	\includegraphics[valign=m,width=\imwidth\textwidth]{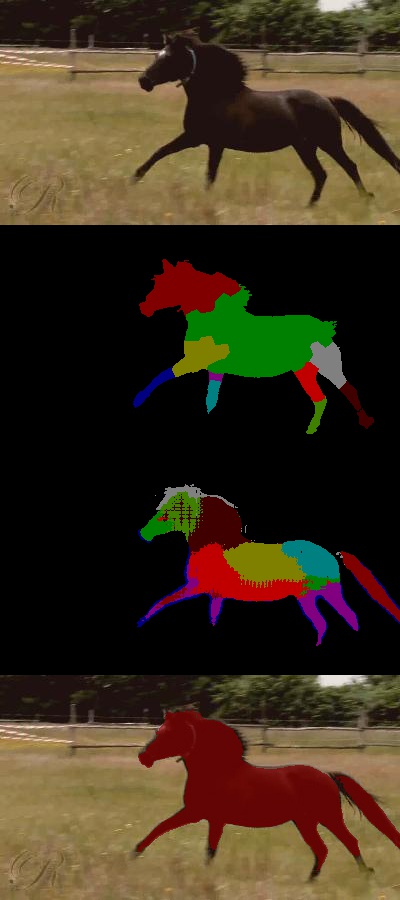}
	 & 
 	\includegraphics[valign=m,width=\imwidth\textwidth]{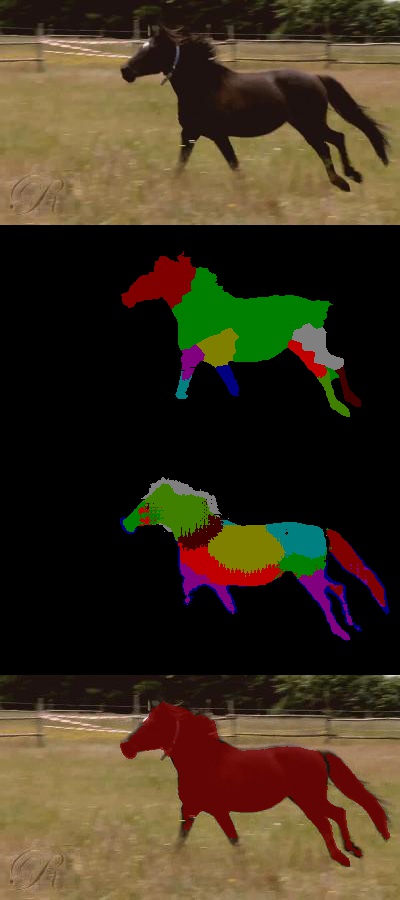}
	&
 	\includegraphics[valign=m,width=\imwidth\textwidth]{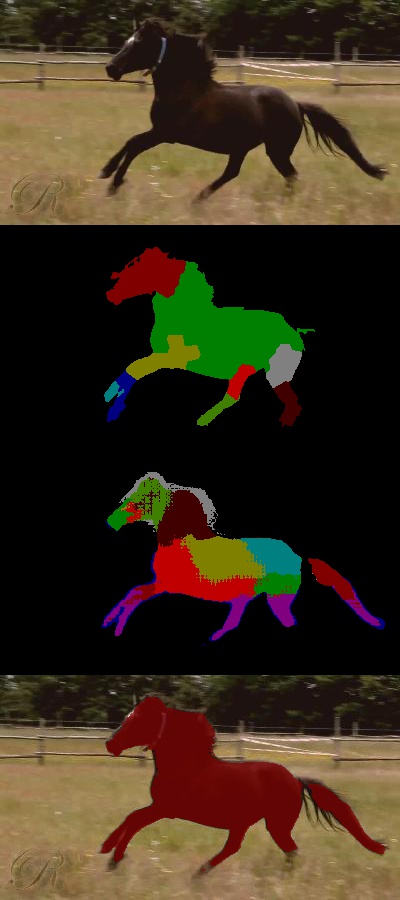}
	&
 	\includegraphics[valign=m,width=\imwidth\textwidth]{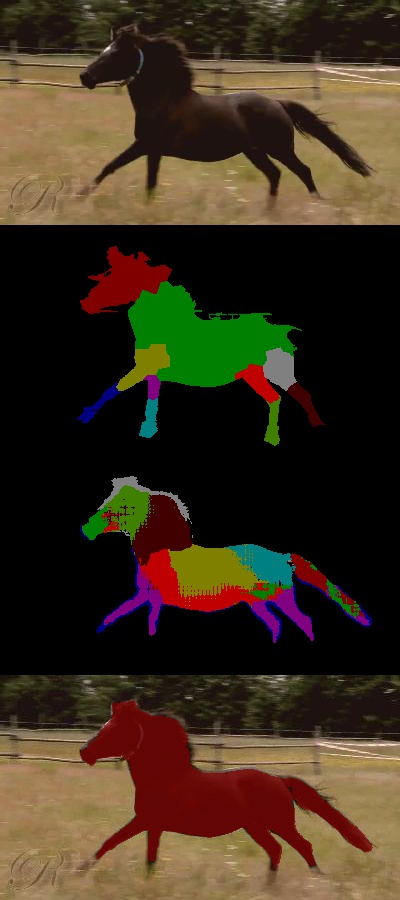}
 	&
 	\includegraphics[valign=m,width=\imwidth\textwidth]{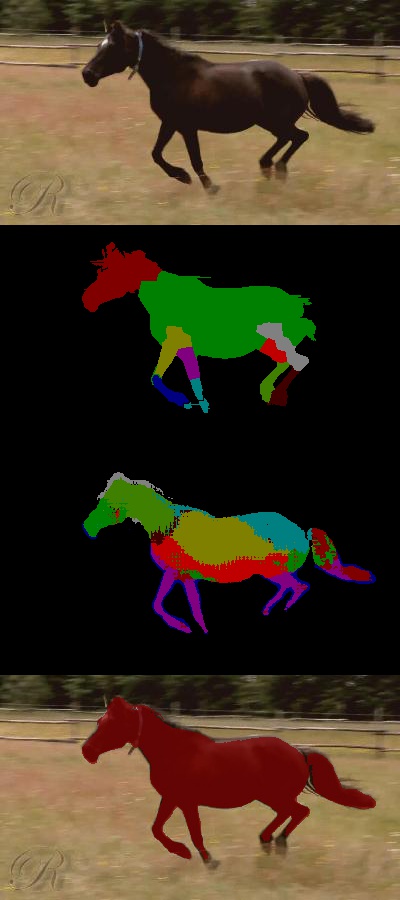}
	\\ [0.8cm]
	
	\end{tabular}
	\vspace{-0.5\baselineskip}
	\caption{\textbf{Qualitative examples of our experiment on the temporal consistency of visual words on the Physical Parts Discovery (PPD) Dataset \cite{del2016discovering}}
	Note how the visual words that our model learns in an unsupervised manner (third row) are consistent over time.
	The PPD dataset contains ground-truth annotations for animal classes (second row) which we leverage for our experiment.
	Note that our visual words will not correspond exactly to the ground truth parts as we learn our visual words in an unsupervised manner.
	}
	\label{fig:parts_experiment}
\end{figure*}

\setlength{\tabcolsep}{6pt}

We use this dataset, as it is the only one that we are aware of that annotates the parts of an object throughout the video.
However, as our method produces object parts in an unsupervised manner, we first form a mapping from each of our visual words to a ground truth object part.
Thereafter, we evaluate the consistency of these visual-word-to-object-part mappings over time. 

\paragraph{Mapping visual words to ground truth object parts}

We initially run our algorithm on the first frame of the video, to obtain $k$ visual words for the object.
We then map each visual word to a ground truth object part.
This assignment is done based on the majority vote from all the pixels in that visual word, \ie a visual word is assigned to the part which it has the maximum spatial overlap with it.
This is then the ground truth assignment of the visual words to the object parts and will be used next for evaluation.

\paragraph{Evaluation}

For all other frames of the video, we run our method and match each pixel to a visual word of the object.
We then find the part that these visual words now belong to, which is following the previous paragraph performed according to the maximum spatial overlap.
We then calculate how many visual words still belong to the part that they were assigned to in the first annotated frame.
The percentage of these consistent mappings, averaged over the entire dataset, is our performance metric, which we denote the ``Part consistency score''.
This metric measures the consistency of our visual words over time, as it indicates if a visual word is still in the same region of the object as it was in the first frame.

\paragraph{Results}

Table \ref{tab:part_consistency} performs this experiment for different numbers of visual words, $k$.
For $k = 10$, the accuracy is 77.4\%, showing that our method is indeed consistently tracking parts of the object.
This is also shown visually in Fig.~\ref{fig:parts_experiment} and in the supplementary video.
Table \ref{tab:part_consistency} also shows that the part tracking accuracy decreases as $k$ is increased.
We believe this is because each of the regions corresponding to a visual word decreases as $k$ is increased, and thus there is higher variance in visual-word-to-object-part mapping.
\begin{figure}[t]
	\centering
	\includegraphics[width =0.48\textwidth]{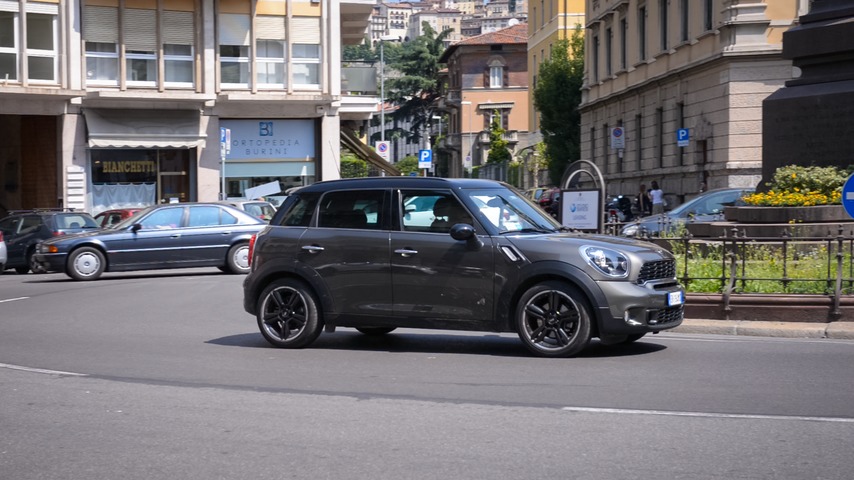}
	\includegraphics[width=0.48\textwidth]{images/online/00027.jpg}
	
	\includegraphics[width=0.48\textwidth]{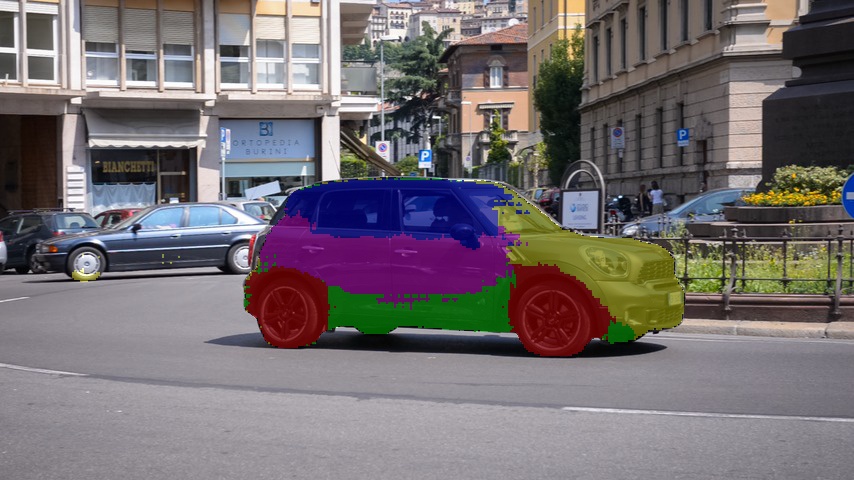}
	\includegraphics[width=0.48\textwidth]{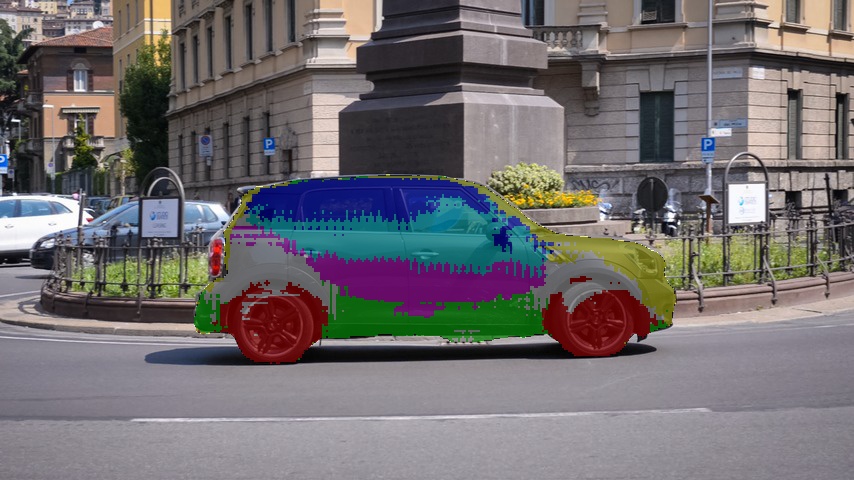}
	\caption{{\bf The effect of online adaptation on the representation of dynamic objects.} As the object pose changes over time, newer visual words (denoted by gray and light-blue colors) are learned and added to the dictionary of visual words.
	}
	\label{fig:online_adaptation_vis}   
	\vspace{-0.1in}
\end{figure}

\begin{table}[]
    \centering
    \small
	\caption{\textbf{Part consistency score as a function of the number of visual words, $k$}.
	The relatively high score (the maximum is 100) suggests that our method indeed forms visual words that consistently correspond to the same parts of the tracked object throughout the video.
	This is performed on the Physical Parts Discovery Dataset \cite{del2016discovering}.} 
	\resizebox{1\columnwidth}{!}
    {\begin{tabular}{cccccccccc}
    \toprule
    Dictionary Size ($k$) & 10 & 15 & 20 & 30 & 40 & 50 & 60 & 80 & 100\\ \midrule
    Part consistency score {\footnotesize (\%)} & 77.4 & 75.1 & 74.1 & 71.9 & 71.1 & 71.1 & 70.9 & 69.9 & 69.6\\
    \bottomrule
    \label{tab:part_consistency}
    \end{tabular}
    \vspace{-0.3in}
    }
\end{table}

\subsection{Effect of Online Adaptation}

Online adapation, as described in Sec.~\ref{sec:online_adaptation}, updates the dictionary of visual words to account for appearance changes in the object throughout the video. Figure~\ref{fig:online_adaptation_vis} shows an example of the online adaptation of our model on the DAVIS-2017 dataset.

\begin{table}[]
    \centering
    \resizebox{0.9\textwidth}{!}{
    \begin{tabular}{cccccccc}
    \toprule
    Interval ($\delta$) & NA & 30 & 20 & 10 & 5 & 2 & 1  \\
    \midrule
    $\mathcal{J}${\footnotesize (\%)} & 55.8 & 58.6 & 59.2 & 61.3 & \textbf{63.9} & 63.7 & 62.8\\
    \bottomrule
    \caption{{\bf The effect of online adaptation on model performance.} 
		$\delta$ denotes the frame interval before every step of online adaptation. Small values of $\delta$ help the model to adapt to quickly changing scenes, but too small a $\delta$ can introduce noise into the visual words. NA: No online adaptation.}
    \label{tab:ablation_online_adaptation}
    \end{tabular}
    \vspace{-0.5in}
    }
\end{table}
\begin{table}[]
	\small
    \centering
    \caption{{\bf The effect of outlier removal and online adaptation}
	}
	\label{tab:ablation_outlier_removal}
	\resizebox{0.8\textwidth}{!}{
	\begin{tabular}{ccc} 
	    \toprule
		Outlier Removal & Online Adaptation & ${\mathcal{J}}${\footnotesize (\%)} \\ \midrule
		\xmark &\xmark & 55.8  \\
		\xmark &\cmark & 60.4  \\ 
		\cmark &\cmark & 63.9  \\ 
	    \bottomrule
	\end{tabular}
	\vspace{-0.5in}
}
\end{table}

Table \ref{tab:ablation_online_adaptation} shows how updating the dictionary, $\mathcal{M}$, improves performance.
Smaller values of the update interval, $\delta$, means $\mathcal{M}$ is updated more frequently and thus helps the system smoothly adapt to dynamic scenes and fast-moving objects.
However, very small values of $\delta$, such as $\delta=1$ also increase the chance of adding noisy visual words, which may explain why $\delta=5$ performs the best.
Table \ref{tab:ablation_outlier_removal} also shows that our simple ``outlier removal'' step improves results as well, by encouraging spatio-temporal consistency from one frame to the next.

Note that our online adaptation system adds little overhead as it simply updates the existing dictionary of visual words.
No additional forward or backward passes through the network are required, unlike methods such as \cite{onavos}.

\paragraph{Effect of $\alpha$ on online adaptation}
\begin{figure}[t]
	\centering
	\resizebox{0.9\textwidth}{!}{
	\includegraphics[width=\textwidth]{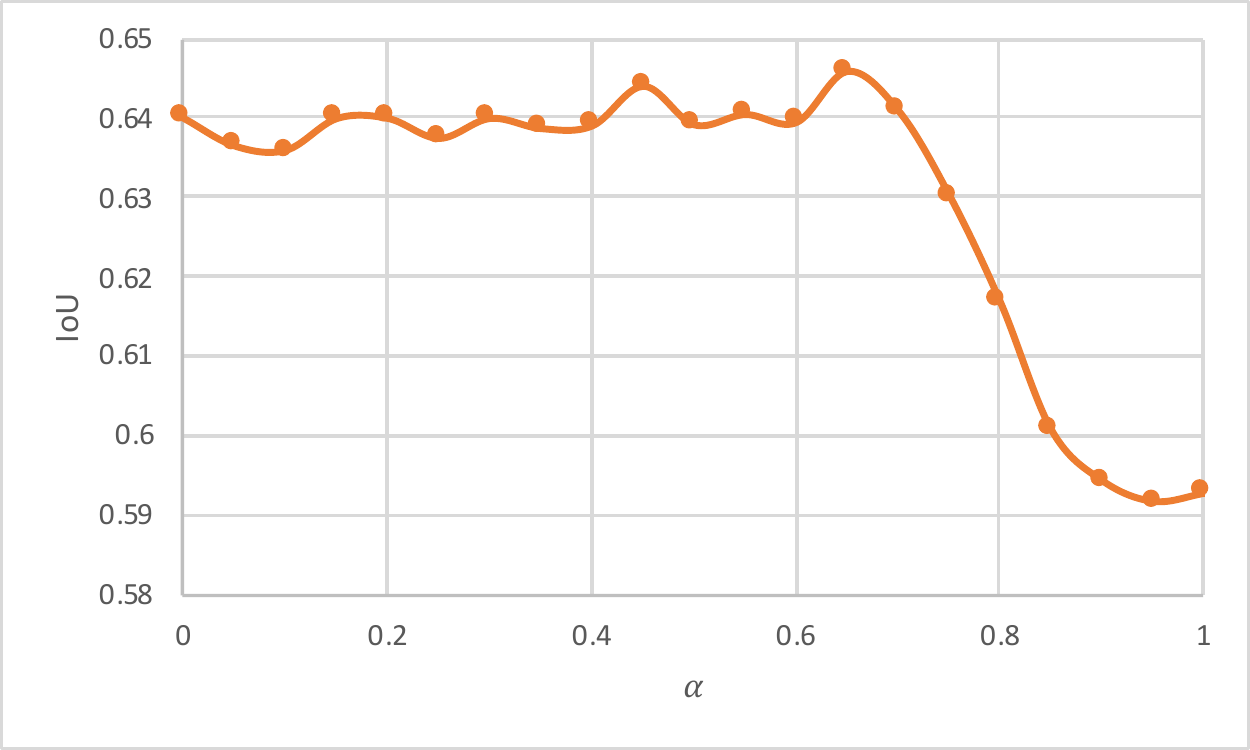}
	}
	\vspace{-0.3cm}
    \caption{{\bf The effect of the $\alpha$ hyper-parameter used in the online adaptation of on our method.}
    Note how our IoU on the DAVIS-2017 validation set barely changes for $\alpha \in [0, 0.7]$ showing that our algorithm is not very sensitive to this hyper-parameter.
    } \label{fig:alpha_online_adaptation}
\end{figure}

Figure~\ref{fig:alpha_online_adaptation} shows the effect of distance threshold in online adaptation, $\alpha$, on our accuracy on DAVIS-2017.
It shows that our algorithm is robust to the choice of alpha, performing similarly in a wide range of values when $\alpha$ lies in the interal $[0, 0.7]$, thus motivating our decision to use $\alpha = 0.5$ in our experiments.
Although our method is tolerant to a wide range of $\alpha$ values, setting it too high ($\alpha \geq 0.8$) introduces noisy visual words into online adaptation process which harms performance.

\subsection{Effect of visual word dictionary size}

\begin{figure}[!]
\vspace{-0.3cm}
	\centering
    \resizebox{0.9\textwidth}{!}{
	\includegraphics[width=\textwidth]{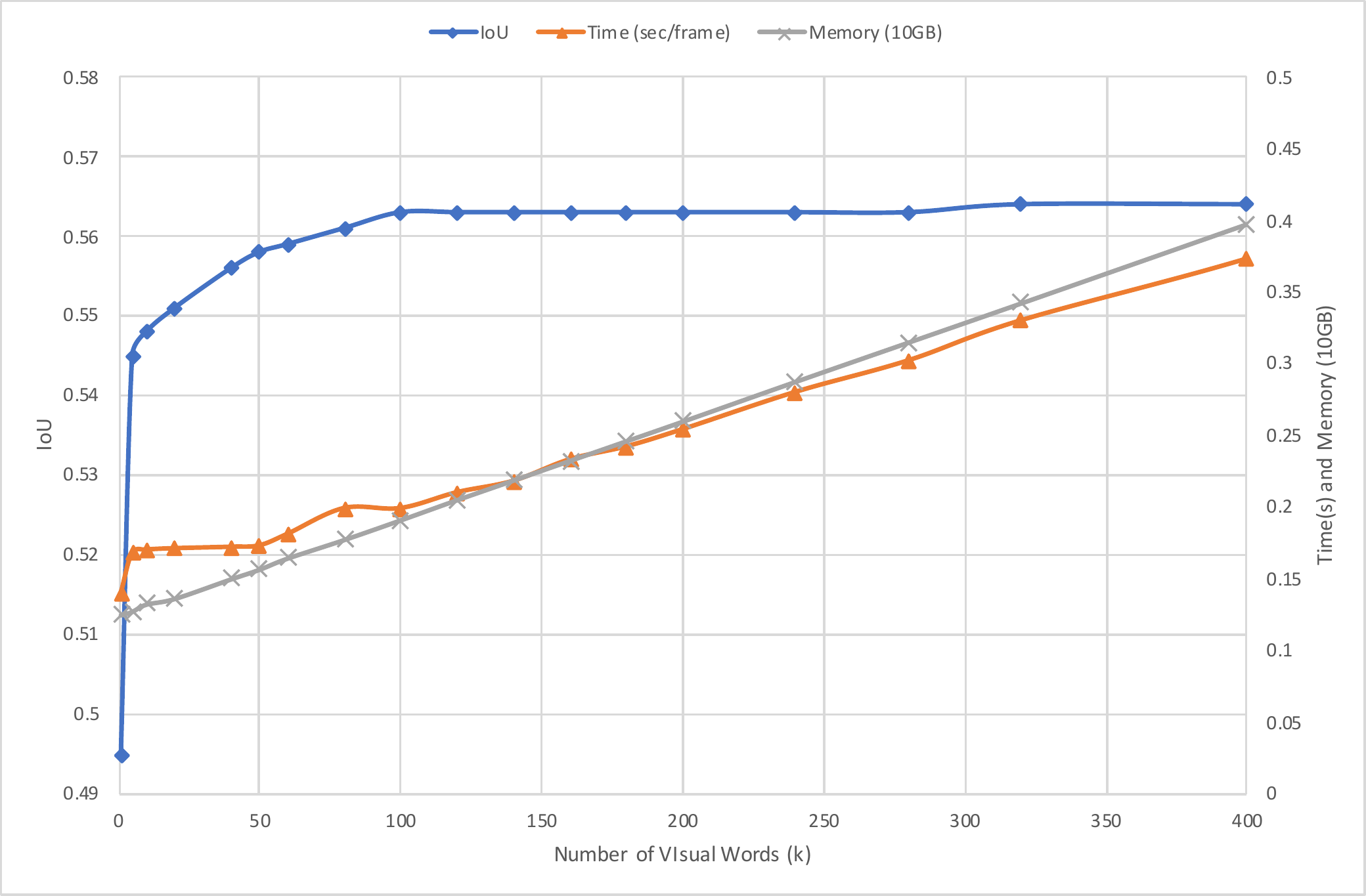}
    \caption{{\bf The effect of visual word dictionary size hyper-parameter on our algorithm's accuracy (IoU), runtime (s) and memory consumption (GB)}.
    The accuracy, in terms of the IoU ($\mathcal{J}$) starts saturating around $k = 50$.
    However, the runtime and memory increase linearly with the number of clusters, $k$.
    	}
    \label{fig:visual_word_acc_time_mem}
    \vspace{-0.4cm}
    }
\end{figure}
Table 3 of the main paper already showed the effect of the visual word dictionary size on performance.
Figure \ref{fig:visual_word_acc_time_mem} examines this in more detail by showing the accuracy (measured by the IoU, $\mathcal{J}$), the runtime in seconds and the memory consumption as a function of the size of the visual word dictionary.
We can see that the accuracy starts saturating after around $k = 50$ clusters.
However, the runtime and memory consumption increase linearly as the number of visual words is increased.
So we use $k = 50$ clusters, as it provides a good balance between accuracy and speed.

\begin{figure*}[]
	\centering
	
	\begin{tabularx}{\linewidth}{YYY}
		Input image & Visual words & Final segmentation 
	\end{tabularx}
		\includegraphics[width=1\textwidth]{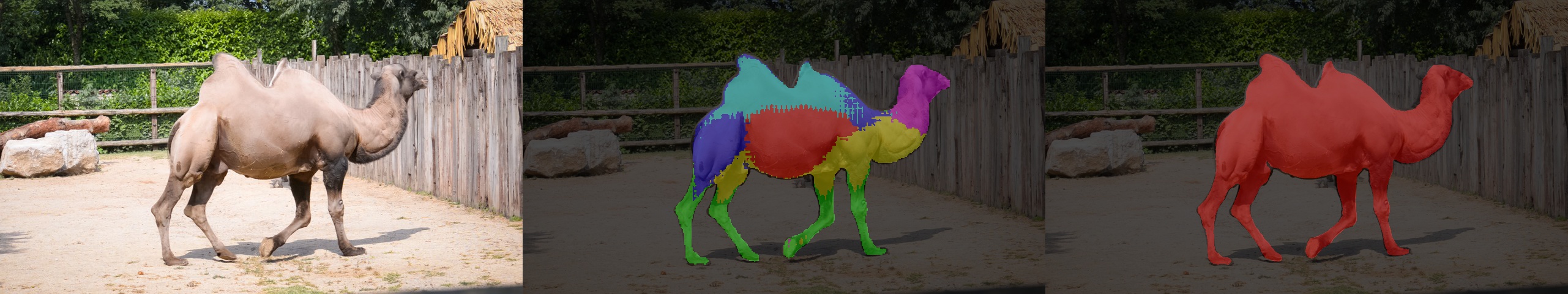}
		\includegraphics[width=1\textwidth]{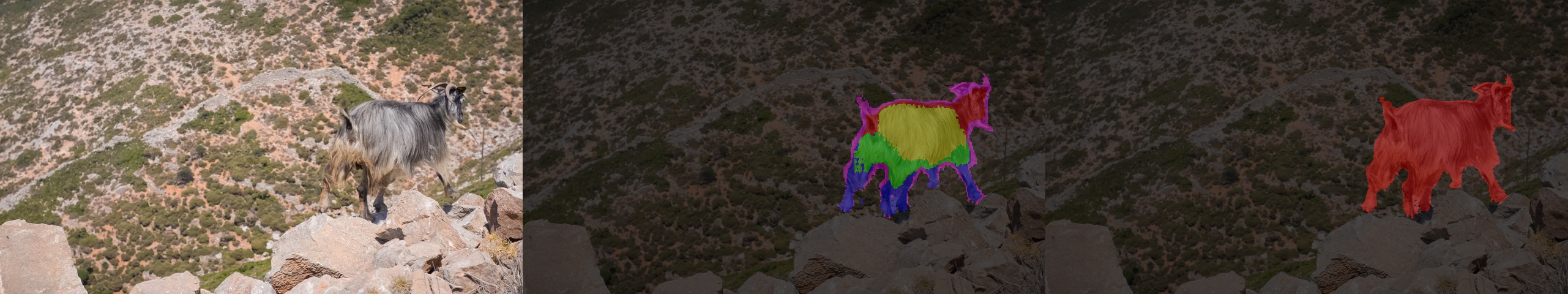}
		\includegraphics[width=1\textwidth]{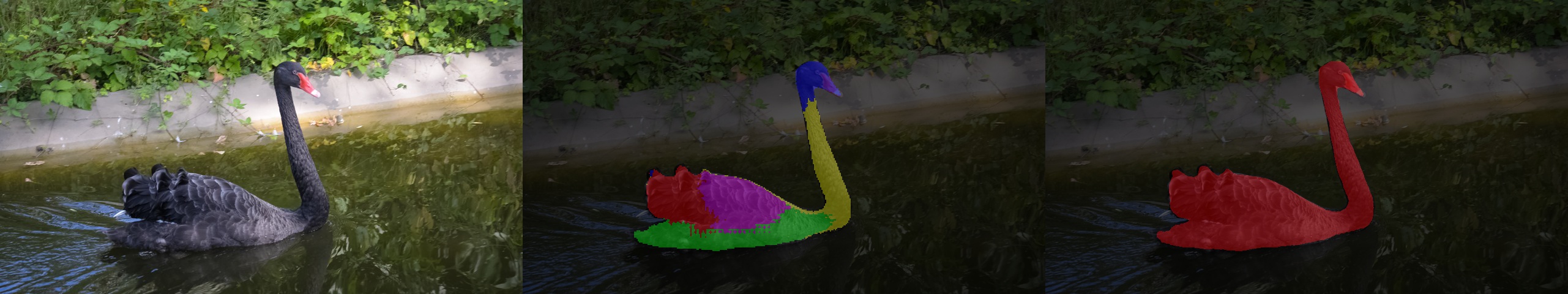}
		\includegraphics[width=1\textwidth]{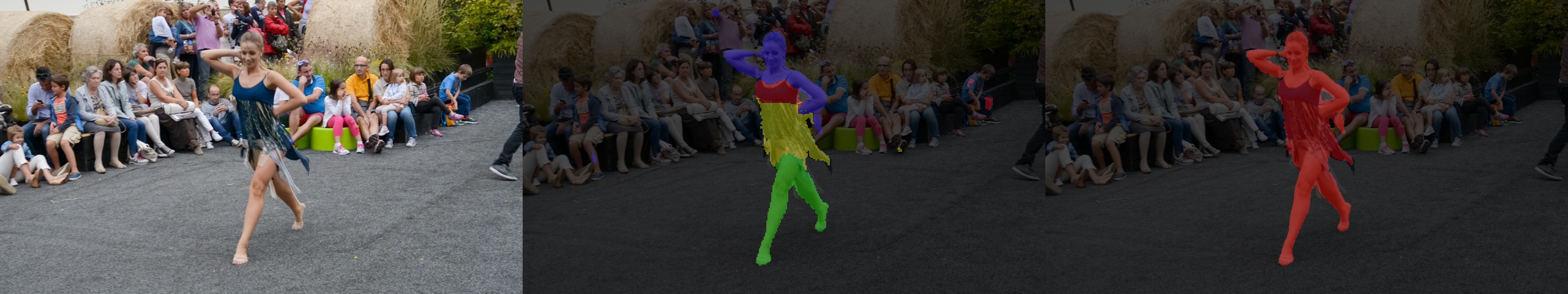}
		\includegraphics[width=1\textwidth]{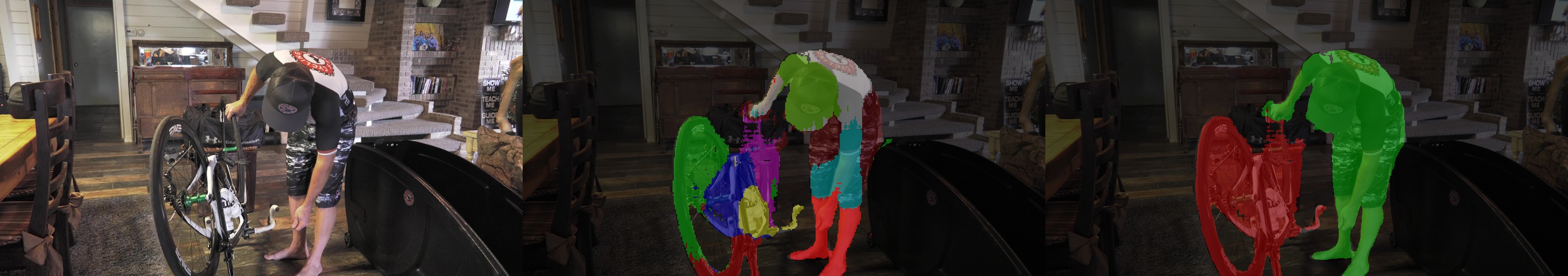}
		
		\caption{ {\bf Visual words formed by our model.} Each row represents a video from DAVIS-2017 dataset, with the original image (left), the object parts formed by our model in different colors (middle), and the segmentation output (right) obtained using our model. It can be seen that our model forms meaningful visual words which represent body parts in objects. 
		}
	\label{fig:part_visualisation}
\end{figure*}

\begin{figure*}[]
    \centering
    \setlength{\tabcolsep}{2pt}
    \begin{tabular}{cccc}
        Original Image & $k = 1$ & $k = 3$ & $k = 50$ \\
        \includegraphics[width=0.24\textwidth]{images/supplementary/increasing_parts/soapbox38.jpg}& \includegraphics[width=0.24\textwidth]{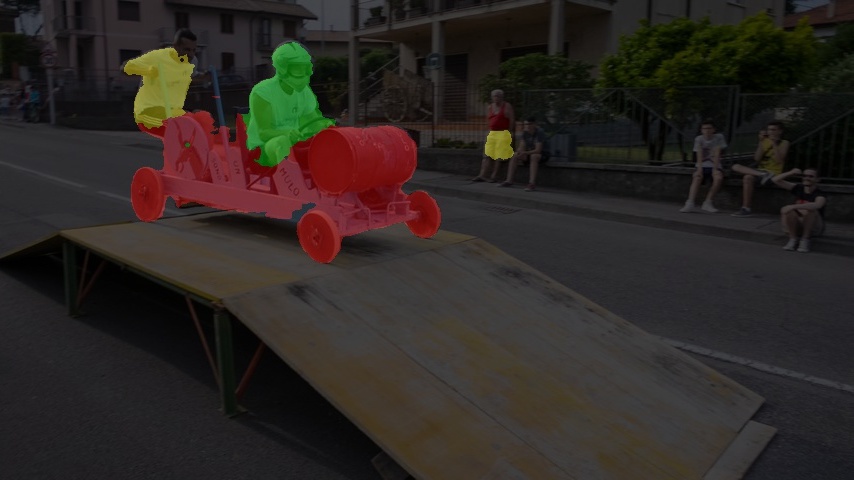}& \includegraphics[width=0.24\textwidth]{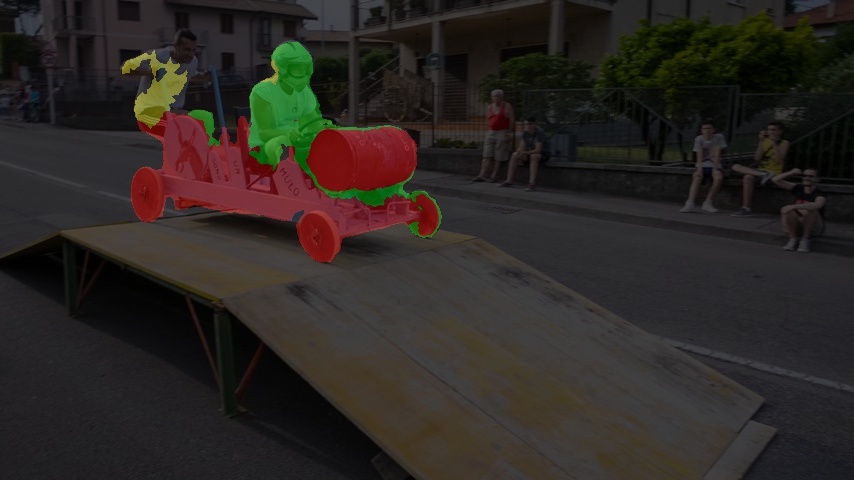}& \includegraphics[width=0.24\textwidth]{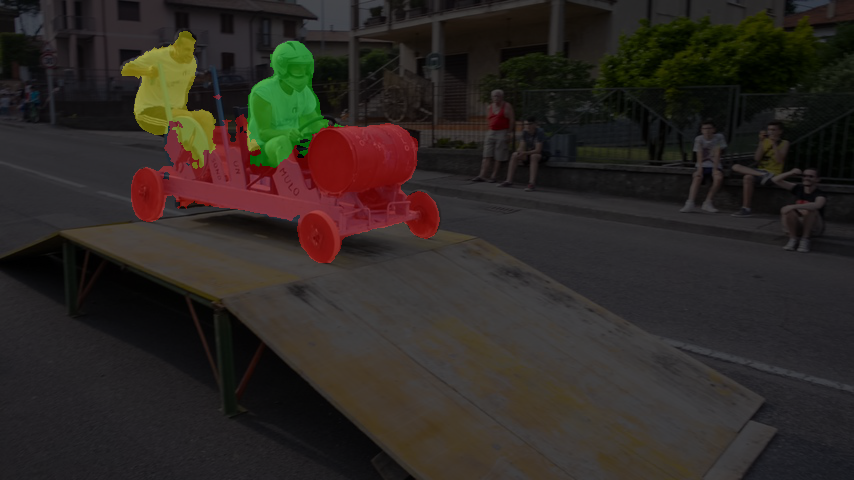} \\
        
        \includegraphics[width=0.24\textwidth]{images/supplementary/increasing_parts/paragliding49.jpg}& \includegraphics[width=0.24\textwidth]{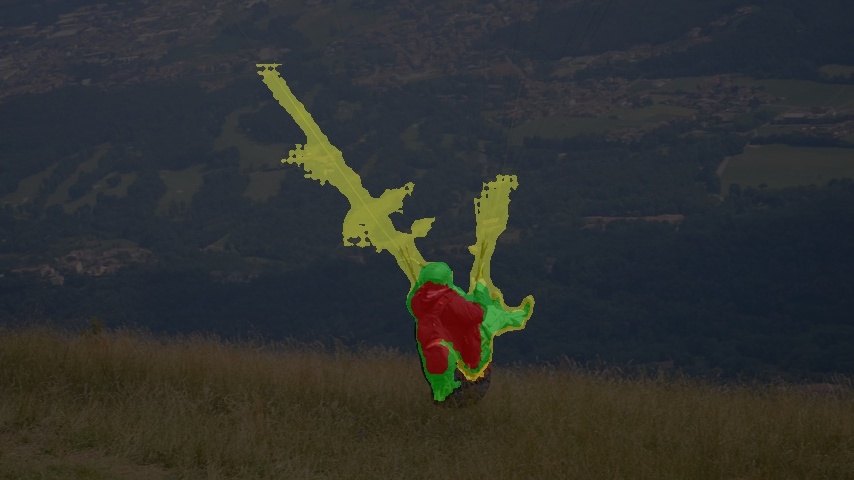}& \includegraphics[width=0.24\textwidth]{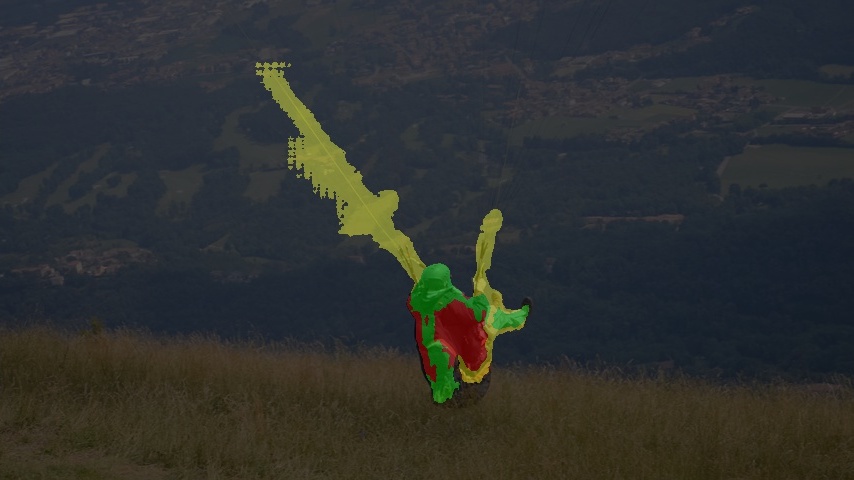}& \includegraphics[width=0.24\textwidth]{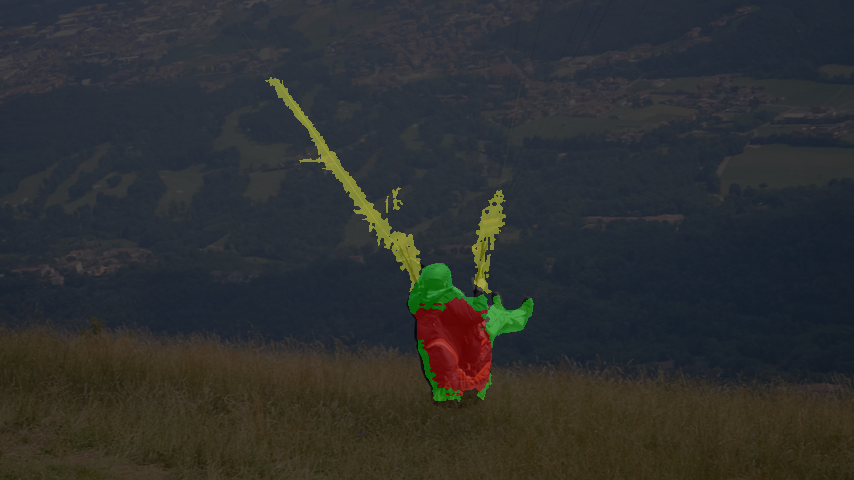} \\
        
        \includegraphics[width=0.24\textwidth]{images/supplementary/increasing_parts/libby5.jpg} & \includegraphics[width=0.24\textwidth]{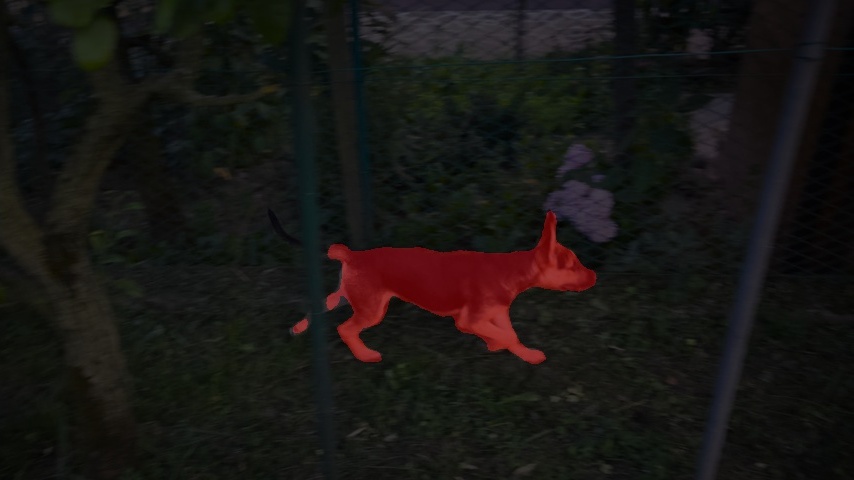} & \includegraphics[width=0.24\textwidth]{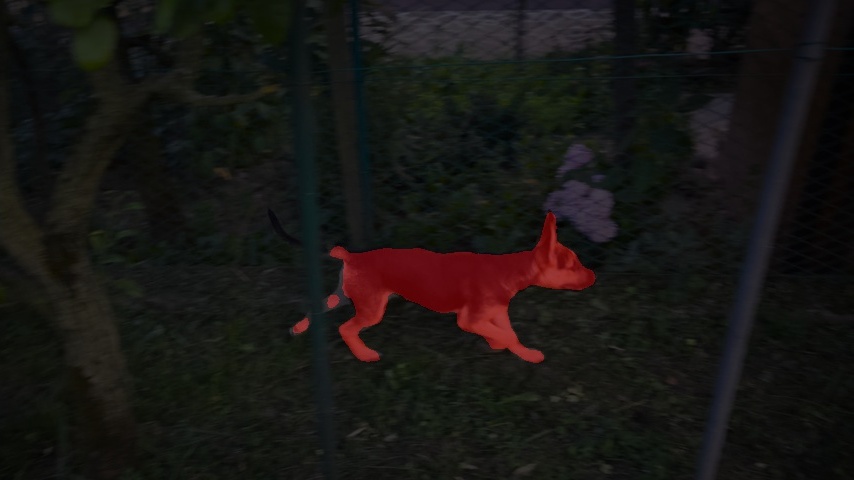} & \includegraphics[width=0.24\textwidth]{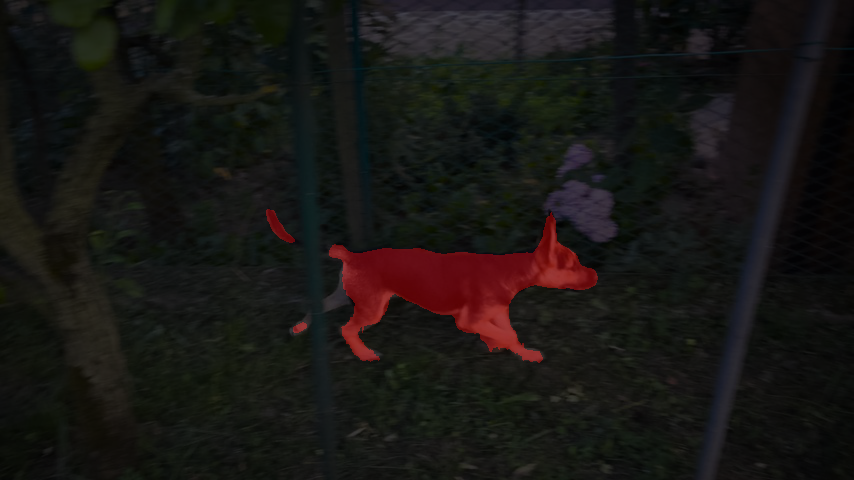}\\
        
        \includegraphics[width=0.24\textwidth]{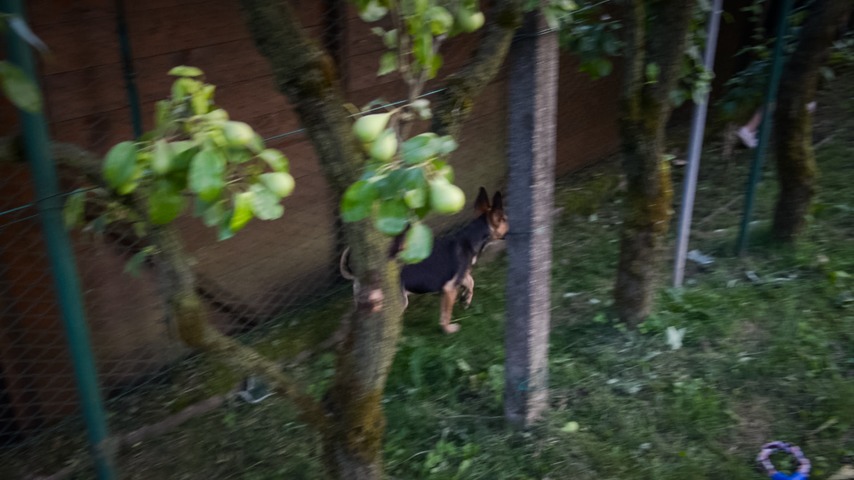} & \includegraphics[width=0.24\textwidth]{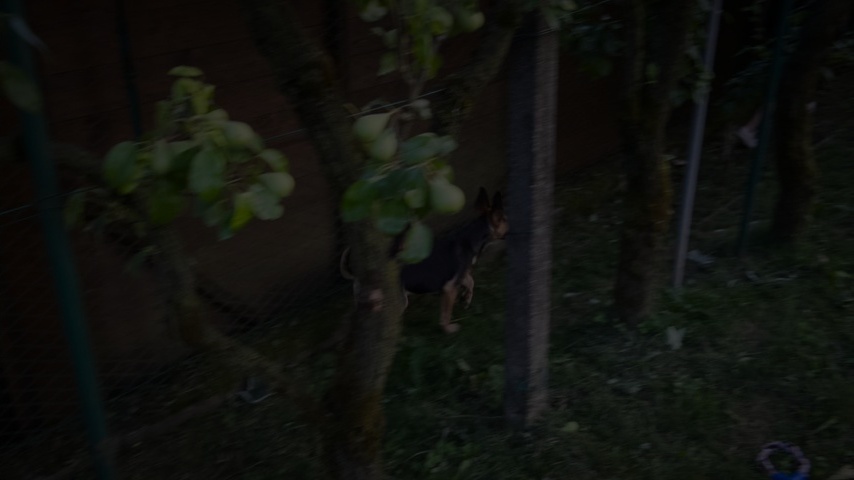} & \includegraphics[width=0.24\textwidth]{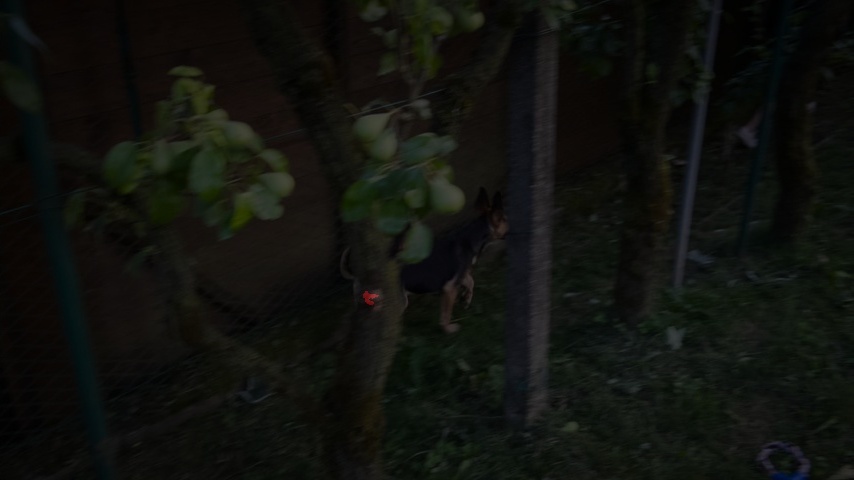} &
        \includegraphics[width=0.24\textwidth]{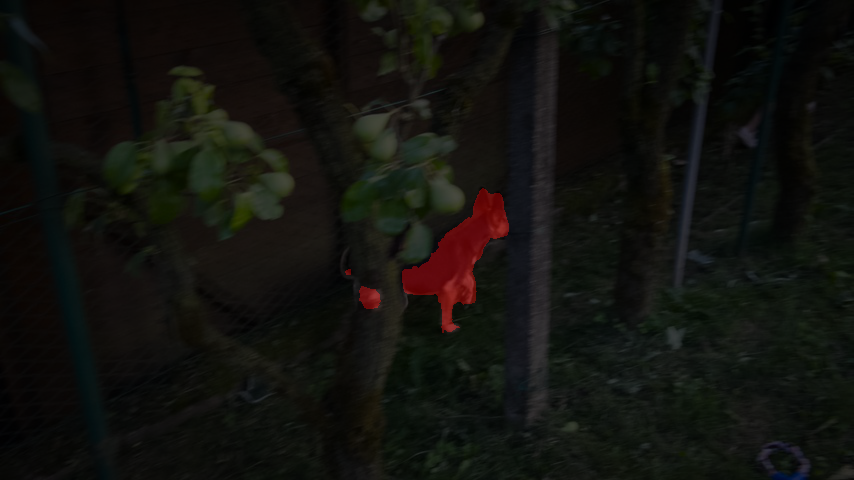} \\
	
	    \includegraphics[width=0.24\textwidth]{images/supplementary/increasing_parts/shooting6.jpg}& \includegraphics[width=0.24\textwidth]{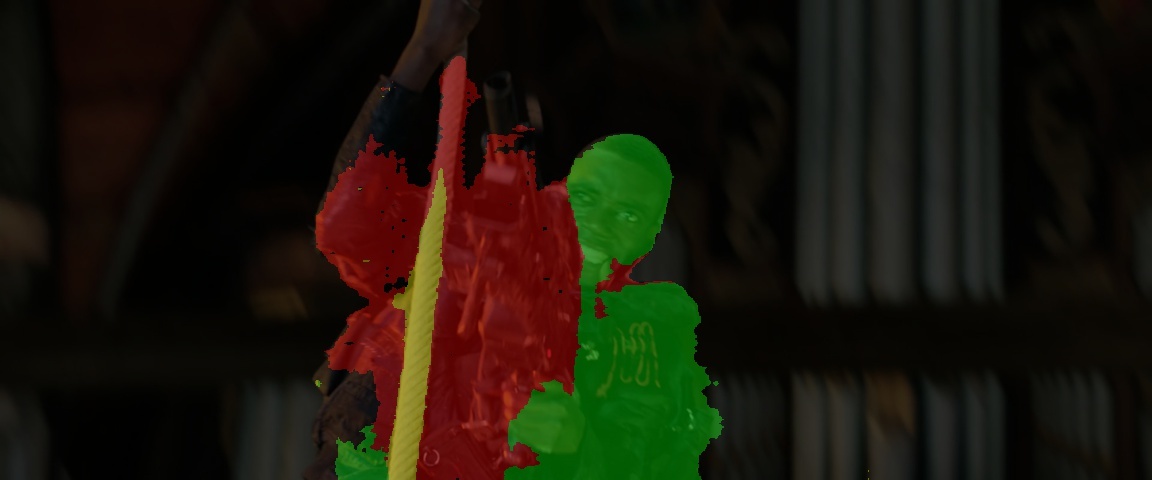}& \includegraphics[width=0.24\textwidth]{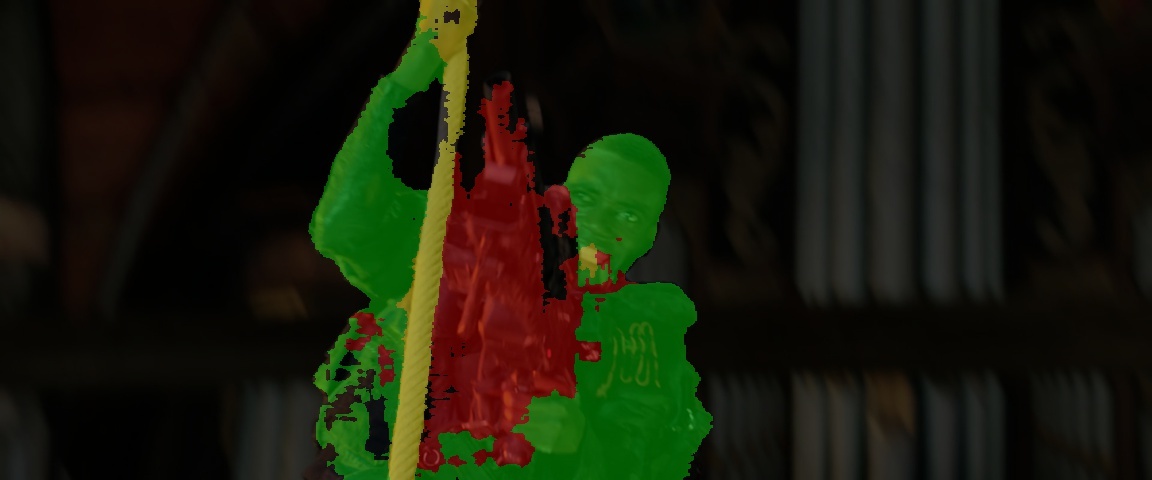}& \includegraphics[width=0.24\textwidth]{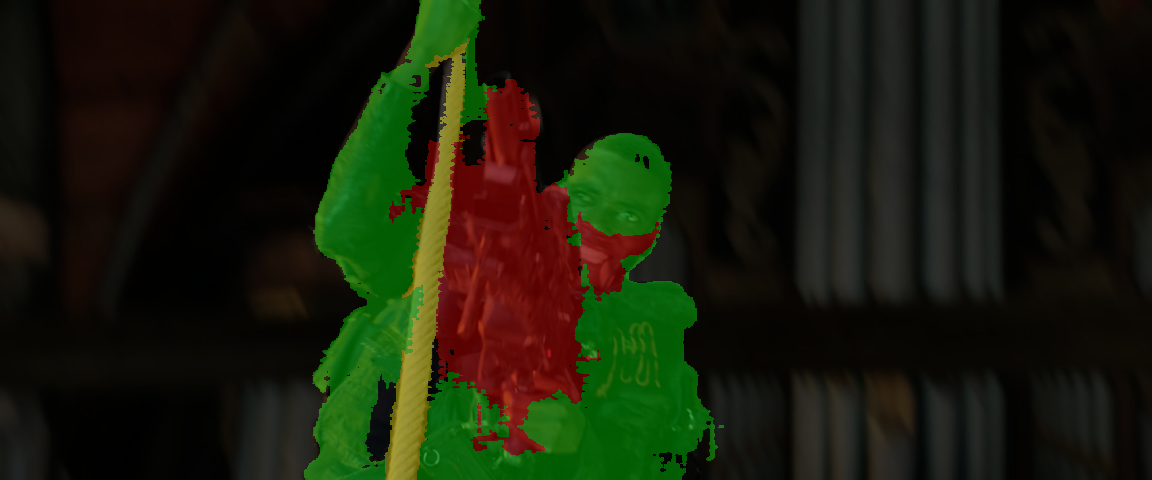} \\
	
	    \includegraphics[width=0.24\textwidth]{images/supplementary/increasing_parts/horse17.jpg}& \includegraphics[width=0.24\textwidth]{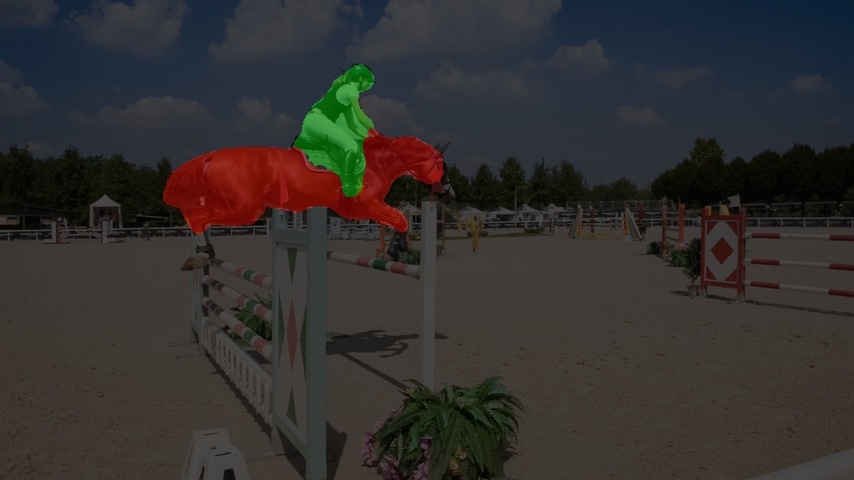}& \includegraphics[width=0.24\textwidth]{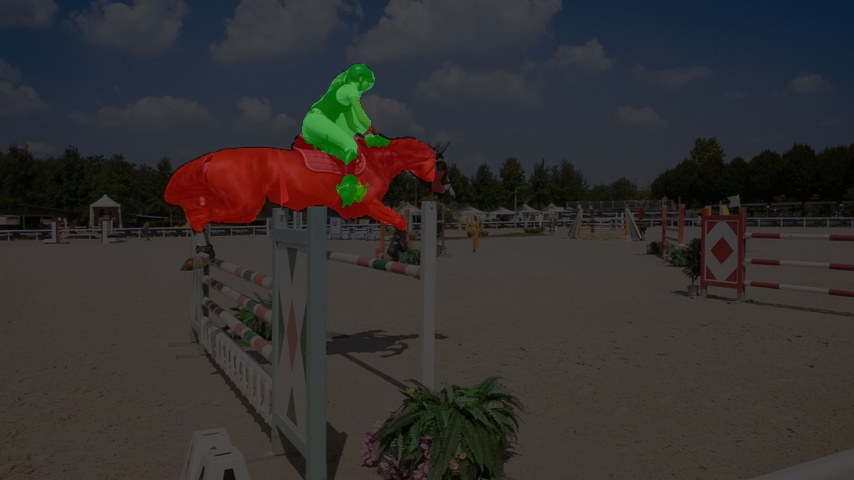}& \includegraphics[width=0.24\textwidth]{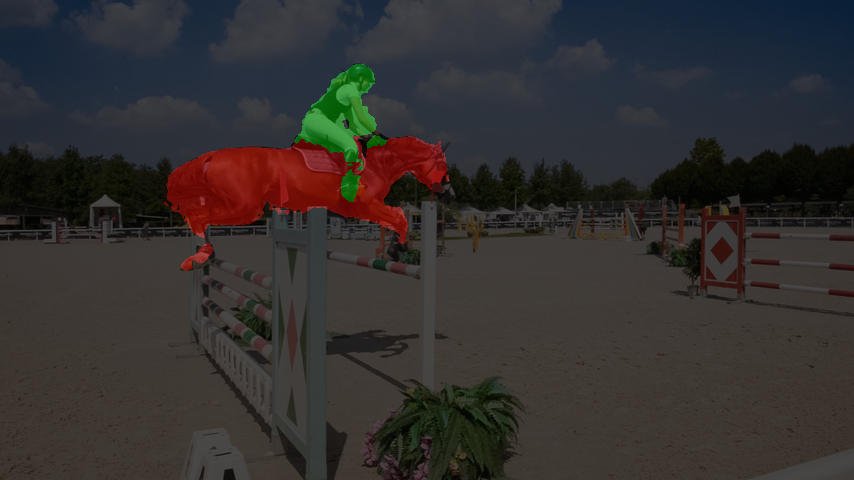}
    \end{tabular}
    \setlength{\tabcolsep}{6pt}
    \caption{ {\bf The effect of visual word object representation on qualitative segmentation outputs.} This figure shows that increasing the size of the visual word dictionary ({K}) improves the representation of the object, and thus improves segmentation outputs (qualitative). 
    This is because it can better capture the intra-object variance. For instance, the lost face of the human in yellow (first row), the lost tail of the dog (third row), and the missing legs of the horse (last row) have been recovered in the last column (50 visual words), because of this property. Similarly, 
    our method can address partial occlusions by representing different object parts using visual words, and tracking them robustly over the video (fourth row). All the visual words are learned in an unsupervised manner to represent object parts, as described in the main paper. The results are obtained without any fine-tuning.}
    \label{fig:visual_word_qualitative_effect}
\end{figure*}

Additionally, Fig.~\ref{fig:part_visualisation} presents some of the visual words, corresponding to object parts, that are automatically learned by our model. Finally, Fig.~\ref{fig:visual_word_qualitative_effect} shows the qualitative effect on our final segmentation results by increasing the number of visual words in the dictionary.


\vspace{-0.3in}
\section{Qualitative Results}
\label{sec:supp_qualitative_res}
Figures \ref{fig:comparison_qualitative} and \ref{fig:comparison_qualitative_failure} present further qualitative comparisons of our method with RGMP \cite{Oh_2018_CVPR}.
\def \imwidth {0.18}
\def \imopts {valign=m}

\setlength{\tabcolsep}{1pt}
\vspace{-0.3in}
\begin{figure*}[h!]
	\begin{tabular}{m{1cm}ccccc}
		
	\footnotesize{Ours}	& 
		\includegraphics[valign=m,width=\imwidth\textwidth]{images/segmentation/scooter-black/00001.jpg}
		&
		\includegraphics[valign=m,width=\imwidth\textwidth]{images/segmentation/scooter-black/00008.jpg}
		&
		\includegraphics[valign=m,width=\imwidth\textwidth]{images/segmentation/scooter-black/00017.jpg}
		&
		\includegraphics[valign=m,width=\imwidth\textwidth]{images/segmentation/scooter-black/00027.jpg}
		&
		\includegraphics[valign=m,width=\imwidth\textwidth]{images/segmentation/scooter-black/00042.jpg} \\ [0.8cm]
		
	\footnotesize{RGMP \cite{Oh_2018_CVPR}}	&
		\includegraphics[valign=m,width=\imwidth\textwidth]{images/rgmp/scooter-black/00001.jpg}
		&
		\includegraphics[valign=m,width=\imwidth\textwidth]{images/rgmp/scooter-black/00008.jpg}
		&
		\includegraphics[valign=m,width=\imwidth\textwidth]{images/rgmp/scooter-black/00017.jpg}
		&
		\includegraphics[valign=m,width=\imwidth\textwidth]{images/rgmp/scooter-black/00027.jpg}
		&
		\includegraphics[valign=m,width=\imwidth\textwidth]{images/rgmp/scooter-black/00042.jpg} \\ [0.85cm]

	\footnotesize{Ours}	& 
		\includegraphics[valign=m,width=\imwidth\textwidth]{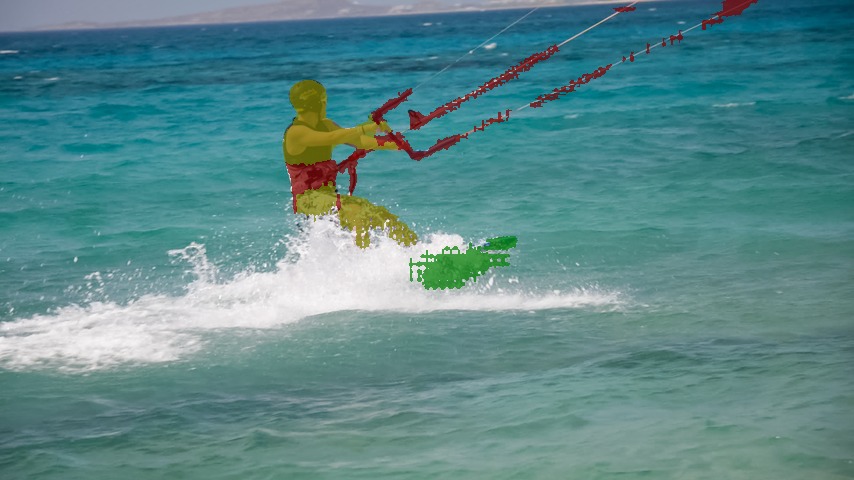}
		&
		\includegraphics[valign=m,width=\imwidth\textwidth]{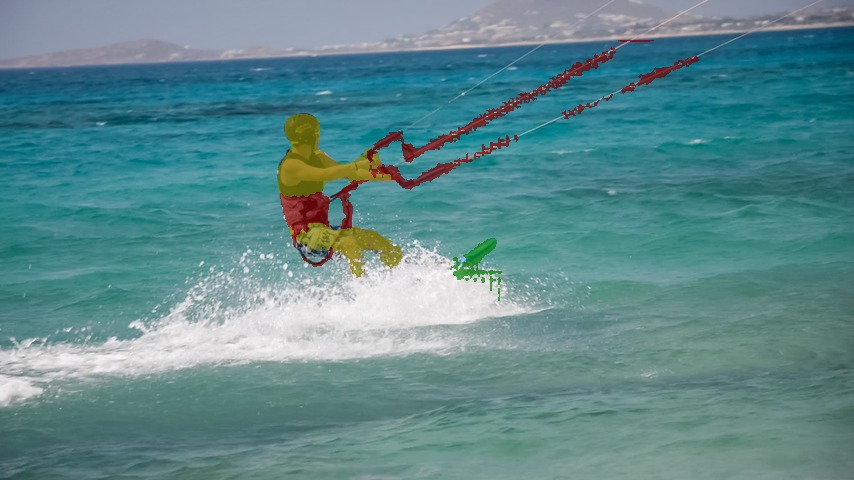}
		&
		\includegraphics[valign=m,width=\imwidth\textwidth]{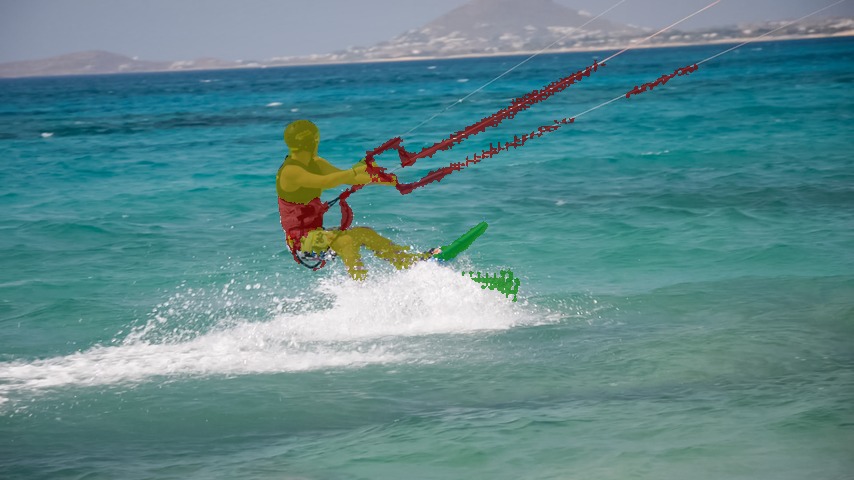}
		&
		\includegraphics[valign=m,width=\imwidth\textwidth]{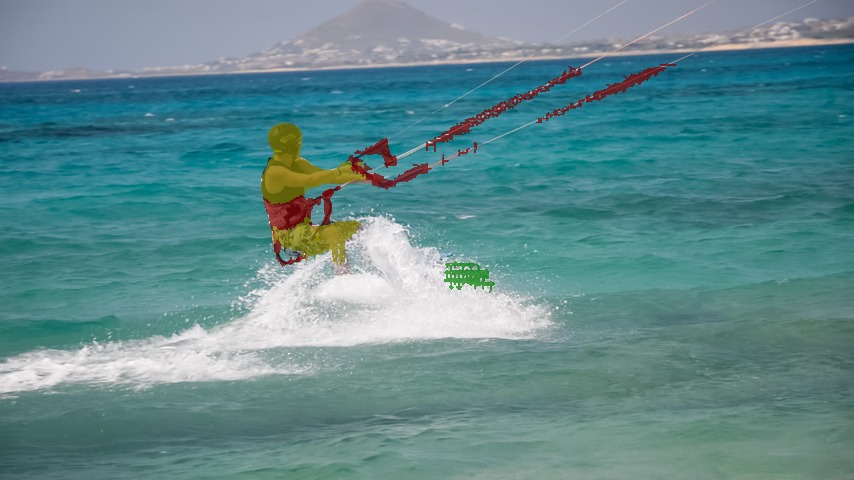}
		&
		\includegraphics[valign=m,width=\imwidth\textwidth]{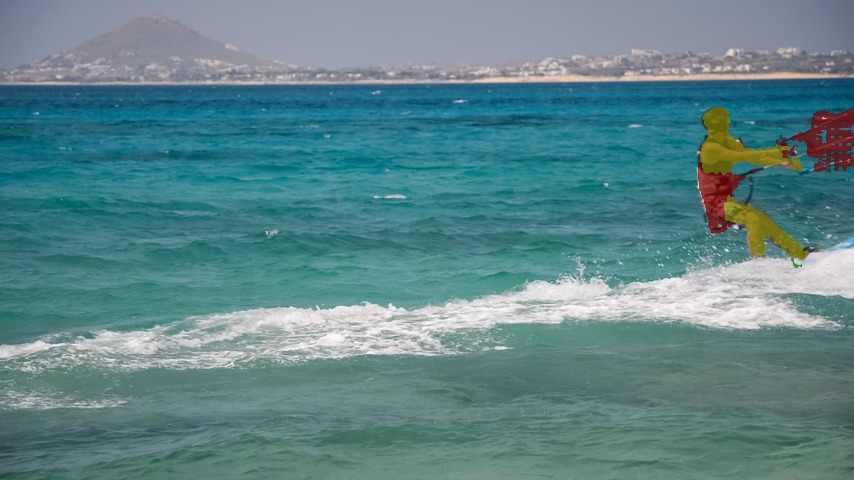} \\ [0.8cm]
		
	\footnotesize{RGMP \cite{Oh_2018_CVPR}}	&
		\includegraphics[valign=m,width=\imwidth\textwidth]{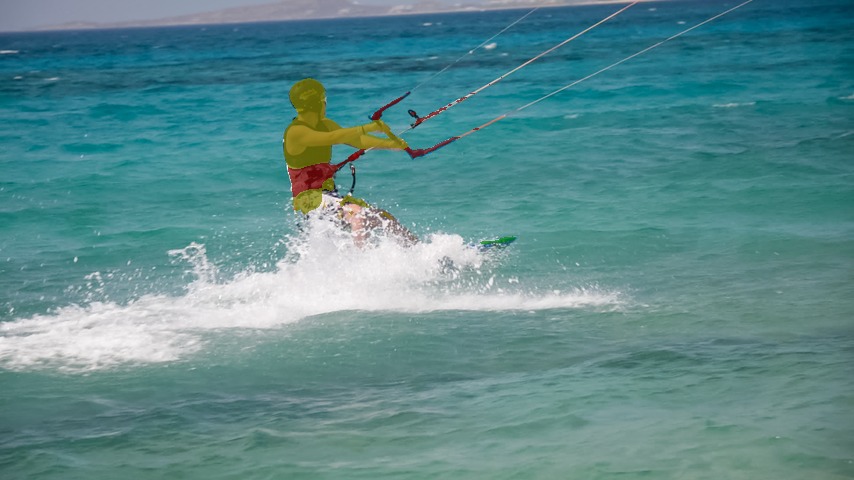}
		&
		\includegraphics[valign=m,width=\imwidth\textwidth]{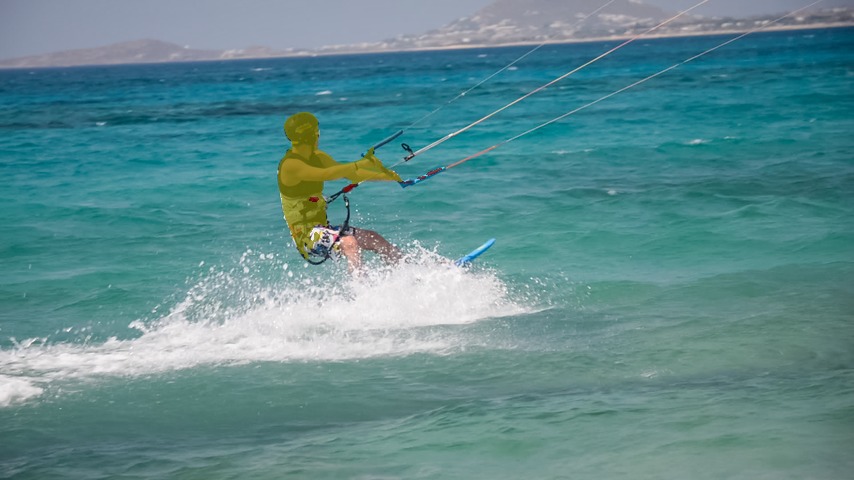}
		&
		\includegraphics[valign=m,width=\imwidth\textwidth]{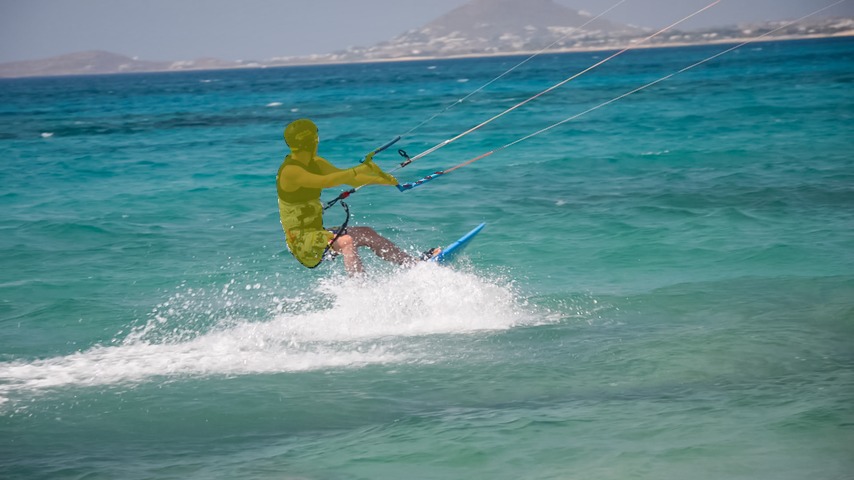}
		&
		\includegraphics[valign=m,width=\imwidth\textwidth]{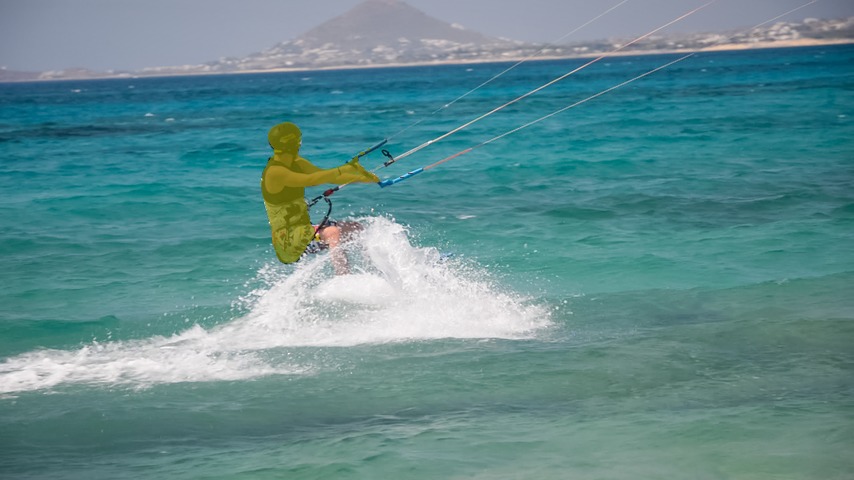}
		&
		\includegraphics[valign=m,width=\imwidth\textwidth]{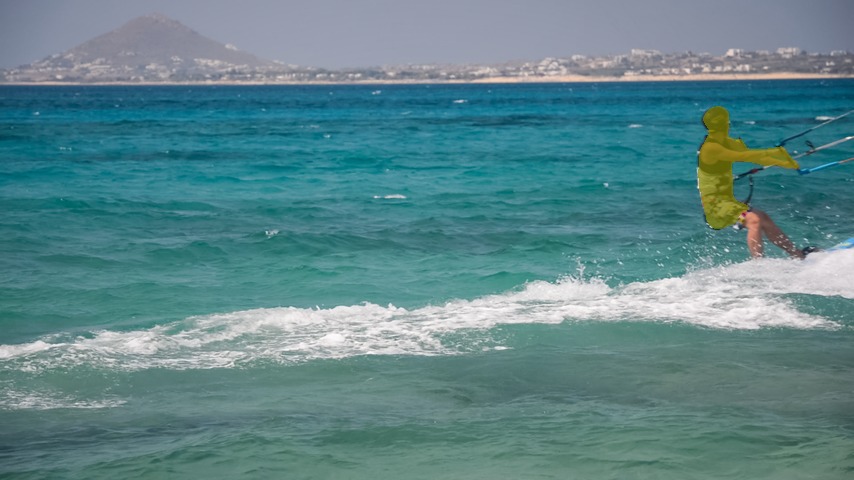} \\ [0.85cm]
		
	\footnotesize{Ours}	& 
		\includegraphics[valign=m,width=\imwidth\textwidth]{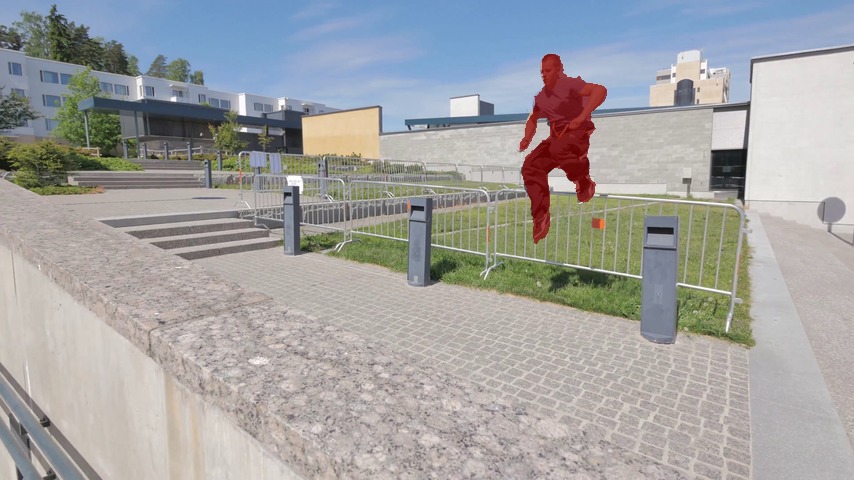}
		&
		\includegraphics[valign=m,width=\imwidth\textwidth]{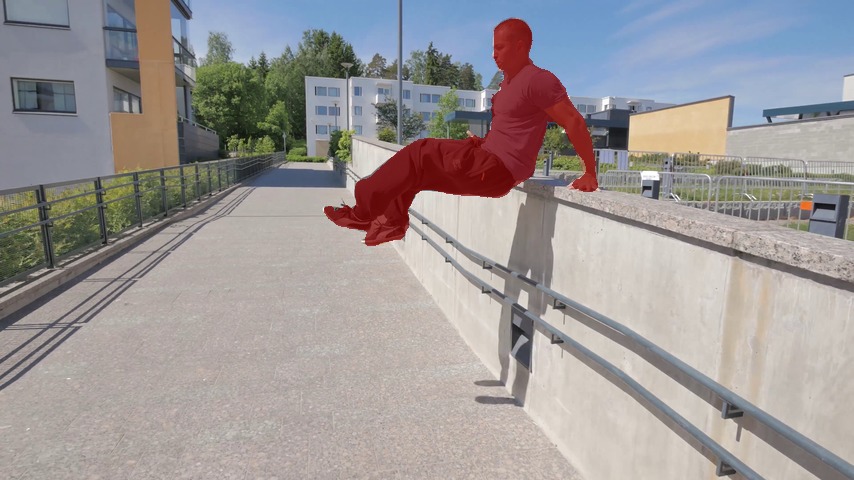}
		&
		\includegraphics[valign=m,width=\imwidth\textwidth]{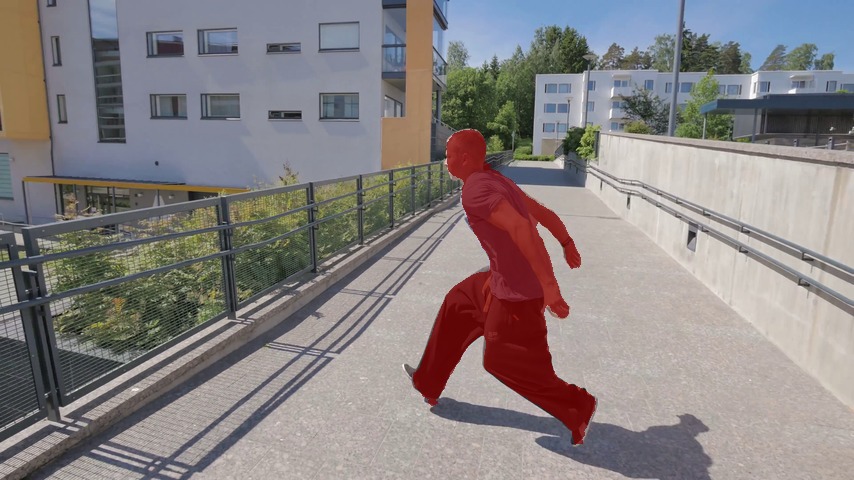}
		&
		\includegraphics[valign=m,width=\imwidth\textwidth]{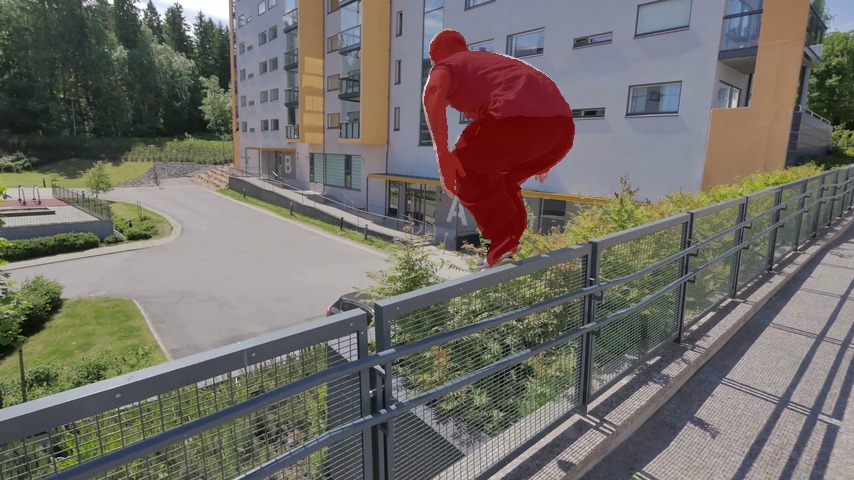}
		&
		\includegraphics[valign=m,width=\imwidth\textwidth]{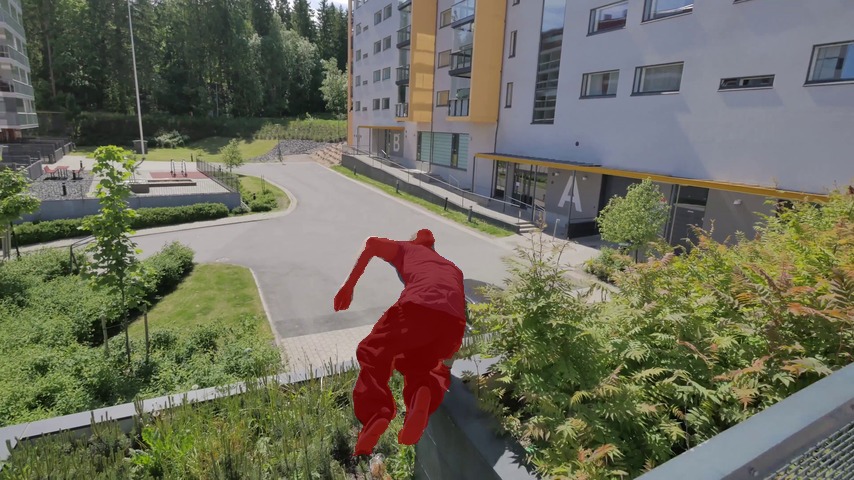} \\ [0.8cm]
		
	\footnotesize{RGMP \cite{Oh_2018_CVPR}}	&
		\includegraphics[valign=m,width=\imwidth\textwidth]{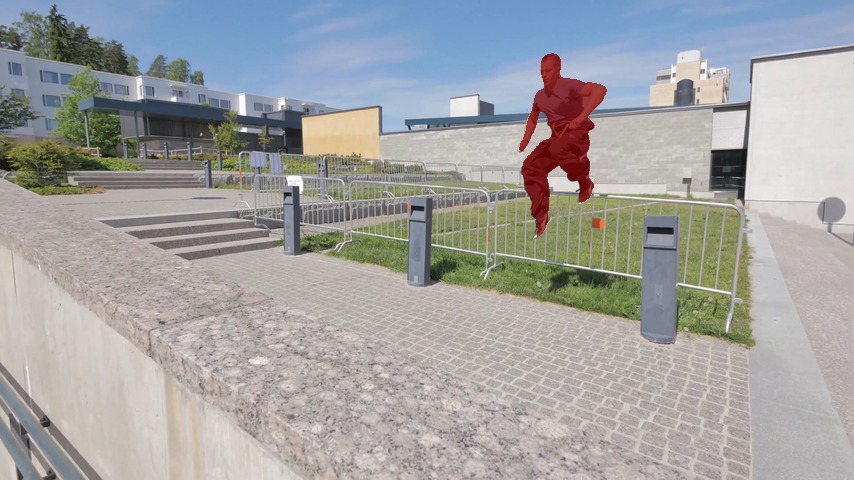}
		&
		\includegraphics[valign=m,width=\imwidth\textwidth]{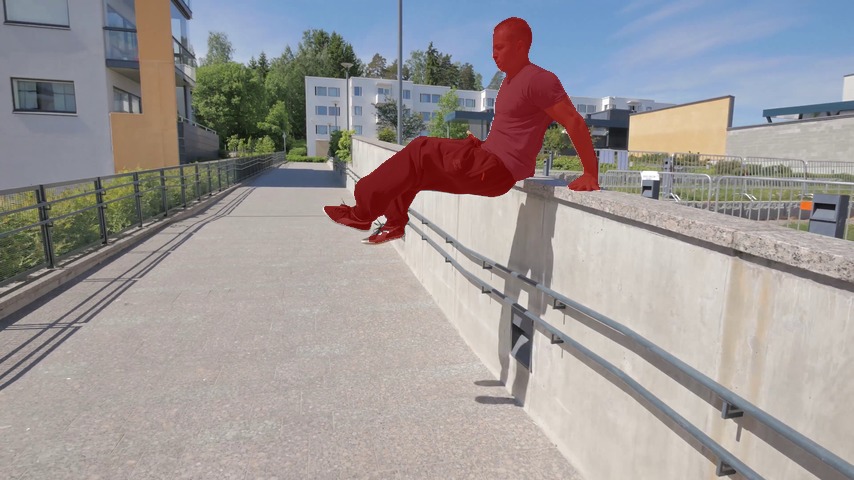}
		&
		\includegraphics[valign=m,width=\imwidth\textwidth]{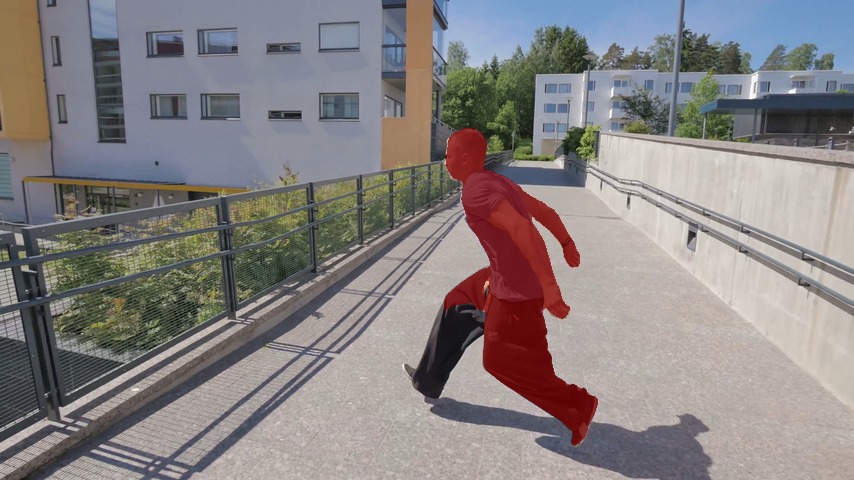}
		&
		\includegraphics[valign=m,width=\imwidth\textwidth]{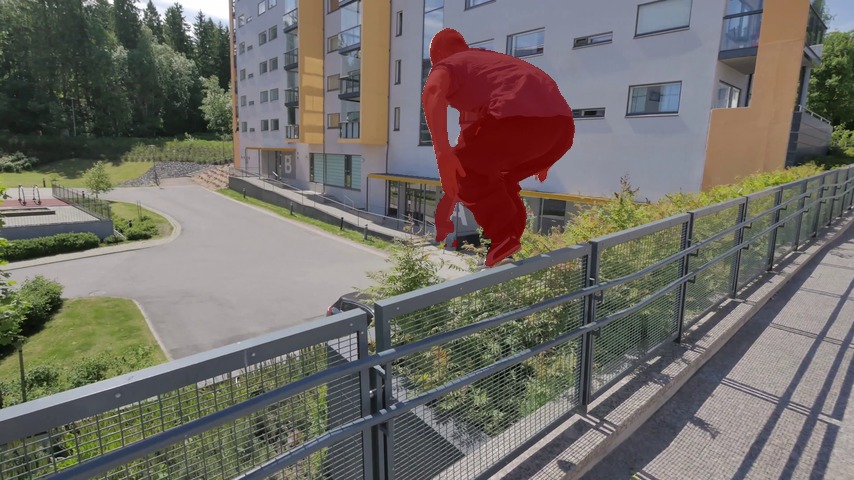}
		&
		\includegraphics[valign=m,width=\imwidth\textwidth]{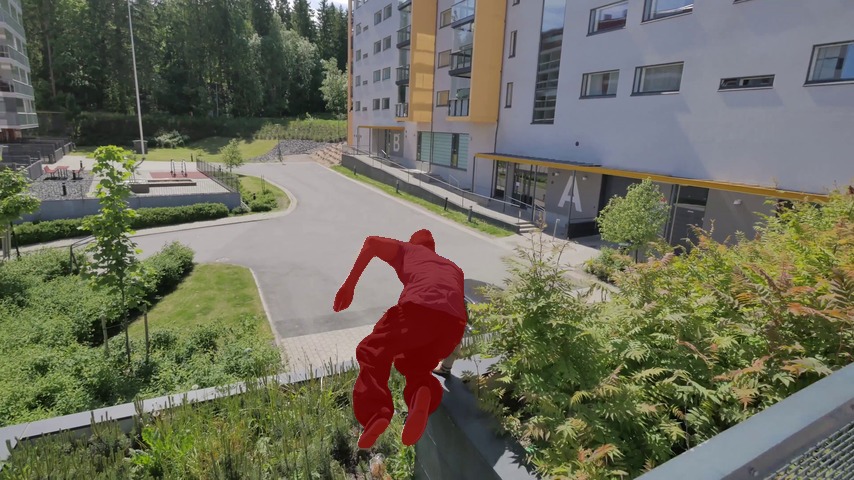} \\ [0.85cm]

	\footnotesize{Ours}	& 
		\includegraphics[valign=m,width=\imwidth\textwidth]{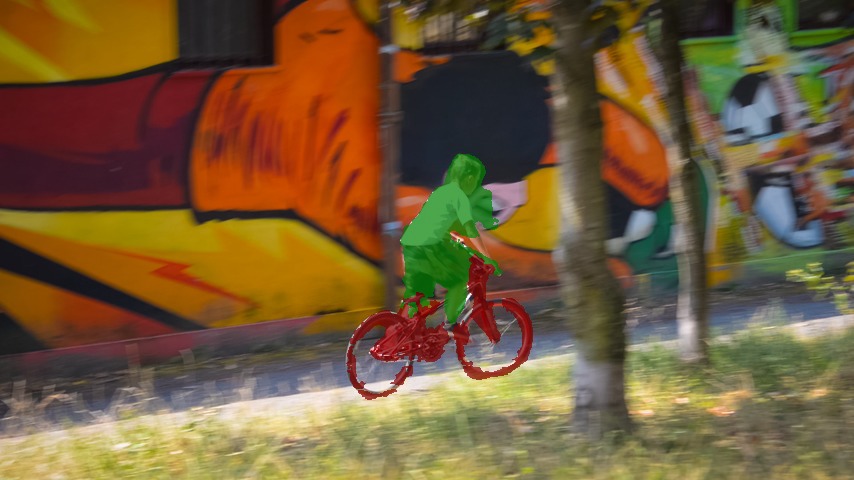}
		&
		\includegraphics[valign=m,width=\imwidth\textwidth]{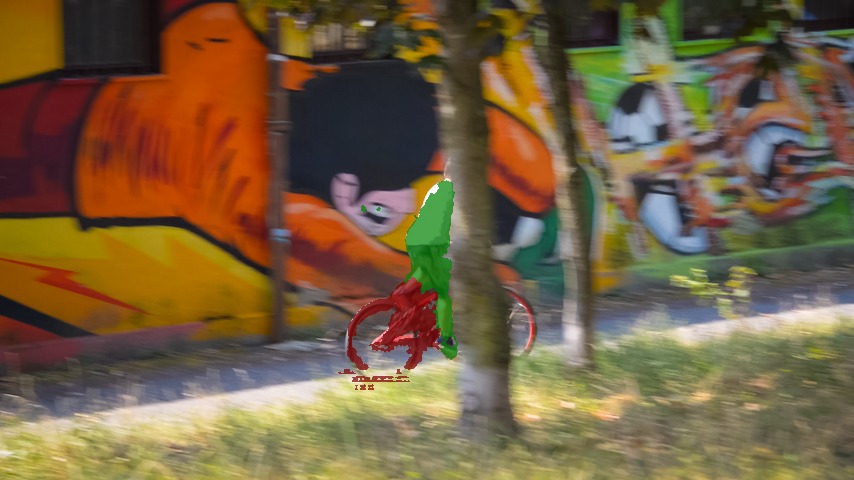}
		&
		\includegraphics[valign=m,width=\imwidth\textwidth]{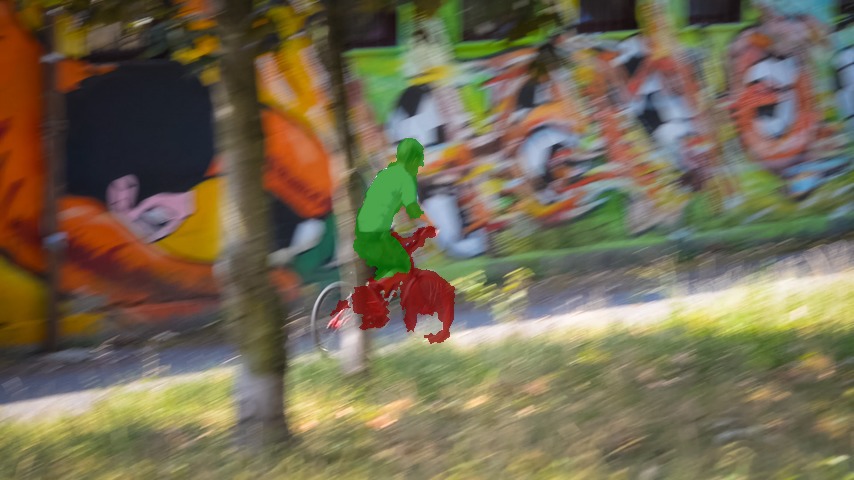}
		&
		\includegraphics[valign=m,width=\imwidth\textwidth]{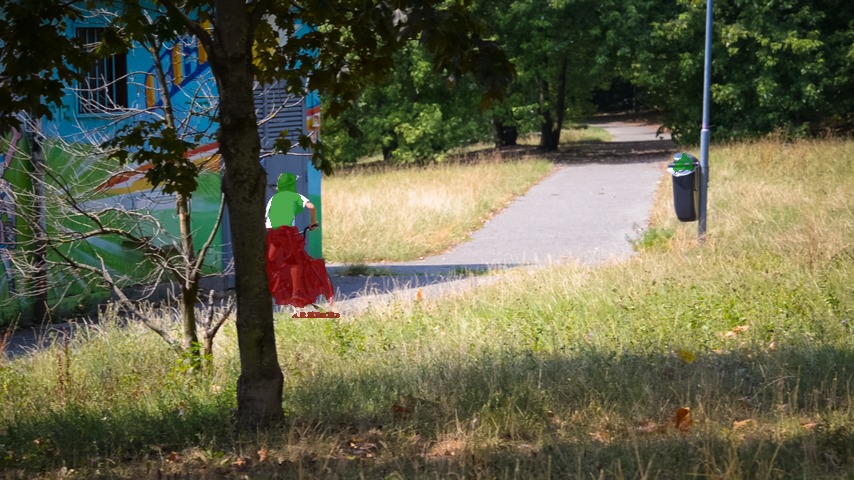}
		&
		\includegraphics[valign=m,width=\imwidth\textwidth]{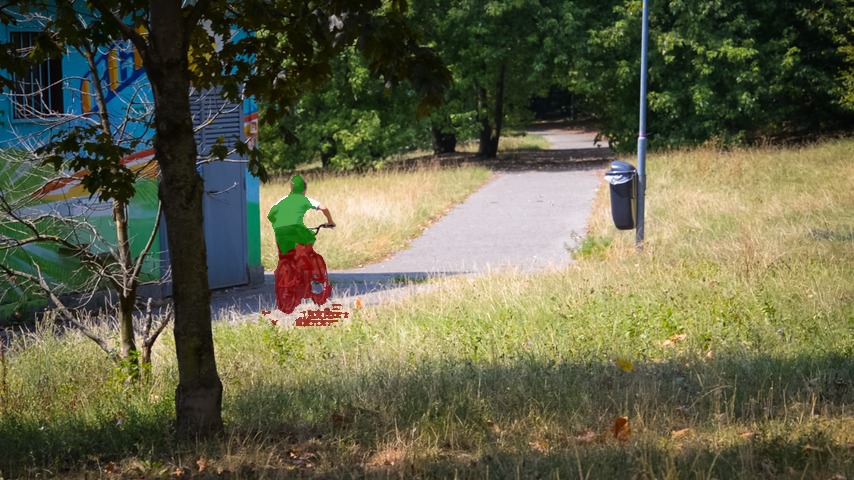} \\ [0.8cm]
		
	\footnotesize{RGMP \cite{Oh_2018_CVPR}}	&
		\includegraphics[valign=m,width=\imwidth\textwidth]{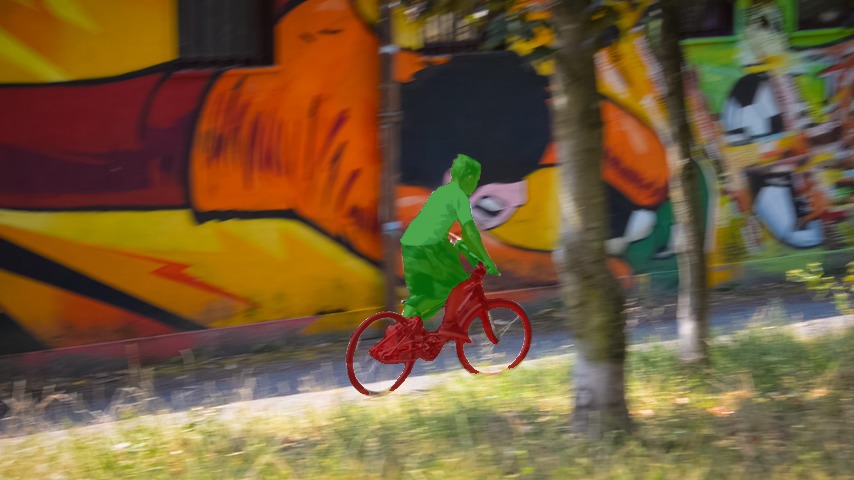}
		&
		\includegraphics[valign=m,width=\imwidth\textwidth]{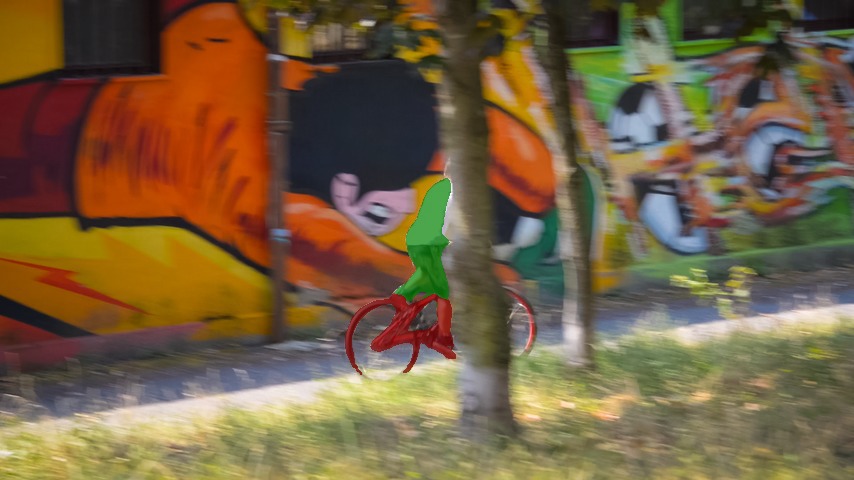}
		&
		\includegraphics[valign=m,width=\imwidth\textwidth]{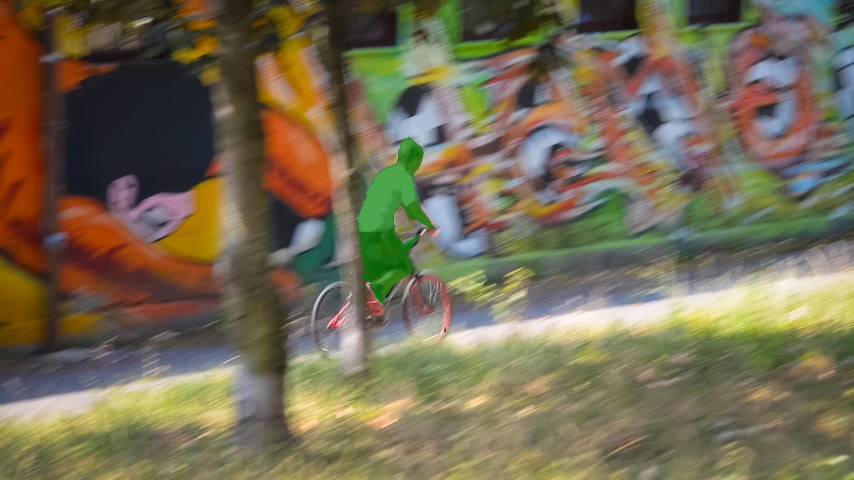}
		&
		\includegraphics[valign=m,width=\imwidth\textwidth]{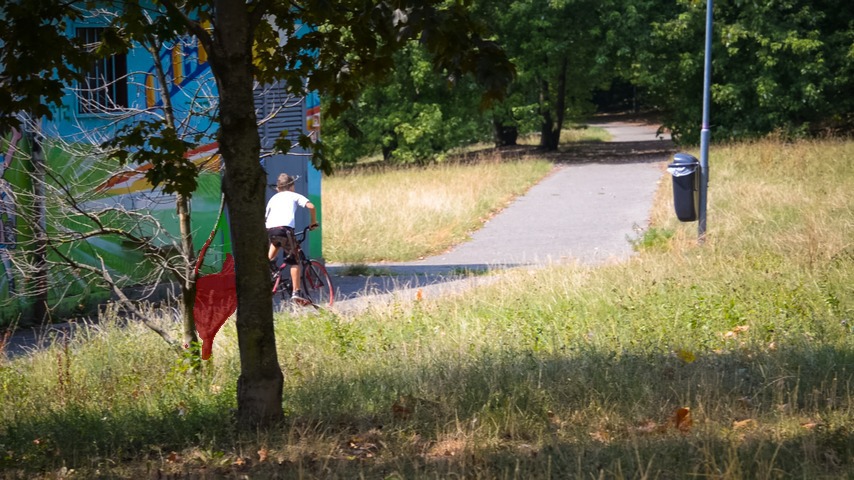}
		&
		\includegraphics[valign=m,width=\imwidth\textwidth]{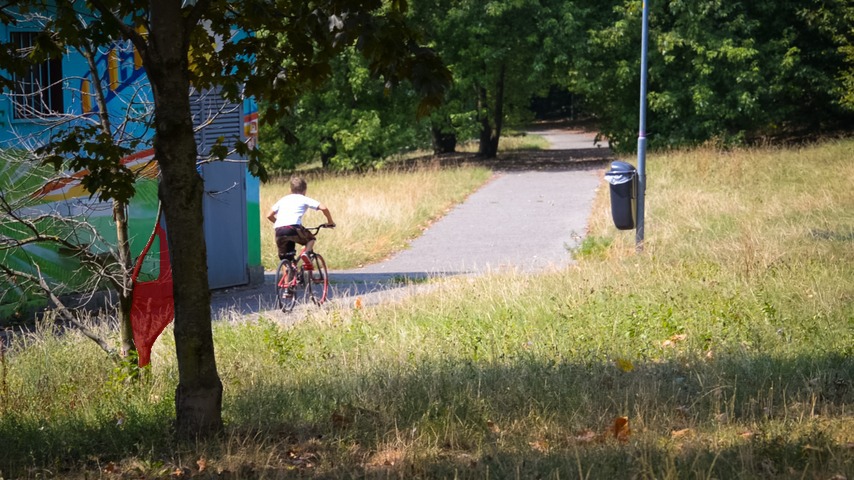}
	
	\end{tabular}
	\vspace{\baselineskip}
	\caption{\textbf{Success cases of our method, and comparison to RGMP \cite{Oh_2018_CVPR}.} In each of these videos, our method is able to accurately track the objects labelled in the first frame throughout the video.
	\textit{First video:} Our algorithm accurately segments the person throughout the video, whilst RGMP cannot deal with the scale and viewpoint changes of the person and mistakes him for the motorbike.
	\textit{Second video:} Our method is able to segment the kite-surfing harness and wires whilst RGMP loses track of these fine structures. Additionally, note how we are able to segment the heavily-occluded surf-board throughout the video, unlike RGMP.
	\textit{Third video:} Both methods perform well on this example.
	\textit{Fourth video:} RGMP loses track of the cyclist from the fourht frame onwards, whereas our method is robust to this occlusion. Mask propagation methods, such as RGMP, struggle with such occlusions. Our representation of objects as visual words is more robust in these situations.
	Full video results of these clips are included in the supplementary video.
	} 
	\label{fig:comparison_qualitative}
\end{figure*}

\setlength{\tabcolsep}{6pt}
\def \imwidth {0.18}
\def \imopts {valign=m}

\setlength{\tabcolsep}{1pt}

\begin{figure*}[t]
	\begin{tabular}{m{1cm}ccccc}
		
	\footnotesize{Ours}	&  \includegraphics[valign=m,width=\imwidth\textwidth]{images/segmentation/loading/00001.jpg}
		& 
	\includegraphics[valign=m,width=\imwidth\textwidth]{images/segmentation/loading/00016.jpg}
	 & 
 	\includegraphics[valign=m,width=\imwidth\textwidth]{images/segmentation/loading/00030.jpg}
	&
 	\includegraphics[valign=m,width=\imwidth\textwidth]{images/segmentation/loading/00047.jpg}
	&
 	\includegraphics[valign=m,width=\imwidth\textwidth]{images/segmentation/loading/00049.jpg}
	\\ [0.8cm]
	\footnotesize{RGMP \cite{Oh_2018_CVPR}}	& 
	\includegraphics[valign=m,width=\imwidth\textwidth]{images/rgmp/loading/00001.jpg}
	&
	\includegraphics[valign=m,width=\imwidth\textwidth]{images/rgmp/loading/00016.jpg}
	&
	\includegraphics[valign=m,width=\imwidth\textwidth]{images/rgmp/loading/00030.jpg}
	&
	\includegraphics[valign=m,width=\imwidth\textwidth]{images/rgmp/loading/00047.jpg}
	&
	\includegraphics[valign=m,width=\imwidth\textwidth]{images/rgmp/loading/00049.jpg}\\ [0.85cm]
	
	\footnotesize{Ours}	& 
		\includegraphics[valign=m,width=\imwidth\textwidth]{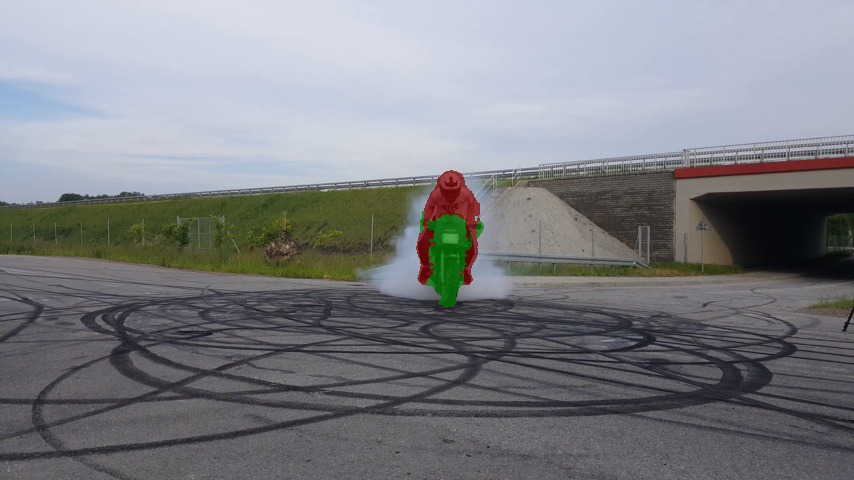}
		&
		\includegraphics[valign=m,width=\imwidth\textwidth]{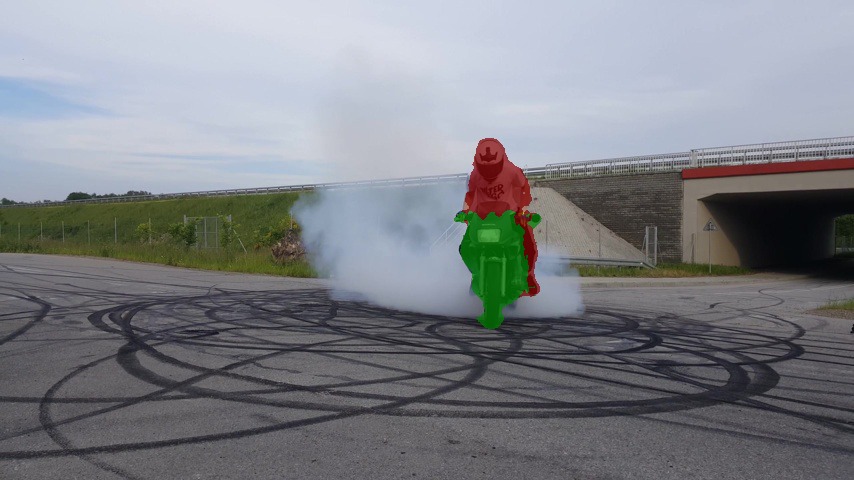}
		&
		\includegraphics[valign=m,width=\imwidth\textwidth]{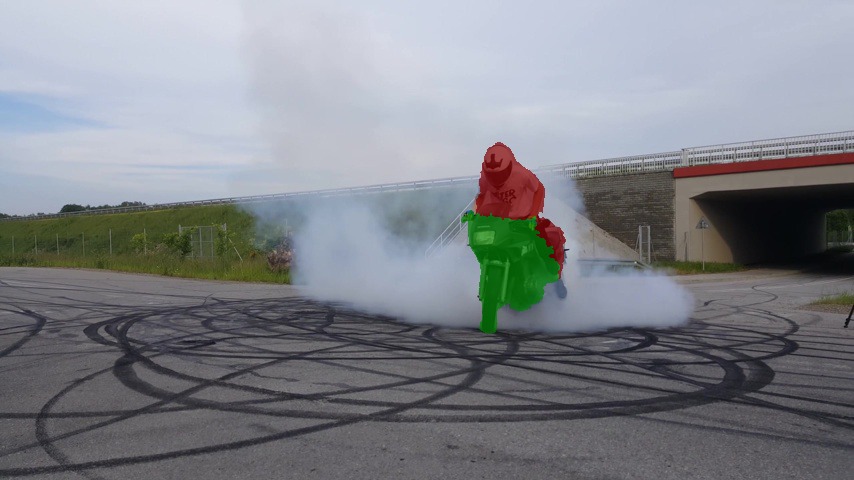}
		&
		\includegraphics[valign=m,width=\imwidth\textwidth]{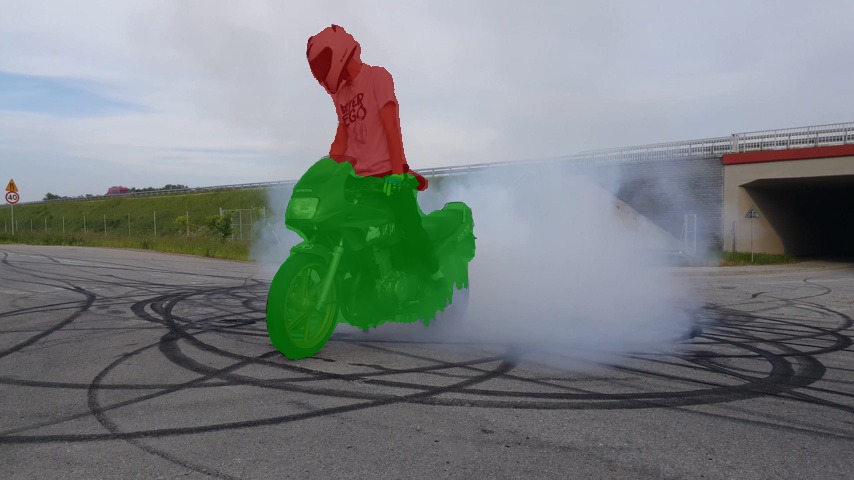}
		&
		\includegraphics[valign=m,width=\imwidth\textwidth]{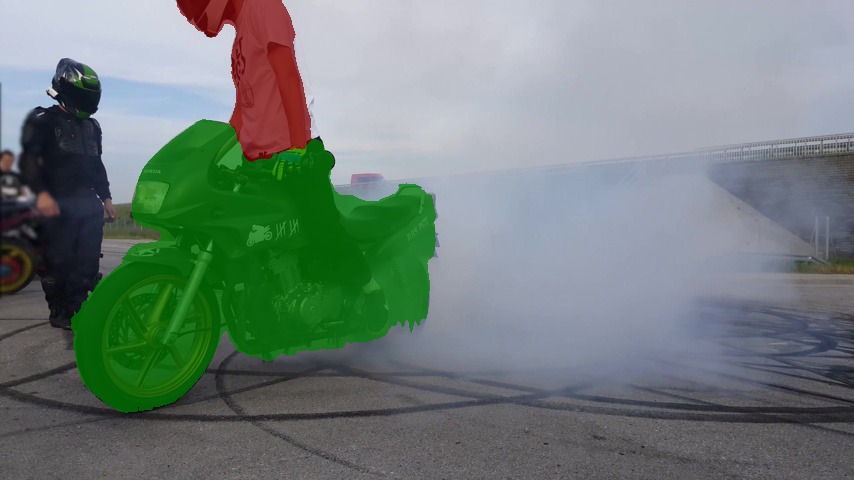} \\ [0.8cm]
		
	\footnotesize{RGMP \cite{Oh_2018_CVPR}}	&
		\includegraphics[valign=m,width=\imwidth\textwidth]{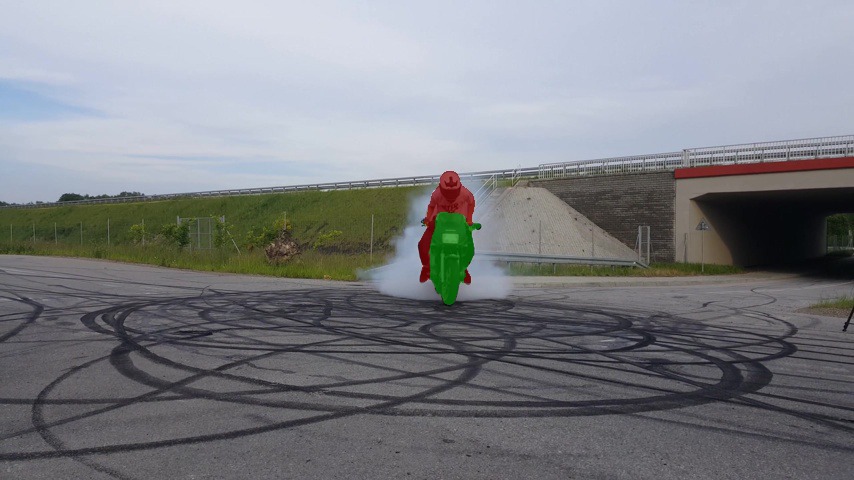}
		&
		\includegraphics[valign=m,width=\imwidth\textwidth]{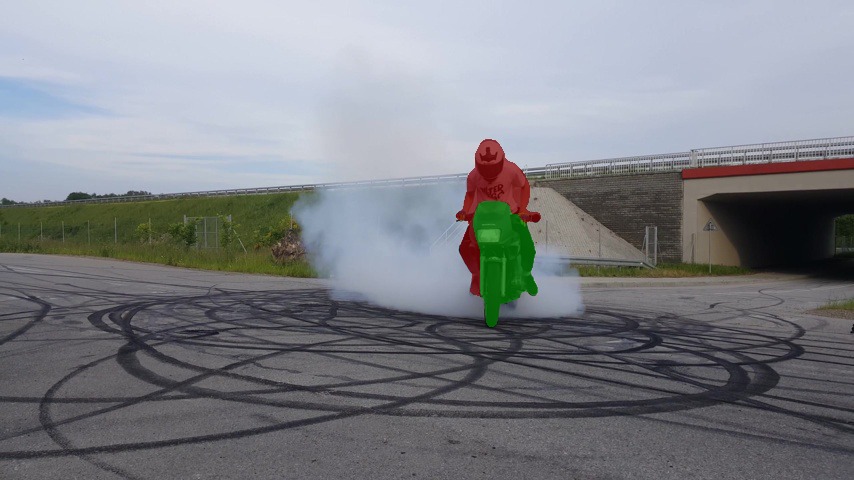}
		&
		\includegraphics[valign=m,width=\imwidth\textwidth]{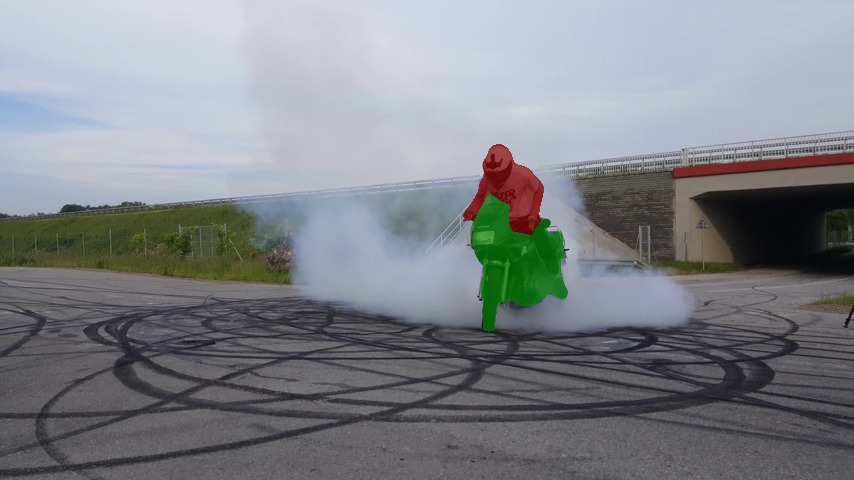}
		&
		\includegraphics[valign=m,width=\imwidth\textwidth]{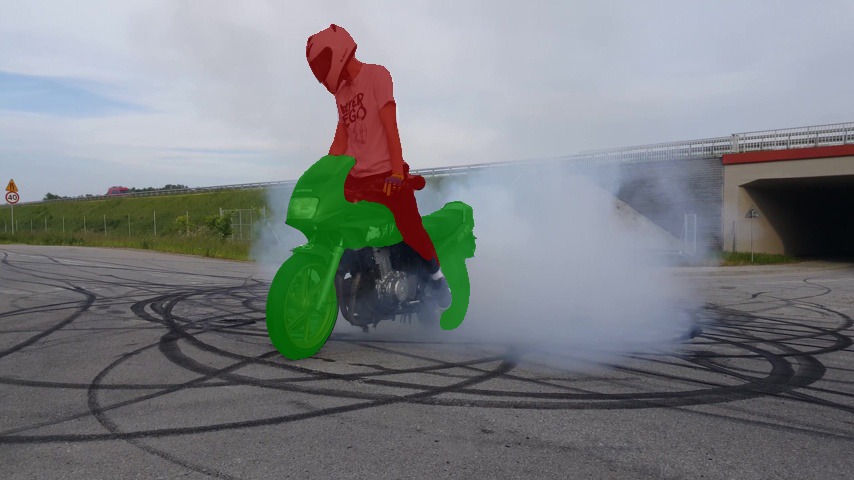}
		&
		\includegraphics[valign=m,width=\imwidth\textwidth]{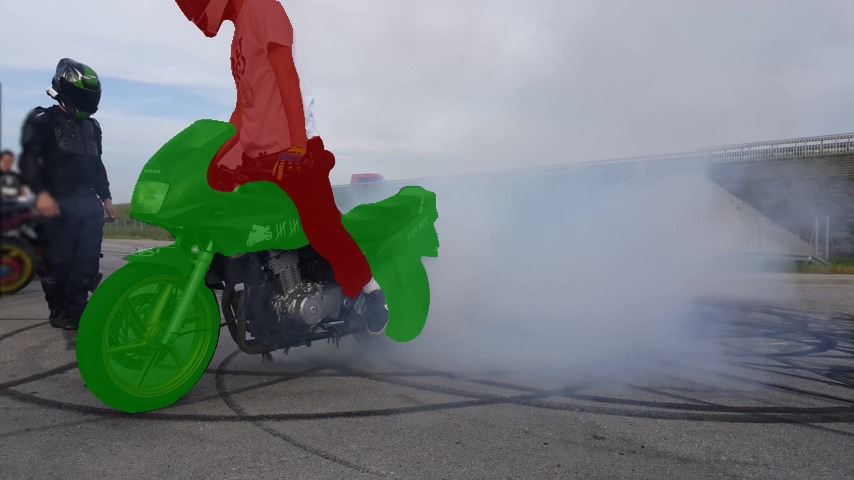} \\ [0.85cm]

	\footnotesize{Ours}	& 
	    \includegraphics[valign=m,width=\imwidth\textwidth]{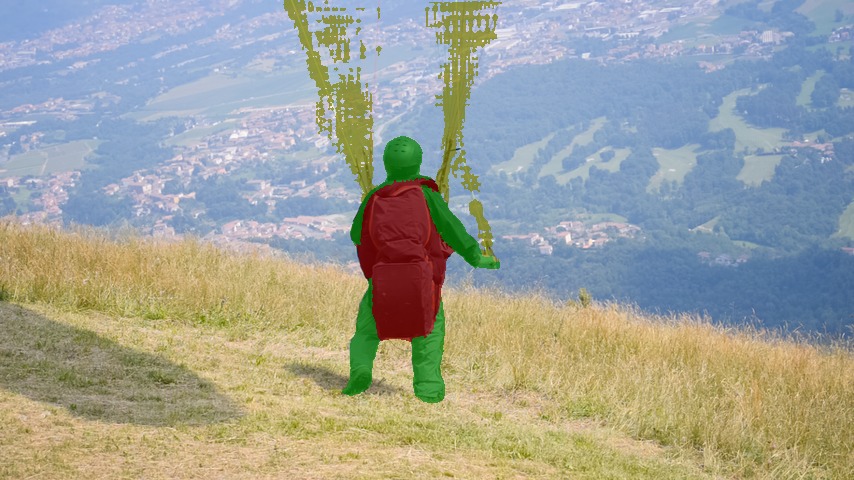}
		&
		\includegraphics[valign=m,width=\imwidth\textwidth]{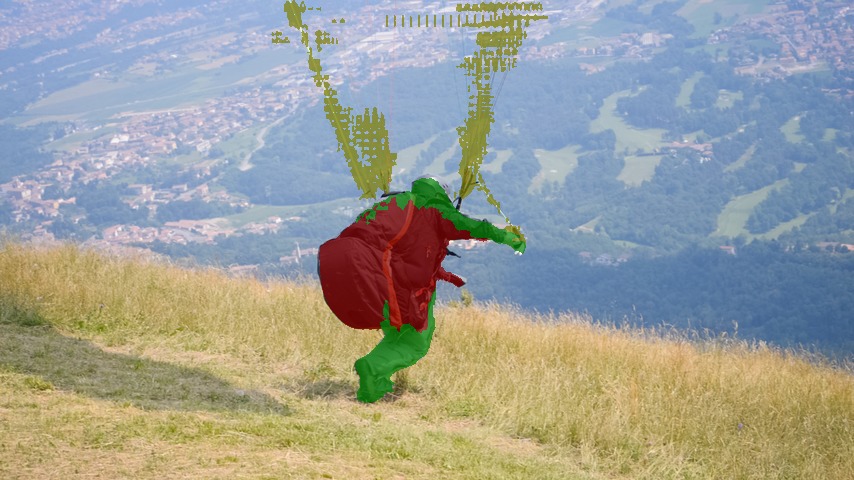}
		&
		\includegraphics[valign=m,width=\imwidth\textwidth]{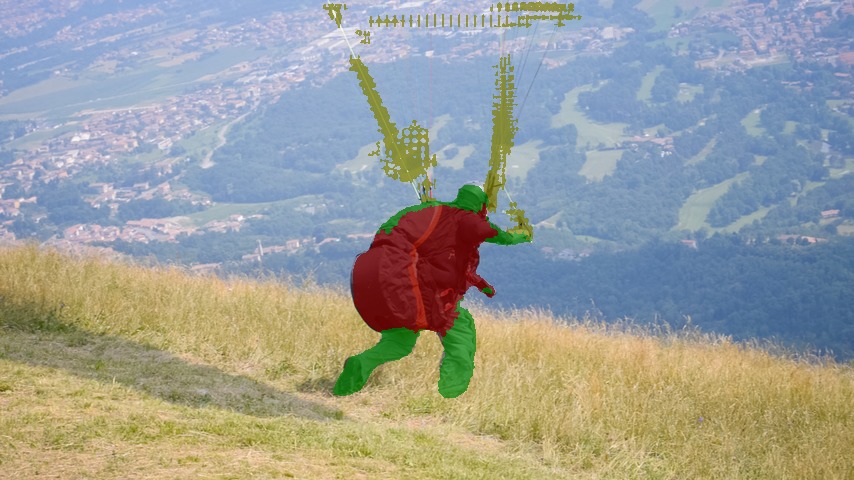}
		&
		\includegraphics[valign=m,width=\imwidth\textwidth]{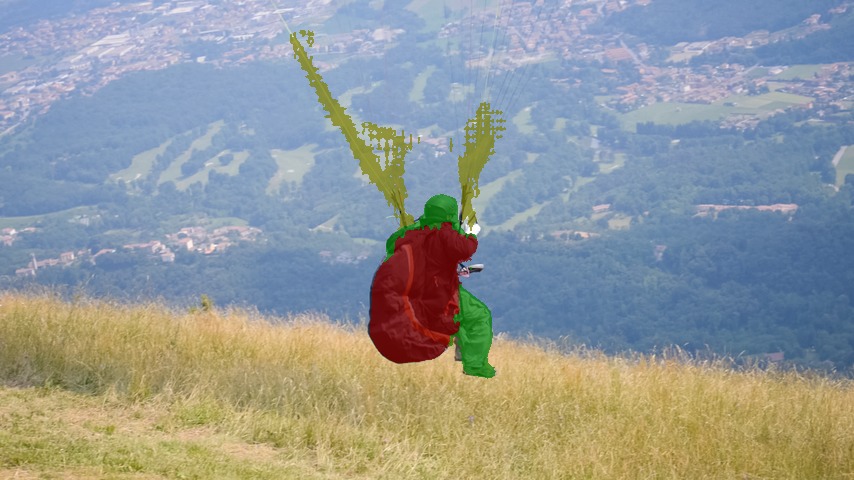}
		&
		\includegraphics[valign=m,width=\imwidth\textwidth]{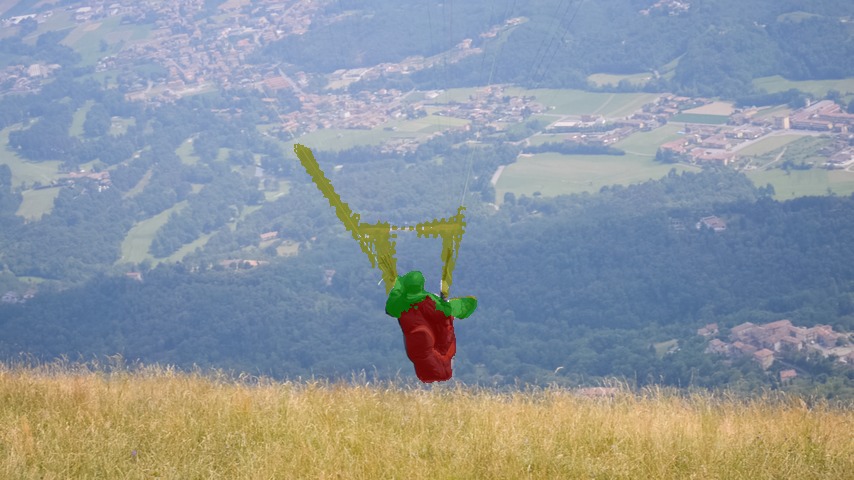} \\ [0.8cm]
		
	\footnotesize{RGMP \cite{Oh_2018_CVPR}}	&
	    \includegraphics[valign=m,width=\imwidth\textwidth]{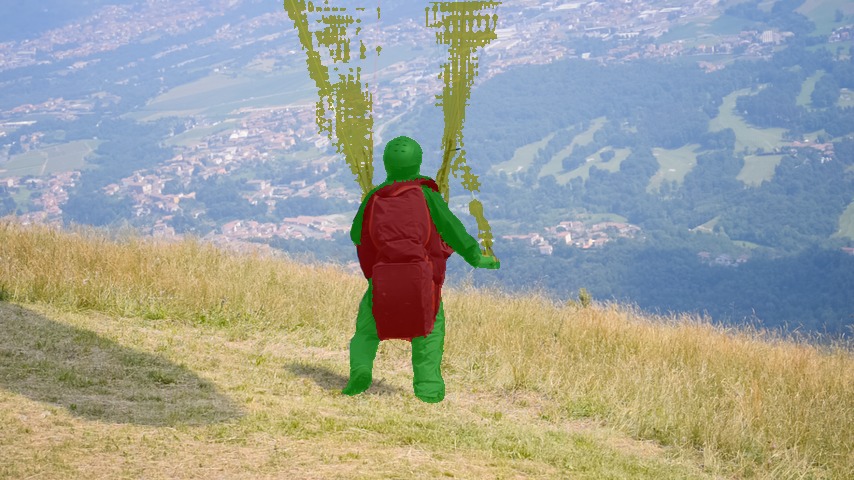}
		&
		\includegraphics[valign=m,width=\imwidth\textwidth]{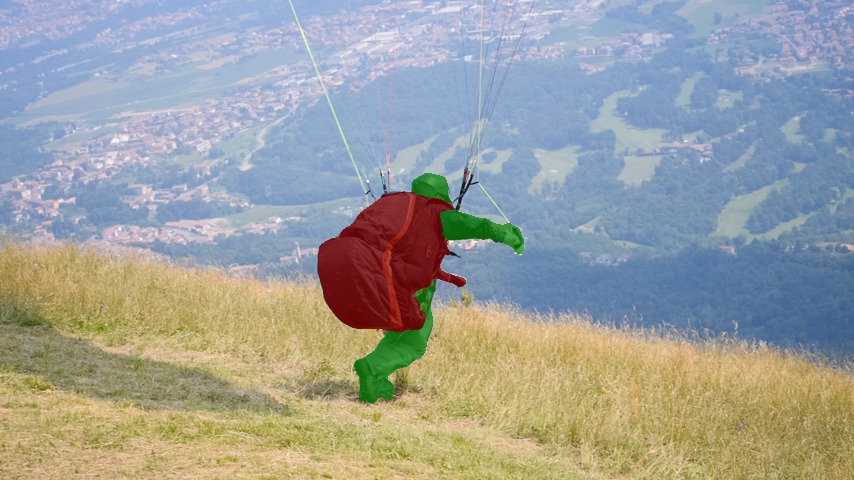}
		&
		\includegraphics[valign=m,width=\imwidth\textwidth]{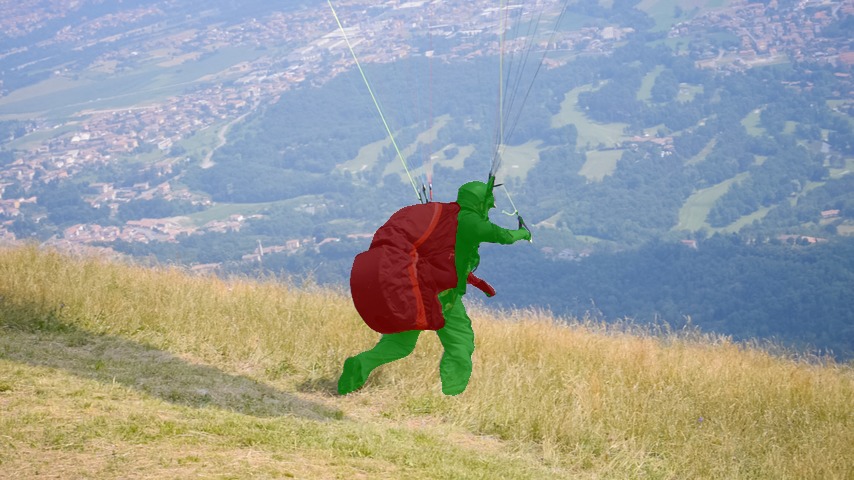}
		&
		\includegraphics[valign=m,width=\imwidth\textwidth]{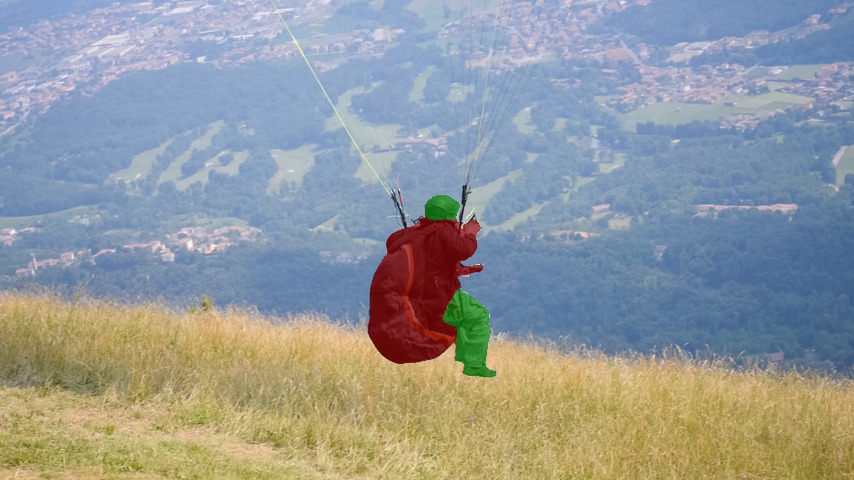}
		&
		\includegraphics[valign=m,width=\imwidth\textwidth]{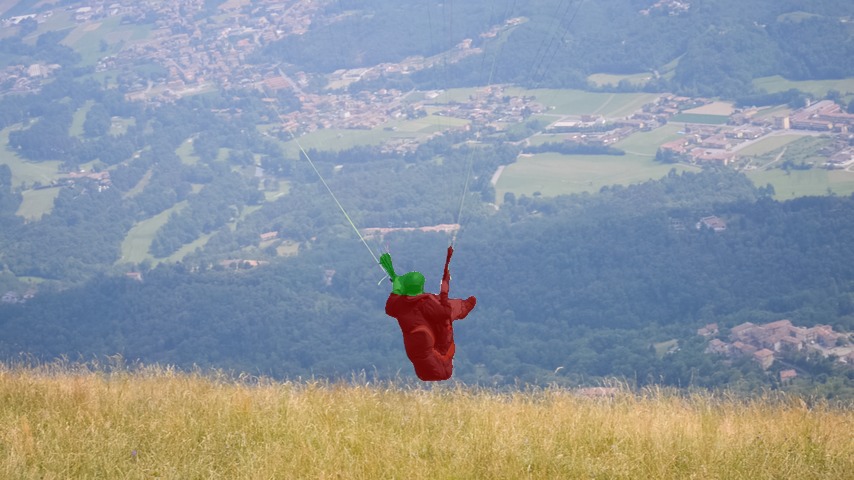} \\ [0.85cm]

	\footnotesize{Ours}	& 
	    \includegraphics[valign=m,width=\imwidth\textwidth]{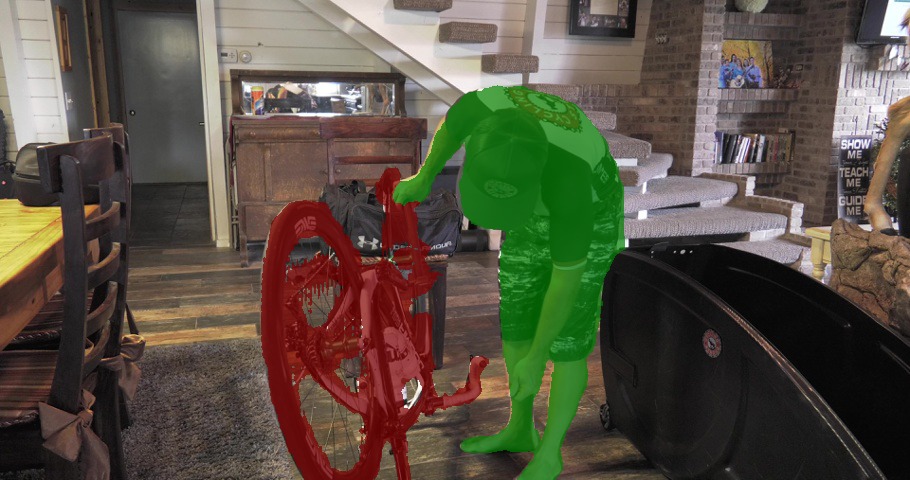}
		&
		\includegraphics[valign=m,width=\imwidth\textwidth]{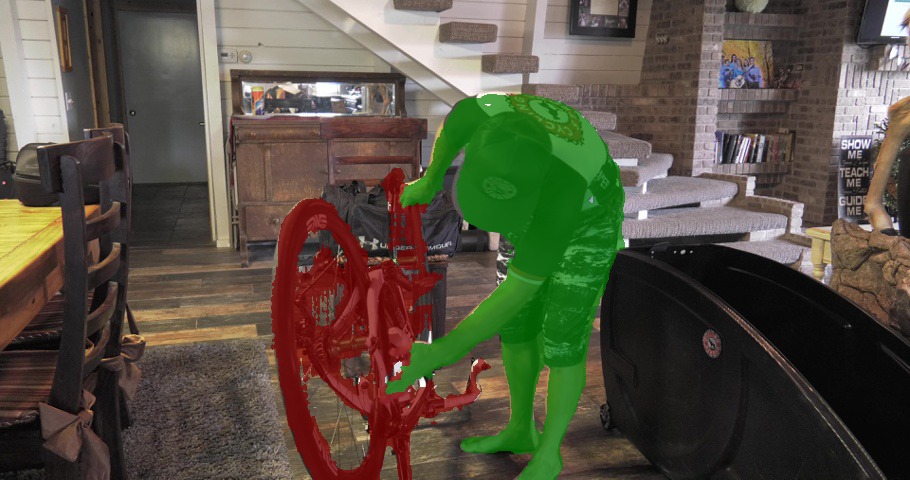}
		&
		\includegraphics[valign=m,width=\imwidth\textwidth]{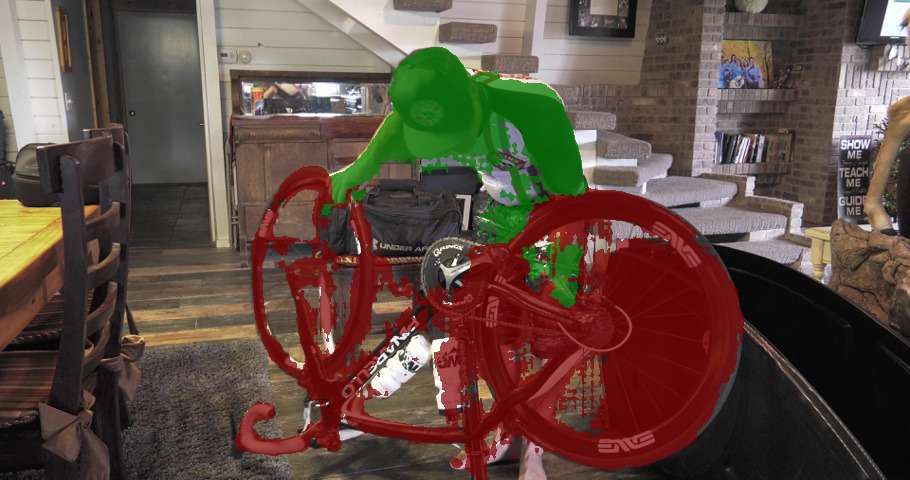}
		&
		\includegraphics[valign=m,width=\imwidth\textwidth]{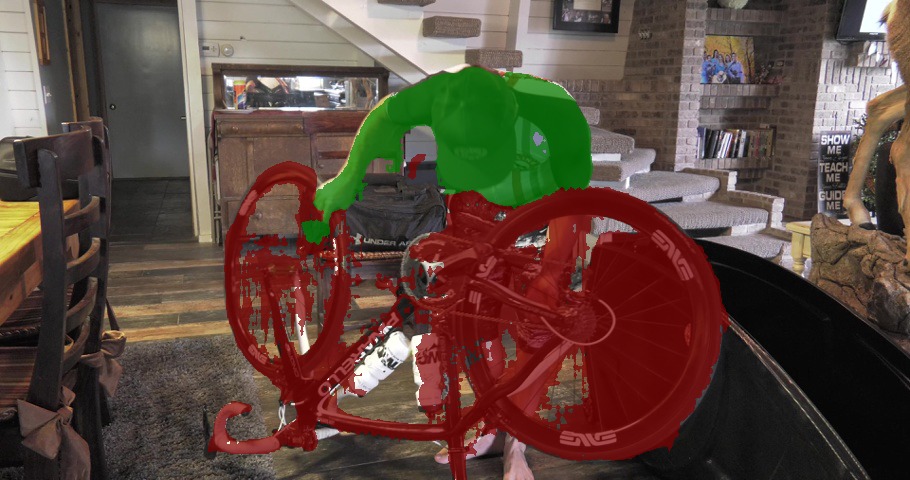}
		&
		\includegraphics[valign=m,width=\imwidth\textwidth]{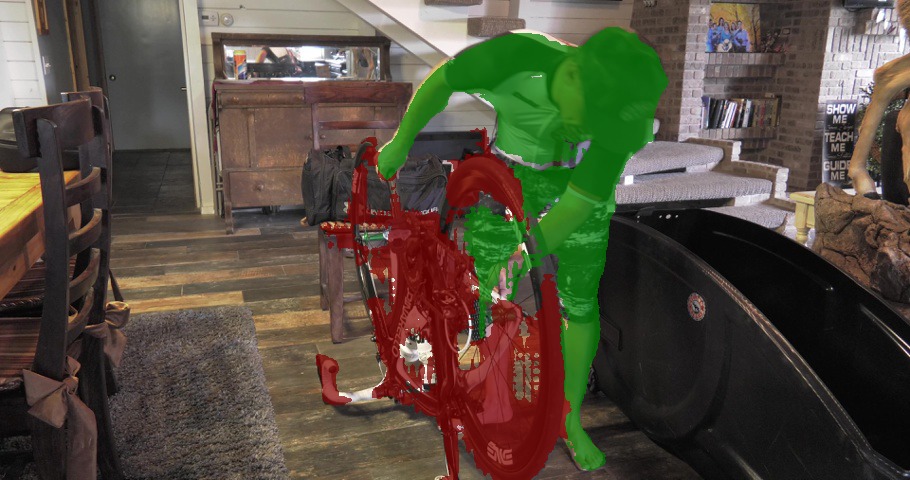} \\ [0.8cm]
		
	\footnotesize{RGMP \cite{Oh_2018_CVPR}}	&
	    \includegraphics[valign=m,width=\imwidth\textwidth]{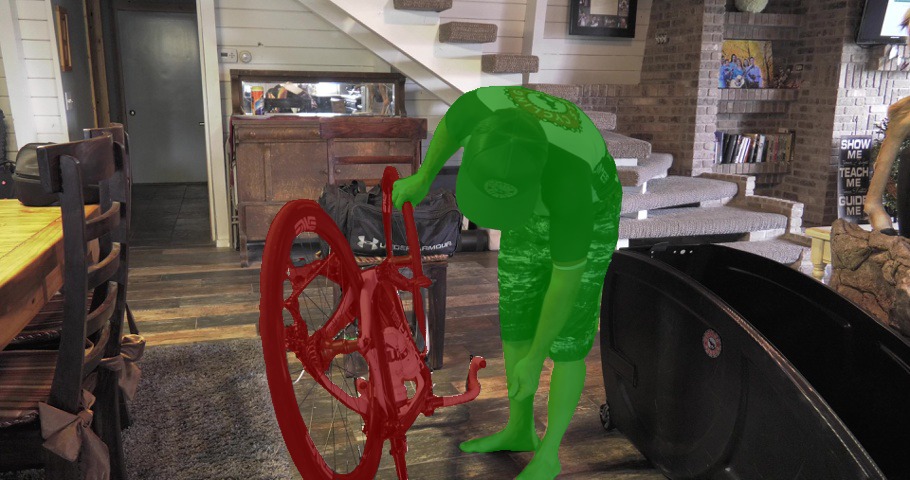}
		&
		\includegraphics[valign=m,width=\imwidth\textwidth]{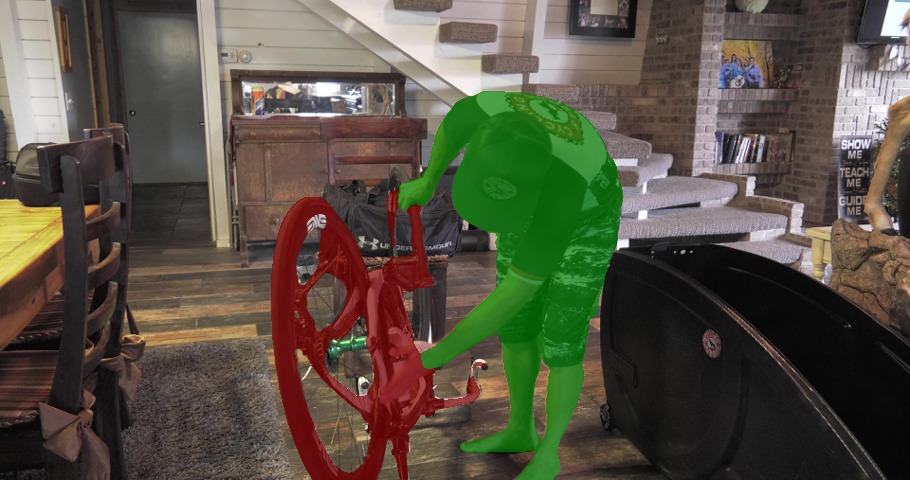}
		&
		\includegraphics[valign=m,width=\imwidth\textwidth]{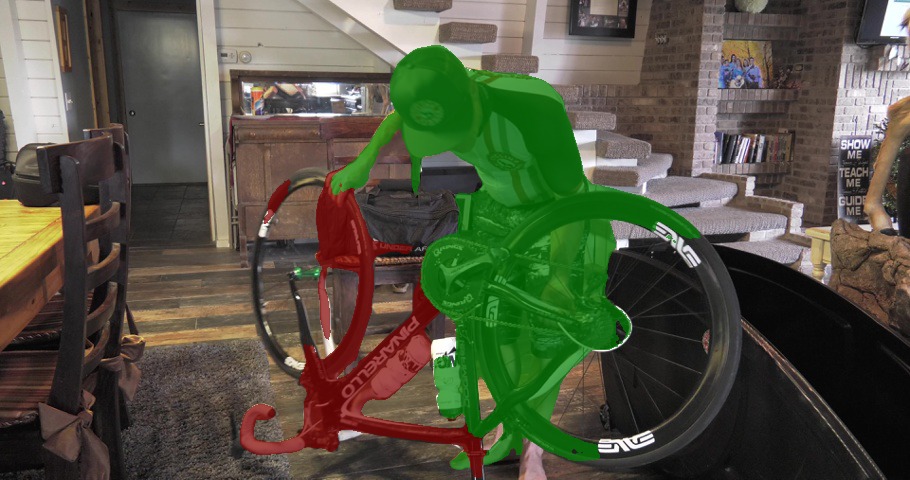}
		&
		\includegraphics[valign=m,width=\imwidth\textwidth]{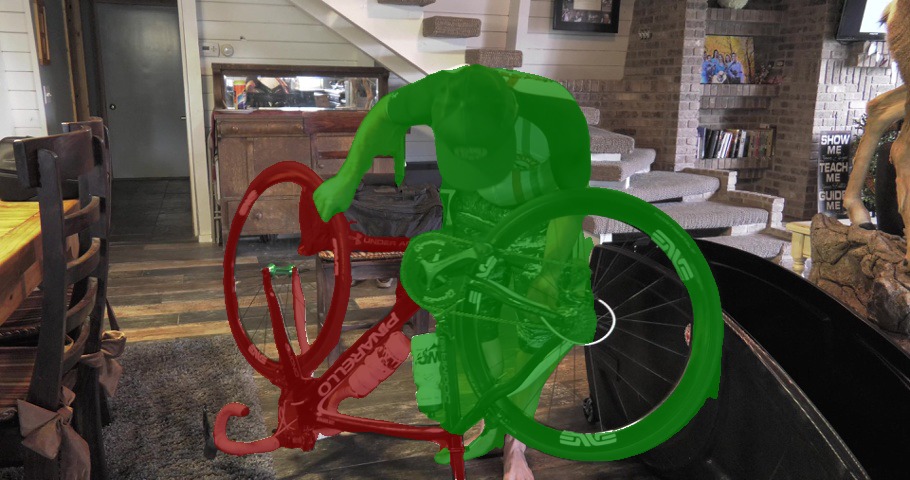}
		&
		\includegraphics[valign=m,width=\imwidth\textwidth]{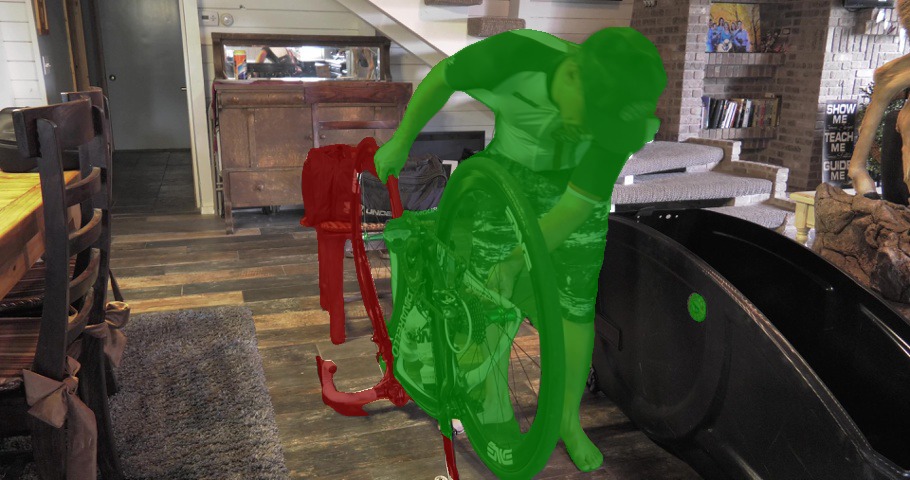} \\ [0.85cm]

	\end{tabular}
	\vspace{\baselineskip}
	\caption{\textbf{Failure cases of our method and RGMP \cite{Oh_2018_CVPR}.} 
	\textit{First video:} Note how our method does not properly segment the green box through the video. RGMP, on the other hand, loses track of the whole box, and also cannot deal with the two people occluding each other in the last two frames.
	\textit{Second video:} In the fourth and fifth frames, our method confuses the cyclists legs and motorbike. RGMP segments the person properly, but not the entire motorbike.
	\textit{Third video:} Our algorithm struggles to segment the fine structures of the parachute. RGMP, on the other hand, completely loses track of the parachute after the first frame.
	\textit{Fourth video:} Our segmentation of the bicycle (particularly its spokes) is not very accurate. RGMP, on the other hand, makes a larger error between the bicycle and person when they occlude each other from the third frame onwards.
	Full video results of these clips are included in the supplementary video.
	} 
	\label{fig:comparison_qualitative_failure}
	\vspace{-\baselineskip}
\end{figure*}

\setlength{\tabcolsep}{6pt}

\clearpage
\onecolumn
\section{Additional quantitative results}
\label{sec:supp_quantitative}

Tables \ref{tab:sequence_davis_2016}, \ref{tab:sequence_davis_2017} and \ref{tab:sequence_youtube} present per-sequence results on the DAVIS-2016, DAVIS-2017 and YouTube-Objects datasets respectively.

\begin{table*}[!h]
\centering
\def\r{30}
\def\d{.7cm}
\small\addtolength{\tabcolsep}{1.5pt}
{
\caption{Per-sequence video object segmentation results for DAVIS-2016 \cite{DAVIS} dataset.}
\label{tab:sequence_davis_2016}	
\begin{tabular}{@{}lp{\d}p{\d}p{\d}p{\d}p{\d}p{\d}p{\d}p{\d}p{\d}p{\d}p{\d}p{\d}p{\d}p{\d}p{\d}p{\d}@{}}
\textbf{Sequence} & \rotatebox{\r}{\bf \scriptsize Blackswan} & \rotatebox{\r}{\bf \scriptsize Bmx-Trees} & \rotatebox{\r}{\bf \scriptsize Breakdance} & \rotatebox{\r}{\bf \scriptsize Camel}    & \rotatebox{\r}{\bf \scriptsize Car-Roundabout} & \rotatebox{\r}{\bf \scriptsize Car-Shadow} & \rotatebox{\r}{\bf \scriptsize Cows}    & \rotatebox{\r}{\bf \scriptsize Dance-Twirl} & \rotatebox{\r}{\bf \scriptsize Dog}     &  \rotatebox{\r}{\bf \scriptsize Drift-Chicane}       \\ 
\hline
F                 & 0.98                    & 0.77                    & 0.70                 & 0.85                 & 0.92                 & 0.99                  & 0.98                & 0.81                      & 0.95                  & 0.72 \\
\hline
J                 & 0.94                    & 0.55                    & 0.69                 & 0.81                 & 0.96                 & 0.96                  & 0.94                & 0.81                   & 0.93                    & 0.65 \\ 
\hline
\textbf{Sequence}  & \rotatebox{\r}{\bf \scriptsize Drift-Straight}     & \rotatebox{\r}{\bf \scriptsize Goat}  & \rotatebox{\r}{\bf \scriptsize Horsejump-High}  & \rotatebox{\r}{\bf \scriptsize Kite-Surf}   & \rotatebox{\r}{\bf \scriptsize Libby}  & \rotatebox{\r}{\bf \scriptsize Motocross-Jump}  & \rotatebox{\r}{\bf \scriptsize Paragliding-Launch}  & \rotatebox{\r}{\bf \scriptsize Parkour}   & \rotatebox{\r}{\bf \scriptsize Scooter-Black}   & \rotatebox{\r}{\bf \scriptsize Soapbox} \\  
\hline
F                                & 0.83                      & 0.91                          & 0.93                            & 0.64                          & 0.89                 & 0.65                 & 0.45                 & 0.94                & 0.82                      & 0.89 \\
\hline
J                               & 0.90                     & 0.87                           & 0.82                         & 0.62                          & 0.75                 & 0.82                 & 0.61                 & 0.89                 & 0.85                          & 0.90 \\
\hline

\end{tabular}
}
 \vspace{0.3cm}
\end{table*}

\begin{table*}[!h]
	\centering
	\def\r{30}
	\def\d{.26cm}
	\small\addtolength{\tabcolsep}{1pt}
	{
		
		\caption{Per-sequence video object segmentation results for DAVIS-2017 \cite{davis_2017} dataset.}
		\label{tab:sequence_davis_2017}
		
		\begin{tabular}{@{}lp{\d}p{\d}p{\d}p{\d}p{\d}p{\d}p{\d}p{\d}p{\d}p{\d}p{\d}p{\d}p{\d}p{\d}p{\d}p{\d}@{}}
			
		\textbf{Sequence} & \rotatebox{\r}{\bf \scriptsize Bike-Packing\_1} & \rotatebox{\r}{\bf \scriptsize Bike-Packing\_2} & \rotatebox{\r}{\bf \scriptsize Blackswan\_1} & \rotatebox{\r}{\bf \scriptsize Bmx-Trees\_1} & \rotatebox{\r}{\bf \scriptsize Bmx-Trees\_2} & \rotatebox{\r}{\bf \scriptsize Breakdance\_1} & \rotatebox{\r}{\bf \scriptsize Camel\_1}    & \rotatebox{\r}{\bf \scriptsize Car-Roundabout\_1} & \rotatebox{\r}{\bf \scriptsize Car-Shadow\_1} & \rotatebox{\r}{\bf \scriptsize Cows\_1}    & \rotatebox{\r}{\bf \scriptsize Dance-Twirl\_1} & \rotatebox{\r}{\bf \scriptsize Dog\_1}     & \rotatebox{\r}{\bf \scriptsize Dogs-Jump\_1}   & \rotatebox{\r}{\bf \scriptsize Dogs-Jump\_2}   & \rotatebox{\r}{\bf \scriptsize Dogs-Jump\_3}      & \textbf{}        \\ 
			\hline
			F                 & 0.81                    & 0.63                    & 0.98                 & 0.72                 & 0.73                 & 0.81                  & 0.88                & 0.96                      & 0.99                  & 0.98               & 0.83                   & 0.95               & 0.12                    & 0.70        & 0.92   &                     \\
			\hline
			J                 & 0.59                     & 0.72                    & 0.95                 & 0.33                 & 0.57                 & 0.76                  & 0.82                & 0.97                      & 0.95                  & 0.94               & 0.82                   & 0.92               & 0.09                   & 0.55                    & 0.85                    &                \\ 
			\hline
			\textbf{Sequence} &  \rotatebox{\r}{\bf \scriptsize Drift-Chicane\_1}  & \rotatebox{\r}{\bf \scriptsize Drift-Straight\_1}     & \rotatebox{\r}{\bf \scriptsize Goat\_1}               & \rotatebox{\r}{\bf \scriptsize Gold-Fish\_1}          & \rotatebox{\r}{\bf \scriptsize Gold-Fish\_2} & \rotatebox{\r}{\bf \scriptsize Gold-Fish\_3} & \rotatebox{\r}{\bf \scriptsize Gold-Fish\_4} & \rotatebox{\r}{\bf \scriptsize Gold-Fish\_5} & \rotatebox{\r}{\bf \scriptsize Horsejump-High\_1} & \rotatebox{\r}{\bf \scriptsize Horsejump-High\_2} & \rotatebox{\r}{\bf \scriptsize India\_1}    & \rotatebox{\r}{\bf \scriptsize India\_2}    & \rotatebox{\r}{\bf \scriptsize India\_3}    & \rotatebox{\r}{\bf \scriptsize Judo\_1}    & \rotatebox{\r}{\bf \scriptsize Judo\_2}  & \textbf{} \\  
			\hline
			F                                & 0.70                      & 0.87                          & 0.90                            & 0.58                          & 0.59                 & 0.49                 & 0.60                 & 0.60                & 0.93                      & 0.89                     & 0.45               & 0.41                & 0.56                & 0.84                & 0.77              & \\
			\hline
			J   & 0.68                      & 0.91                           & 0.86                         & 0.62                          & 0.56                 & 0.60                 & 0.65                 & 0.75                 & 0.78                      & 0.67                      & 0.46                & 0.45                & 0.60                & 0.81               & 0.76               &     \\
			\hline

			\textbf{Sequence} & \rotatebox{\r}{\bf \scriptsize Kite-Surf\_1}    & \rotatebox{\r}{\bf \scriptsize Kite-Surf\_2}    & \rotatebox{\r}{\bf \scriptsize Kite-Surf\_3} & \rotatebox{\r}{\bf \scriptsize Lab-Coat\_1}  & \rotatebox{\r}{\bf \scriptsize Lab-Coat\_2}  & \rotatebox{\r}{\bf \scriptsize Lab-Coat\_3}   & \rotatebox{\r}{\bf \scriptsize Lab-Coat\_4} & \rotatebox{\r}{\bf \scriptsize Lab-Coat\_5}       & \rotatebox{\r}{\bf \scriptsize Libby\_1}      & \rotatebox{\r}{\bf \scriptsize Loading\_1} & \rotatebox{\r}{\bf \scriptsize Loading\_2}     & \rotatebox{\r}{\bf \scriptsize Loading\_3} & \rotatebox{\r}{\bf \scriptsize Mbike-Trick\_1} & \rotatebox{\r}{\bf \scriptsize Mbike-Trick\_2} & \rotatebox{\r}{\bf \scriptsize Motocross-Jump\_1} & \textbf{}\\ 
			\hline
			F                 & 0.67                    & 0.26                    & 0.94                  & 0.44                 & 0.51                 & 0.49                  & 0.41                & 0.62                      & 0.91                  & 0.89               & 0.55                   & 0.89               & 0.75                   & 0.76                   & 0.57                   &          \\ 
			\hline
			J                 & 0.28                    & 0.16                    & 0.72                 & 0.07                  & 0.13                 & 0.55                  & 0.43                & 0.66                      & 0.76                 & 0.93               & 0.47                   & 0.84               & 0.62                   & 0.75                   & 0.48                 &        \\ 
			\hline
			
			\textbf{Sequence} &\rotatebox{\r}{\bf \scriptsize Motocross-Jump\_2} & \rotatebox{\r}{\bf \scriptsize Paragliding-Launch\_1} & \rotatebox{\r}{\bf \scriptsize Paragliding-Launch\_2} & \rotatebox{\r}{\bf \scriptsize Paragliding-Launch\_3} & \rotatebox{\r}{\bf \scriptsize Parkour\_1}   & \rotatebox{\r}{\bf \scriptsize Pigs\_1}      & \rotatebox{\r}{\bf \scriptsize Pigs\_2}      & \rotatebox{\r}{\bf \scriptsize Pigs\_3}      & \rotatebox{\r}{\bf \scriptsize Scooter-Black\_1}  & \rotatebox{\r}{\bf \scriptsize Scooter-Black\_2}  & \rotatebox{\r}{\bf \scriptsize Shooting\_1} & \rotatebox{\r}{\bf \scriptsize Shooting\_2} & \rotatebox{\r}{\bf \scriptsize Shooting\_3} & \rotatebox{\r}{\bf \scriptsize Soapbox\_1} & \rotatebox{\r}{\bf \scriptsize Soapbox\_2} & \rotatebox{\r}{\bf \scriptsize Soapbox\_3} \\
			\hline
			F    & 0.52                      & 0.84                          & 0.83                           & 0.58                          & 0.95                 & 0.55                 & 0.62                 & 0.84                 & 0.77                      & 0.77                      & 0.31                & 0.55                 & 0.68                & 0.82               & 0.81               & 0.84     \\
			\hline
			J      & 0.69                      & 0.77                         & 0.58                          & 0.15                          & 0.90                   & 0.52                 & 0.48                & 0.93                 & 0.64                      & 0.80                      & 0.33                & 0.59                & 0.44                & 0.84               & 0.75               & 0.76       \\
			\hline
		\end{tabular}
	}
\end{table*}

\begin{table*}[!h]
	\centering
	\def\r{30}
	\def\d{.7cm}
	\addtolength{\tabcolsep}{1.5pt}
	{
		\caption{Per-sequence video object segmentation results for Youtube-Objects dataset~\cite{youtube1, youtube2}.}
		\label{tab:sequence_youtube}
		\resizebox{0.8\textwidth}{!}{
		\begin{tabular}{@{}lp{\d}p{\d}p{\d}p{\d}p{\d}p{\d}p{\d}p{\d}p{\d}p{\d}@{}}
			\textbf{Sequence} & \rotatebox{\r}{\bf \small Aeroplane} & \rotatebox{\r}{\bf \small Bird} & \rotatebox{\r}{\bf \small Boat} & \rotatebox{\r}{\bf \small Car}    & \rotatebox{\r}{\bf \small Cat} & \rotatebox{\r}{\bf \small Cow} & \rotatebox{\r}{\bf \small Dog}    & \rotatebox{\r}{\bf \small Horse} & \rotatebox{\r}{\bf \small Motorbike}     &  \rotatebox{\r}{\bf \small Train}       \\ 
			\hline
			${\mathcal{J}}${\footnotesize (\%)} & 87.1                  & 84.7                    & 81.0                 & 82.9                & 78.8                 & 76.3                  & 80.3                & 72.9                     & 77.8                  & 89.5 \\
			\hline
		\end{tabular}
	}
	}
\end{table*}

\end{document}